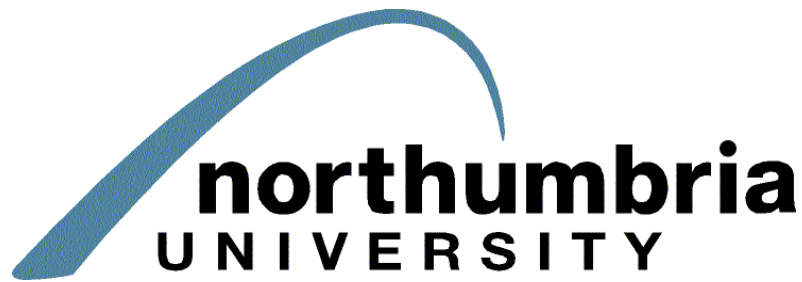

# School of Computing, Engineering and Information Sciences

## CG0174 Individual Computing Project, MSc

# Intelligent Search Optimization using Artificial Fuzzy Logics

Student: **Jai Manral**

Programme: **Computer Science, MSc**

Supervisor: **Prof. Alamgair Hossain**

Second Marker: **Prof. Michael Brockway**

Session: **2011-2012**

# 1 Declaration

The copyright of this dissertation rests with the author. No part of it should be published without his prior written consent and any information derived from this report should be acknowledged.

I certify that the work contained in this report is the sole work of the author except where indicated. All material that has been taken from other sources has been clearly acknowledged. Quotations from other sources have been clearly marked, using quotation marks or a block quote.

Signed……Jai Manral………………..

Date………17/05/2012………………...

# 2 Acknowledgement

The author expresses heartfelt gratitude to the following highly inspiring individuals, whose time, advice and guidance were extremely valuable and greatly appreciated, during the development of the project.

- Professor Alamgir Hossain whose encouragement, supervision and support from the preliminary to the concluding level enabled me to develop an understanding of the subject.

- Professor Michael Brockway who showed faith in the research and provides valuable insight in it.

- Dr. Becky Strachan whose teaching and guidance helped me in writing this research work.

  .
- Shelagh Keogh for her kind words of encouragement that made an otherwise hard journey seems infinitely simple.

I do like to thank all the interviewees and students who were patient while answering questions and reviewing product.

Finally, to the much needed support and understanding from family and friends that made this day possible.

# 3 Abstract

Information on the web is prodigious; searching relevant information is difficult making web users to rely on search engines for finding relevant information on the web. Search engines index and categorize web pages according to their contents using crawlers and rank them accordingly. For given user query they retrieve millions of webpages and display them to users according to web-page rank. Every search engine has their own algorithms based on certain parameters for ranking web-pages. Search Engine Optimization (SEO) is that technique by which webmasters try to improve ranking of their websites by optimizing it according to search engines ranking parameters.

It is the aim of this research to identify the most popular SEO techniques used by search engines for ranking web-pages and to establish their importance for indexing and categorizing web data. The research tries to establish that using more SEO parameters in ranking algorithms helps in retrieving better search results thus increasing user satisfaction.

In the accomplished research, a web based Meta search engine is proposed to aggregates search results from different search engines and rank web-pages based on new page ranking algorithm which will assign heuristic page rank to web-pages based on SEO parameters such as title tag, Meta description, sitemap etc. The research also provides insight into techniques which webmasters can use for better ranking their websites in Google and Bing.

Initial results has shown that using certain SEO parameters in present ranking algorithm helps in retrieving more useful results for user queries. These results generated from Meta search engine outperformed existing search engines in terms of better retrieved search results and high precision.

# 4 Table of content

# 5 List of Figures

# 6  List of Tables

# 7 List of Abbreviations

| Abbreviation | Word/Phrase | Explanation |
|---|---|---|
| *Backlink* | inbound links/in links | Links which are diverted towards webpages |
| *Black hat* | SEO technique | Search Engine banned SEO techniques which aim to provide better web-page ranking unethically. Mostly used by spammers and advertisers |
| *IR* | Information retrieval | Area of science dealing in searching for documents |
| *Keyword* | Query term | Word/phrase enter by user in search engine to find relevant information |
| *MSE* | Meta Search Engine | Web based searching tools which retrieve results from various search engines |
| *PPC* | Pay per click | Internet advertising model which pays website for every ad clicked by user in webpage |
| *PR* | Page Rank | Methods used by search engines for ranking pages in search results |
| *SE* | Search Engine | Web-based searching tool which retrieve documents in web for user query |
| *SEO* | Search Engine Optimization | Process to increase web site visibility/ranking in search engines |
| *SERP* | Search Engine Result Pages | Listing of search results for query given by users |
| *White Hat* | SEO techniques | Process to increase website ranking in search results by improving its content/structure. |
| *WWW* | World Wide Web | Network of hyperlinked documents |

This space intentionally left blank

# 1   Introduction

## 1.1  Motivation

World Wide Web has grown to be the biggest source of information from last decade. This has already changed the way people socialize, do business, entertain themselves and use it for finding information (Berners-Lee and Fischetti, 2000). The tremendous growth and amorphous nature of the Web makes it hard for users to find relevant information.  Evolution of search engines changed the way user interacts with the web. Users rely on search engines for extracting relevant information from among enormous data on the web (Berman and Katona, 2011). This trend has made search engines as information-seeking vehicle and online advertisement medium. Search engines index web-pages and categorize them according to its content. For any user query, search engines returns millions of results, ranked according to their ranking parameters (Evans, 2007). According to Jansen and Spink (2006, cited in Evans, 2007) 73% users never look beyond first page of search results. This has resulted in websites competing for higher rank in search engines.

In order to achieve higher ranking in search engine results, webmasters use certain techniques for optimizing their websites, known as SEO (Search Engine Optimization). SEO can be defined as an art which helps websites to be more visible to search engines. Understanding those factors which influence page ranking of search results is crucial for webmasters/developers (Evan, 2007). Although the market of SEO has crossed million dollar mark, not much research is done in this area (Berman and Katona, 2011; Gupta and Aggarwal, 2012). A study conducted by (Pringle *et al.*, 1998) suggests the use of certain SEO parameters to improve web-page ranking in search engines. Clay (2005) in his paper claims Pringle's study to be 'old' and 'inexplicit' for new era of search engines. In another research (Berman and Katona, 2011) do conclude that, using SEO improves the search engine's ranking and its user's satisfaction but their approach was based on Black Hat[1] techniques. The main motivation of this research is to analyse those White Hat[2] SEO techniques which is useful in indexing and categorising web data, hand out these techniques to webmasters for better optimizing their websites and develop a Meta search engine for the company (Abster-iT) which uses these techniques for providing better results to users query.  The focus of this paper is on white hat SEO techniques, which will be plainly called SEO throughout the paper.

## 1.2  Research Aims and Objective

The undertaken research tries to *find important SEO techniques used for ranking result pages by search engine; how they are important in indexing and categorizing web-data and by using them, optimize search results from search engine: Google and Bing, to show their importance in generating better results to user query.* It is important to achieve following objectives in order to accomplish the aim of this research:

- To find relevant literature from articles, web sources and interviews to understand the working of search engines, i.e. how they retrieve data from web, categorize them and present it to users.
- To develop new ranking algorithm for retrieving better results to user query.
-  To demonstrate the weightage of those SEO factors important in optimizing websites and improve ranking in search engines result pages.
- To develop an intelligent Meta search engine, this can retrieve better results than present search engines.
- To help webmasters by providing important SEO techniques for their websites.

## 1.3  Work Done and Results

Literature review for this topic revealed many claims of using SEO techniques in indexing and categorizing web data but fails to show experimental proofs. Most studies on SEO are based on economic effects of it rather than its importance in information retrieval. Further research found SEO importance in ranking websites in search engine results. Some research focuses on using Meta search engines for understanding optimization parameters. Resource materials from interviews and online lectures helped in finding some relevant SEO methods. Practical work is conducted in two Steps: Step 1 aimed to find SEO techniques used by search engines for indexing, categorizing and ranking webpages. Step 2 involves designing and implementing Meta search engine which collects search results from Google and Bing search engines. These results are then merged and ranked using page ranking algorithm which is trained to rank optimized[3] webpages higher in ranking.

---

[1] Type of SEO techniques banned by search engines which improves the ranking of a website.
[2] SEO techniques which improve website content and structure for websites to rank higher in in search engine result page.

As part of experiment, 50 students were selected from university library and asked to use Google, Bing and iral Meta search engine for same 'query[4]' and rate their performance[5]. Precession[6] of iral Meta search engine outperformed Google and Bing. The results reveal that SEO techniques are not only important for ranking a webpage higher in search engines result page but using these techniques assist search engines in indexing and categorizing web data which helps user's to find relevant information based on keywords thus increases user satisfaction.

## 1.4 Structure of Thesis

The rest of dissertation paper is presented as follows: Section 2 is literature review which covers broad area of previous research on search engines, Meta search engines, search engine optimization and web-mining. Section 3 highlights the practical work done with brief overview of architecture, implementation and testing of proposed Meta search engine. Section 4 presents results, their analysis and evaluation confirming literature hypothesis to experimental results. Section 5 evaluates the purpose of this dissertation. Finally, section 6 summarizes the paper and provides future recommendation.

---

[3] Here optimized web-pages refer to those web pages which use certain selected SEO techniques. These techniques are listed in practical work done section.
[4] Query here indicates the keyword 'alcoholism' and 'local computer shop'. These terms were used for finding information from search engines.
[5] Performance here means 'which search engine provides better result for given query'
[6] Precision is the calculation done for estimating search engines performance for given query. It is given by number of relevant documents retrieved by total number of documents retrieved by search engine.

## 2 Literature Review

## 2.1 Introduction

The information on the web is prodigious; searching relevant information is difficult for users. The accuracy of search results is measured by relevancy of query term to web-pages ranked and displayed by search engines. Even a small query term generates thousands of pages, which if not sorted can be difficult to retrieve relevant information. Search engines work hard on structuring and ranking these search results. Different methods and approaches for retrieving and ranking these search results have been suggested. There are several parameters on which search engine assign page ranking to webpages. In order to understand these parameters, we need a detailed study about search engines and their page ranking strategies.

## 2.2 Search Engines

### 2.2.1 Overview

Search engines can be defined as the most useful and high profile resources available on the internet. These are powerful tools used to assist the users to find the information in a large pool of data in World Wide Web (Dreilinger and Howe, 1997). According to Beigi *et al.*1998, Explosive growth of world wide web resulted in motivation of development of search engines to assist the users for finding desired information from pool of unmanageable data. Search engines presents information in real time by using numerous algorithms and are thus better than web directories; which stores information in databases and match query for retrieving results. Some popular search engines are Google, Yahoo, Bing, MSN and Alta Vista.

### 2.2.2 Evolution

In the early years when World Wide Web was relatively small finding information on the internet was done by using Web-directories. By 1990's it was observed that human powered categorization model was insufficient for faster and better search in World Wide Web (Metaxas, 2009). As elaborated in (Bowmen, 1994) Archie; the first ever search engine created by Alan Emtage, 1990; index files and menu name of FTP by using script based information gatherer. This system supports only name based searches by matching user query to its database. Other models developed were Veronica, which suffers from same problem of indexing. It served the same purpose as Archie while working on plain text file working for Gopher (Weiss, *et al.*, 1996). These early developments of search engines focused mainly on its ability to index web resources and retrieve the best results for users query. Studies on web search engines were small and publications like (Wildstrom, 1995; Shirky, 1995; Taubes, 1995) are descriptive in nature.

With the introduction of web crawlers, the World Wide Web Worm (McBryan, 1994) indexing of documents becomes effective and efficient. The WWWW was used to index the URLs and HTML files; using title string. By 1996, Infoseek and Lycos search tools make huge collection of documents by indexing WWW; allowing users to search using keyword-based queries (Gauch, Wang and Gomez, 1997). With emergence of Google (Brin and Page, 1998) led new standards in search engines. The unique way of ranking retrieved page increases its quality.

Search engines has received good attention lately (Funkhouser *et al.*, 2003; Cohen *et al.*, 2003; Strohman *et al.*, 2004; Theobald and Weikum, 2002; Lei *et al.,* 2006; Bajracharya *et al.*, 2006; Ferragina and Gulli, 2008; Horowitz and Kamvar, 2010; Hogan *et al.*, 2011) describes about new algorithms to search dynamic web pages, 3d objects and images.

Fuzzy logic based experimental search engines are described in (Widyantoro and Yen, 2001; Liu *et al.,* 2006;   Aquin and Motta, 2011). Wolfram Alpha are new generation answer engines which performs computation using information retrieval and NLP to answer the query rather than providing list of documents (Gray, 2012).

### 2.2.3  Classification

 According to Lin, Tang and Chun (2011), search engines can be classified in three categories; robot based, directory based and Meta search engines; according to information collection and service deliverably. Robot based search engines use software robots to collect and index web sites, download documents in their databases having large indexes. Upon receiving a query they search their database to generate results relevant to query. Examples are Infoseek (1994), Alta Vista and WebCrawler. Directory based search engines organize resources in tree structured directories sorted according to subject area. They collect information by artificial based or by the authors of the websites. Meta search engines are based on multiple individual search engines to extract the data.

### 2.2.4  Limitations

Size of internet and its exponential growth (Gebert *et al.*, 2012) has makes it impossible for any single search engine to index more than $1/3^{rd}$ of 'index able' web (Information retrieval on internet). Crawling web pages for information gathering and storing them to enormous databases is costly and time consuming (Joachims, 2002). The researcher thus designed a Meta search engine for experiments.

## 2.3  Meta search engine

### 2.3.1  Overview and Importance

A Meta search engine transmits users search query simultaneously to several search engine databases of websites individually and retrieve information from all those search engines queried (Shu and Kak, 1999). The purpose of creating a Meta search engine is to save lot of time by initiating a search which can be processed by different search engines simultaneously thus saving lot of time in studying different search engines, generating broad range of results (Mahabhashayam and Singitham, 2). The idea of developing a Meta search engine due to the fact that World Wide Web is a huge unstructured corpus of information which can be really difficult to index the web-pages for research. Incorporating various search engines such as Google, MSN and Bing under one Meta search engine helps to cover a large portion of World Wide Web, satisfying goals of maximizing search quality and minimizing resource consumption (Zilberstein, 1995 cited in Dreilinger and Howe, 1997).  Other important factor is web-crawling, process used by search engines to index websites in their databases. Web-crawling is a long and expensive process (Menczer *et al.*, 2001). Both storage cost of the database and time frame is important factors in it. As the content of websites keep changing from time to time thus keeping the updated information is rigid. Meta Search engine thus helps to eradicate this problem of crawling and indexing.

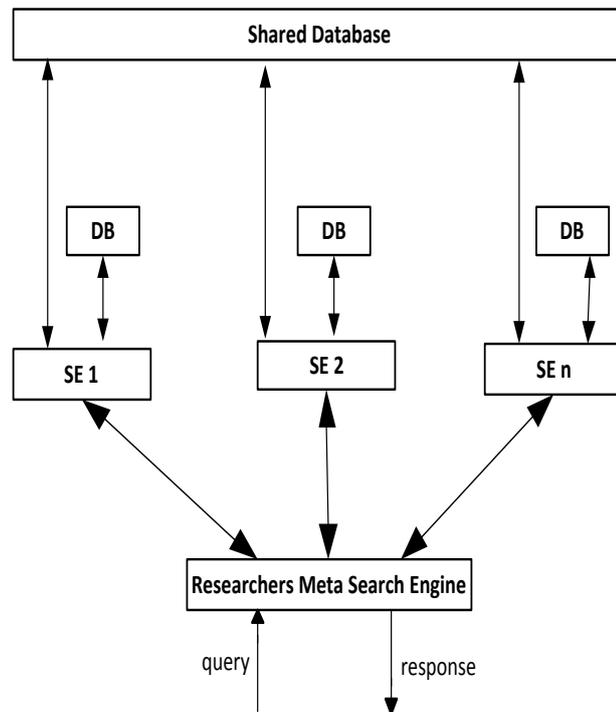

**Figure 1: Meta search engine basic architecture**

### 2.3.2 Previous work

Meta search engines has received lot of attention in academic researches, (Gauch *et al.,* 1996; Beigi *et al.*, 1998; Joachims, 2002; Lin *et al*,. 2011) uses meta search engines for generating search results based on their experimental information retrieval techniques. The concept of Meta search engine came from All-in-one Search page (Cross, 1996) in which users has to select search engines from pool of Search engine's. Using intelligent agents to retrieve search results were discussed in (Balabanovic *et al.,* 1995; Knoblock *et al.,* 1994; Chen and Liu, 2005). Selecting most promising search engine for information retrieval, SavvySearch (Dreilinger, 1996) was developed in which, query was send parallel to 2-3 search engines but results displayed were not merged. Neural network based intelligent Meta search engines (Gauch *et al.,* 1996; Shu and Kak, 1999; Mahabhashyam and Singitham, Tadpole; Hamdi, 2011). The ProFusion System (Gauch *et al.,* 1996) supports automatic and manual query dispatching, combing several features of other Meta search engines. Experimental Meta search engines focus on system architecture such as sorting algorithms (Huang *et al.*, 2000) and adaptive behaviour (Fan and Gauch, 1999).

Work of Aslam and Montague (2001) uses two models of Meta search engine. Their study was based to enhance the performance of Metasearch engines using two different algorithms, Borda-fuse and Bayes-fuse. ProThes (Braslavaski *et al.,* 2004) revamped the design of Metasearch engine by including a thesaurus component.

There are various Meta search engines available on the internet, categorized as following:

1. Deep study Meta Search engines for categorized based searches. Ex: Informiti (earlier known as Surfwax)
2. Meta search engines which do not aggregate the results before presenting them to users. Ex. Dogpile
3. MSE's which retrieve information from various search engines and aggregates them. Ex: Ixquick

## 2.4  Search Engine Optimization

### 2.4.1  Overview

A set of techniques or actions that a web-site can undertake in order to improve its ranking/position in search engine is known as search engine optimization. It essentially involves in providing a higher ranking in search results when a user submit related query to search engines.  This ranking is essential for generating higher traffic to the web-sites. As online search engines are popular tools for discovering information on the web, this has created new form of advertisement i.e. online ads.

Search engines generate their revenue from advertisement and sponsored links. To accommodate advertisers, search engine usually divide their result pages into organic and a sponsored part. These sponsored links are costly and not every website owner can bid for it. In order to gets a higher page ranking from among millions of websites, owners/developers of websites need to invest time in organic listing (Rutz and Bucklin, 2007). The collection of techniques for improving the organic rank of websites is called as search engine optimization (Berman and Katona, 2011). There are two type of SEO techniques mentioned in Kisiel, (2010) White hat SEO and Black hat SEO. Improving site content and user satisfaction comes under white hat techniques, while black hat techniques are used to increase the ranking of websites by eliciting external linking. Search engines are very cautious about SEO techniques and have special guidelines, failing which they remove sites from search results (Fortunato *et al.,* 2006).

### 2.4.2  Previous Work

Most of the research is based on economic aspects of search engine optimization. Work by (Rutz and Bucklin, 2007; Ghose and Yang, 2009) focus on advertising efficiency for business growth through search advertisements. Although SEO techniques are common in practice, researchers such as (Pasquale, 2007 and Mercadante, 2008) states, manipulating search results through any means has negative impact. The benefit of SEO techniques for generating the volume of traffics to company sites and thus increasing revenue was proposed by (Kisiel, 2010). Research by (Berman and Katona, 2011) concluded that, some level of SEO can be useful for search engines as it make it easy for indexing and categorizing web-sites thus providing better results to consumers. Article by (Joran *et al.,* 2010) proposes the idea of academic search engine optimization (ASEO). It proposed new techniques to optimize scholarly literature for Google Scholar and other academic search engines. Analysing search engines ability to identify relevant websites using robust and quality algorithms and optimizing web-sites were achieved in experiments by (Xing and Lin, 2006).

Some old studies on ranking search results using regression analysis and decision tress from search engines like AltaVista, Excite, InfoSeek and Lycos were conducted by (Pringle *et al.,* 1998). Their study concludes using keywords, Meta fields, informative titles, headings and Meta text as important factors for ranking in search results. This study is quite old and most of the search engines are out of business (Clay, 2005). In other study by (Bifet *et al.,* 2005) they used different factors for their ranking functions using SVM and compared their ranking with actual Google results for particular keywords, ex. 'Art'. The results highlighted other factors for ranking the websites in Google, moreover there SVM did not work and queries were arbitrary in nature (Evans, 2007). Using simple linear regression on predicting absolute PageRank (Khaki-Sedigh and Roudaki, 2003; Fortunato *et al.,* 2006) were conducted. Factors such as number of in-bound link (also known as backlink, referencing between web-sites) for PageRanking were done in Fortunato *et al.* (2006).

Some commercial studies conducted by SEO firms (Seoconsult.com, SearchEngineWatch.com, HighRanking.com, etc.), have partial knowledge of all the factors responsible for SEO. Most of these studies are based on trial and errors with no evidence (Fortunato *et al.,* 2006).

### 2.4.3  Optimizing Factors

The factors for optimizing websites can be categories as: 1. Keyword Analysis (Searching for best suitable keywords) 2.On-page optimization (set of techniques implied to web-pages) 3. Off-page optimization (efforts using social networking and back linking) (Slawski, 2011; Bifet and Castillo, 2005; Ledford, 2009)

### 2.4.3.1  Keyword Analysis

The quires entered by user in search engines are known as keywords and their combination is collectively known as key phrase. Keywords are important and essential for optimization of any web-site (Viney, 2008). A thorough research for keywords should be done before choosing a domain name as it accounts for 20% of SEO efforts (Viney, 2008). According to (Jones, 2010 p.14) it is the first step for optimization and arguably the most important one. An effective keyword is one which has less competition (no. of websites having same keyword) and high volume (number of search for a particular keyword over a period of time) of search. Keywords can either be head terms (one or two words with high search volume) or tail terms (three or more words with less search volume) (Jones, 2010 p.16). As an example, '*computer*' is a head term; keyword '*cheap computer for student*' is considered as a tail term.

Google suggestion tools are useful for finding the density of any keywords and its global search volume (Clifton, 2012 pp.12). Different other tools; Compete.com for analysing keyword competition, Keywordspy.com for keyword research on pay-per-click basis, Wordtracker.com for keyword suggestion are mentioned in (Jones, 2010 pp.14-35).

### 2.4.3.2  On-page optimization

These are the techniques which are implied on websites/web-pages for optimization. Search engines 'crawls' the content of websites to know its category or topic area it is relevant to. Advanced tools such as keyword analysis, content analysis and language analysis are used by search engines to categorize a certain web-page (Odom and Lynell, 2012 pp.16-18). For example a search for cricket should generate those pages which are related to it. Elements which need to be considered optimizing are:

a) **Page Title:** It tells about the topic of a web-page both to its users and search engines. It is the first element to get crawled by SE's (Odom and Lynell, 2012 pp.23-28). Google use page title in its result page as a summary. It is written in HTML as, `<title> content goes here</title>`

Title tag in homepage may contain the business/website name. There is no limit in title tag length but according to World Wide Web Consortium (W3C) standards, length should not increase more than 64 characters including spaces.

b) **Meta Keywords**: They are used to define the content of a web-page. They provide bunch of keywords specific to sites content. Most search engines (Google, Yahoo) penalize for abusing its use. They are widely used to provide synonyms to title tags (Odom and Lynell, 2012).

Syntax: `<meta name="keywords" content="keyword1, keyword2, keyuword3">`

c) **Meta Description:** These tags are used to describe web-pages in a short plain language text. They are often considered as summary of the websites. Ideally they are written in 20-30 words and can be seen in search engines result pages (SREP) below title tags.

Syntax: `<meta name="description" content="description goes here">`

d) **Meta Content:** It is used to declare character set of website. For example if the document is in UTF-8 punctuation character but been displayed in ASCII or ISO, will not cause any display problem ().

Synatx: `<meta http-equiv="Content-Type" content="text/html; charset=UTF-8">`

e) **Meta Expires tag:** This tag tells the search engine about expiry of the web-page. It can be set for any date and time or never.

Syntax : `<meta http-equiv=''expires'' content=''Thr,12 Jul 2012 8 :30 :00 GMT''>`

**f) Meta Revisit tag:** It tells the web-crawler when to come indexing next. Usually used in those websites which change their content frequently. Syntax: `<meta name="revisit-after" content="12 days">`

**g) Meta Robots:** It tells search engines either to index the web-page or not to. If there are certain pages in the website which are not required to be indexed, value can be set to 'noindex'. Syntax: `<meta name=''robot'' content=''noindex''/>`

For Google: `<mata name="GOOGLEBOT" content="NOARCHIVE">`

**h) Meta Distribution:** It defines the degree of distribution for website and its classification based on World Wide Web distribution methods (). Three form of degree are (a) Global- webpage intended for world wide, (b) Local-intended for local coverage and (c) Internal use (IU)- not intended for others

Syntax: `<meta name=''distribution'' content=''global''>`

**i) Other Optional Tags**

i. Meta Author- It display the author of the webpage or website.

   Syntax: `<meta name="Author" content="Jai Manral, jaimanral@northumbria.co.uk">`

ii. Meta Copyright- it includes information regarding patents, copyright, trademark etc.

   Syntax: `<meta name="copyright" content="Copyright 2012">`

iii. Meta Language- It is used to declare, in what language website is made. It is recommended only for those websites which are build outside English speaking countries.

   Syntax: `<meta name="Language" content="french">`

**j) Image Alt Attributes:** Search engine crawlers fails to read image contents in websites whereas human eyes can interpret the image to its meaning. By providing text description of images, it become easy for crawlers to better understand the meaning of website if it has lot of images embed in it. It is also helpful for image search as the alt text (used to define image) is treated similar as anchor text (Odom and Lynell, 2012 pp.76-80).

```
Syntax: <img src=''pic1.jpg'' alt=''descriptive text
goes here''>
```

**k) Breadcrumb Trial:** It is navigational tool used to show user path or hierarchy of website. It is helpful for fast and easy navigation through the website. According to some webmasters it is useful for in-bound linking of web-pages which alternatively gets some points from search engines in ranking. Use of images in breadcrumb helps in earning extra points, instead of text home hyperlink a home image is more suitable as it can further contain alt text, explained above,  thus treated as anchor text ().

**l) Sitemap:** Generally sitemaps can be further divided to two types: a) HTML page listing which is used for easy navigation for web-users. These are divided into sections and categories depending on the size and pages a website contains.

b) XML Sitemap- It was introduced by Google for dynamic web-pages. It helps crawlers to crawl webpages developed in Flash or AJAX ((Odom and Lynell, 2012 pp.86-90). It is more precise than normal HTML pages as syntax errors are not tolerated.

Sample coding for XML Sitemap:

```
<?xml version="1.0" encoding="UTF-8"?>
<urlset     xmlns="http://www.sitemaps.org/schemas/sitemap/0.9">
         //current protocol

   <url>        //parent tag for url entry
     <loc>http://www.mysite.com/</loc>    //Page url
        <lastmod>2012-02-21</lastmod> // Last modification date
   <changefreq>weekly</changefreq> //frequency of web-page changing( hourly,weekly,
         monthly, never)
   <priority>1.0</priority> // priority of this page in website value0.0-1.0 default-0.5
                                   </url>
                                   </urlset>
```

### 2.4.3.3 Off-page optimization

Those strategies which are used outside the webpage and are not directly related to modification on page content are known as off-page optimization techniques. It is more related to page-linking techniques and Social media marketing. They are mostly used to maximize keyword performance of websites (). Some popular methods are Blog Commenting, posting on forums, bookmarking on social sites, article submission and RSS feed submission. It is a lengthy and continuous process. Anchor text linking is most popular as in this, webmasters use words to hyperlink their websites.  For example, a computer shop website may be linked to a blog discussing computer studies.

```
<a href="http://mycomputerwebsite.com">Computer Studies</a>
```

 This research mainly focuses on-page optimization techniques as off-page optimization techniques are often misused and are widely used by spammers.

## 2.5  Web Mining

### 2.5.1  Overview

Traditional data mining techniques applied to gather information over the World Wide Web is known as Web mining. It can be categorized into three parts: Web content mining, Web usage mining and Web structure mining (Stumme *et al.,* 2006).

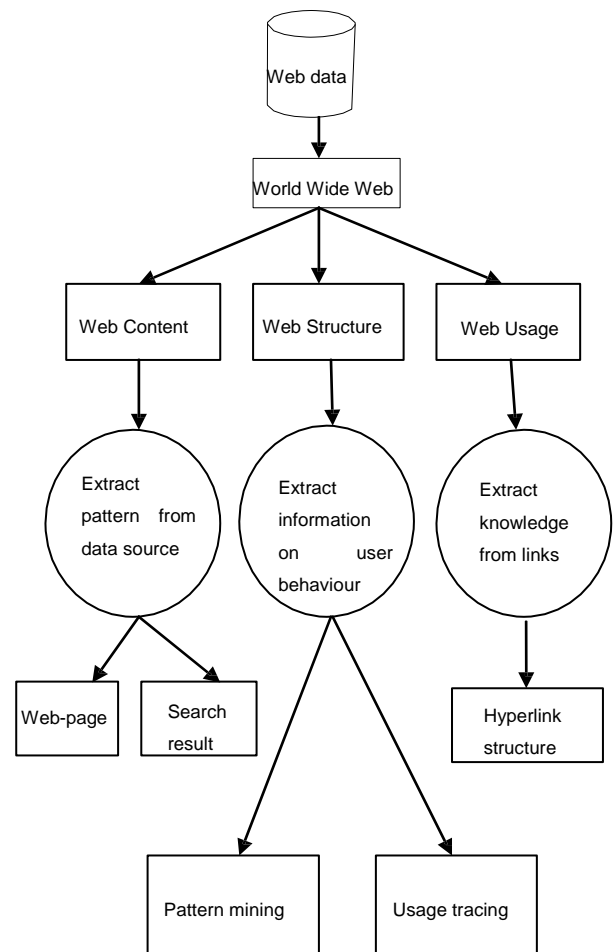

**Figure 2: Web Mining categories and objects**

#### 2.5.1.1  Web Content mining

It deals with mining, extraction and integration of information, knowledge and data (multimedia, audio, video, text and hypertext) from web page contents (Liu, 2005). A wrapper is used for extraction of data using machine learning methods, trained in manually labelled pages used as training set. Stalker (Muslea *et al.*, 1999), WL2 (Cohen *et al.*, 2002) and Thresher (Hogue and Karger, 2005) are some popular research based wrapper induction systems.

#### 2.5.1.2  Web Structure mining

It is the process of getting knowledge about hyperlinks on web-page. It is useful in generating patterns or deriving relationship between web-pages (Torres *et al.,* 2010). Web-page content can be organized, depending on XML and HTML, in tree structure (Sharma and Bhartiya, 2012).

### 2.5.1.3  Web Usage mining

It is the study of records by visitors on any website, mostly stored in web server log (Arotaritei and Mishra, 2004). Various tools such as Google Analytics are used for data log extraction (Devi *et al.,* 2012). Carmona *et al.,* (2012) studies web usage mining techniques such as clustering, subgrouping and association rules for improving website design for an e-commerce website (OrOliveSur.com).

## 2.5.2  Related Work

As mentioned in (Munibalaji and Balamurgan, 2012) web is large pool of data and finding relevant information is difficult, web mining helps in extracting useful pattern and develops abstract from diversified sources. Intelligent web agents, Harvest (Brown *et al.,* 1994), Information Manifold (Kirk *et al.,* 1995) and ParaSite (Spertus, 1997) uses pre-specified domain information for retrieving and interpreting documents. Whereas, Internet Learning Agent (Perkowitz and Etaioni, 1995) uses content mining for information retrieval and populates its own hierarchy.

Page Rank Algorithm (Brin and Page, 1998) and weighted page rank algorithm (Haveliwala, 2003) use web structure mining for ranking the page. A new approach, Web page content ranking algorithm (Munibalaji and Balamurgan, 2012) uses web structure mining for page importance (via. back-linking) and page relevancy is calculated using web content mining techniques. There detailed study is available at Appendix [4].

## 2.6  Summary

There has been little research done on search engine optimization, while the market has grown to a multi-billion dollar business (Berman and Zsolt, 2011). Most of the earlier researches are based on the economic aspects of SEO rather than the science behind using these techniques for usefulness to search engines and users. In this chapter, various search engine optimization techniques was highlighted. A detailed study on evolution of search engines and Meta search engines were discussed. The accuracy of search results is measured by relevancy of query term to web-pages ranked and served by search engines. Even a small query term generates thousands of pages, which if not sorted can be difficult to retrieve relevant information. Search engines work hard on structuring and ranking these search results. Different methods and approaches for retrieving and ranking these search results have been discussed above. Search engine optimization techniques are used by websites for providing a better indexing to search engines. This chapter can be concluded on the note of making a Meta search engine to provide a structure for using SEO techniques to rank web-sites and study there relevancy to the content and user satisfaction.

## 3   Practical Work Undertaken

## 3.1 Overview:

The literature review highlighted some major search engine optimization methods and some recent research by (Gupta and Aggarwal, 2012; Berman and Katona, 2011; Chen *et al.,* 2011; Chandra *et al.,* 2011; Buha, 2010; Evans, 2007) done in this context. World Wide Web is an ocean of information which people use every day to find relevant information. Lately due to continuous expanding size and complexity of web, retrieving useful information, effectively and efficiently has become a major concern. Various techniques have been proposed both by theoretical researchers and Company researchers. Developing Meta search engines and using different algorithms for information retrieval (Mahabhashyam and Singitham, 2004; M and Jacob, 2008) to variations in page ranking algorithms (Joachims, 2002; Bifet *et al.,* 2005; Modi *et al.,* 2011) has been discussed in chapter above. This chapter proposed a new Meta Search engine model, *iral*, retrieving information from current search engines (Google, Yahoo, Bing and MSN) and ranking the relevant pages using Naïve Bayesian Classifier based on training sets. Training sets are based on Search engine optimizing (SEO) parameters.

## 3.2 Architecture of Meta search engine

**3.2.1  General Architecture**: In this section a high level overview of how system works is given. In order to elicit data and provide a framework for testing page ranking algorithm based on SEO techniques, researcher has implemented a WWW Meta search engine called '*iral*'.  Meta search engines (MSE) are tools used for retrieving relevant information from the web by combing results of various other search engines without having a database of own (Joachims, 2002).

As mentioned in (Dreilinger and Howe, 1997) Meta search engines can be broken into three components:

1) Dispatch Component- In this user can provide a decision making algorithm for selecting a particular Meta search engine for relevant query. This research is devoted to results page sorting instead analyzing this component.

2) Interface Agents- This is used for interacting with search engines by passing user's query to them. They are also responsible for matching the query format to search engines format.

3) Display Components- Results from various search engines need sorting before been displayed to users. These sorting can include different page ranking techniques, removing duplicates and verifying broken links. The experimental portion of this research is based on this component.

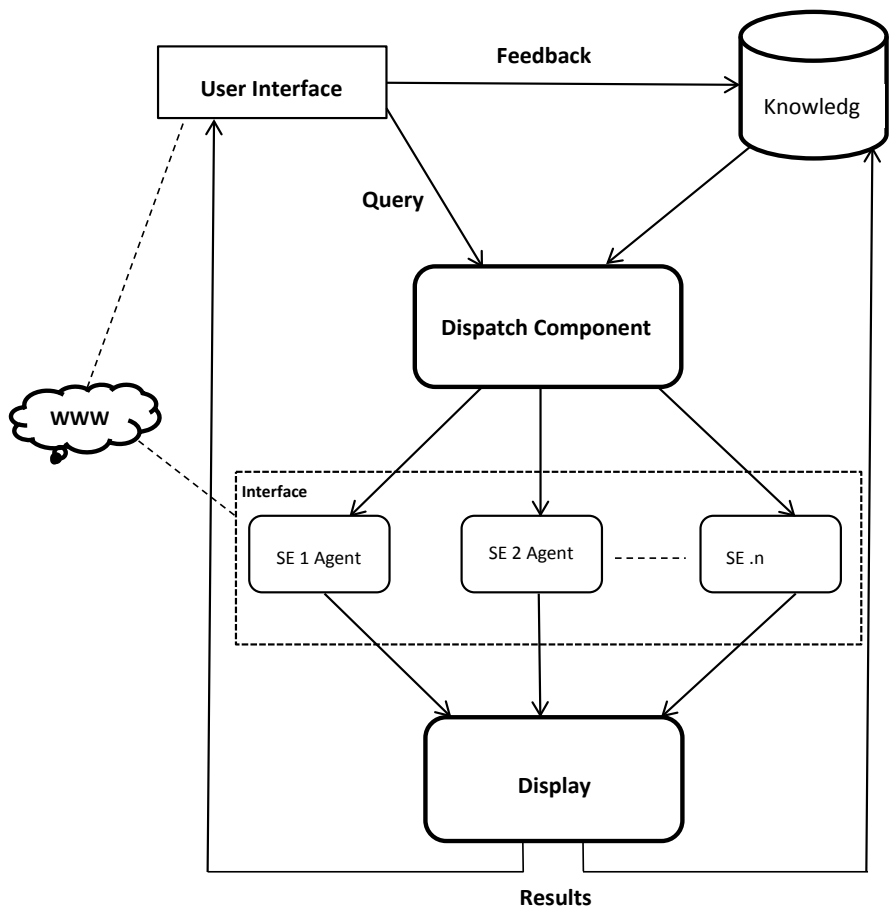

**Figure 3: General framework for categorizing of Meta search engine**

### 3.2.2 Proposed iral Architecture

In this section architecture of iral (Meta search engine) is explained. iral accepts a single query from the user and sends it to multiple search engines. This architecture supports two search engines: Google and Bing.

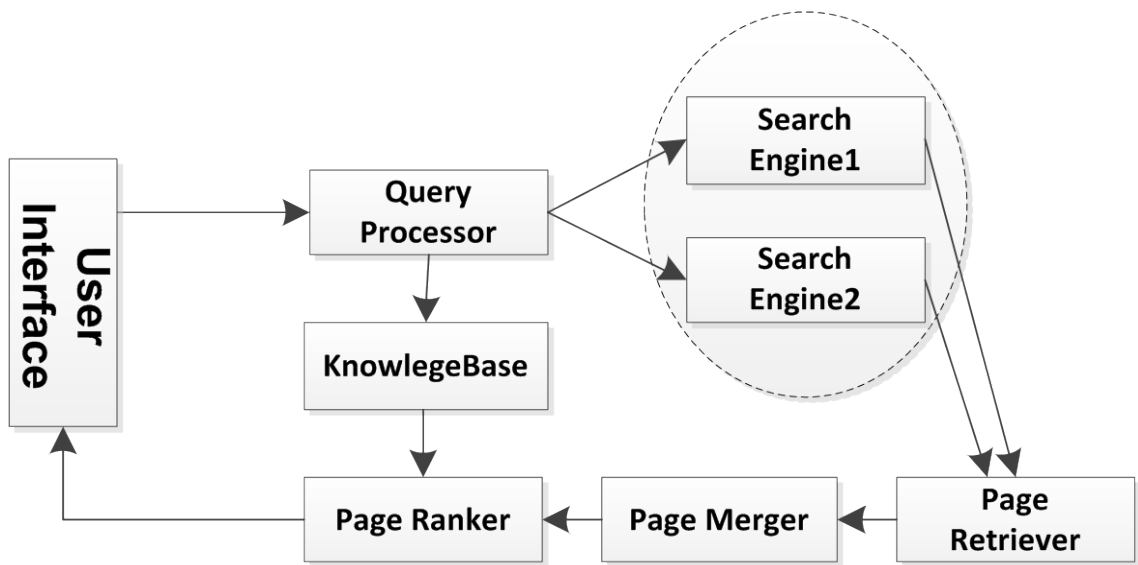

Figure 4: Proposed iral Meta search engine architecture

The user interface allows the user to submit their queries. The query processor will pass the query to search engines and knowledge base. The search engines will process the query and the results generated from them will be retrieved by page retriever. As two search engines can have similar search results for same query so to eliminate duplicates all the retrieved pages will be send to page merger. Duplicate webpages will be removed by assigning hash value to every page and sorting them in hash table. Links which produces same hash value will be considered as duplicates and removed. The merge results without any duplicates will send to page ranker where pages will be ranked. A page ranking algorithm will give heuristic page rank to merged pages using SEO parameters.

### 3.1.1.1 User Interface

An information system is connected to data sources which contain large collection of data. To access the data economically and effectively, data must be easily reachable to the users (Wellhausen, 2006). User interface can be described as a junction between user and a computer program. User Interface (UI) in basic can be of two types: a) command driven (where user enters command) and (b) menu-driven (where user selects commands from menu). Search engines employ a search text box for users to enter their query.

In present system, User Interface allows user to submit their search queries. The UI deployed here is a search box which supports text based searches. User enters the keyword or combination of keywords for any search query. This query will then be forwarded to query processor by using a Hypertext Pre-processor scripting language.

### 3.1.1.2 Query Processor

The aim of query processor is to accept user query/keyword and to modify it for best utilizing the search engines. User's query for any desired results are not expressed alone by keywords (Glover *et al.*, 1999 and Souldatos *et al.*, 2005). Use of query processor is used for explicit notation of user preferences. Query processor is connected to search engines via API (application programming interface) provided by Google and Bing. Other end of query processor is connected to a knowledge base.

### 3.1.1.3 Search Engines

The use of search engines is vital in retrieving results for queries in Meta search engines. Here the researcher has used two search engines namely, Google and Bing[7]. The query process send the keyword to search engines, API's maintains all the required setting for the communication with remote search engines. Once the keywords are entered in both search engines, they dispatch the results in HTML format known as Search result records (SRRs).

---

[7] Bing is a search provider to Yahoo and MSN.

### 3.1.1.4 Page Retriever

The use of page retriever is to collect all the results from search engines individually. The Search Return Result (SRR) is dynamically generated web usually enwrapped in HTML tags. It contains useful information such as web-link, domain name, meta-tag, title description and snippet.

### 3.1.1.5 Page Merger

Page merger is an important application which is used for merging the pages form different search engines. The aligned pages from page retriever is send to page merger where it remove any duplicate pages such as same website link from two different search engine.

### 3.1.1.6 Knowledge Base

The use of knowledge base is to get user query from query processor and find the synonyms of one work query. As there is no database for the system, knowledge base works by sending the query to dictionary.com via its API. The system then gets the results and send it to page ranker where it search for the related keywords in meta description, snippet and title tags.

### 3.1.1.7 Page Ranker

The purpose of page ranker is to order web-pages according to the ranking algorithm. It use search engine optimization parameters for ranking web-page. The detailed list of parameters and ranking algorithm is described in later section.

## 3.3  Implementation and Design of Meta search engine-iral

This section demonstrates insight of designing and implementation of proposed iral Meta search engine. It presents various Unified Modelling Language (UML) diagrams of iral Meta search engine.

UML can be defined as a standardized language for writing software blueprints. It is widely used to visualize, construct, understand and documentation of software artefacts (Chiang, 2009). Right set of UML diagrams are provided for visualization of Meta search engine in this chapter. For system development file see Appendix [1]

After system architecture is been discussed, next logical step is to present the design of the system. This chapter also includes functional and non-functional requirements.

### 3.3.1 Functional Requirements

Functional requirements can be described as 'what system should do?'. Given below are list of functionalities this system is capable to do.

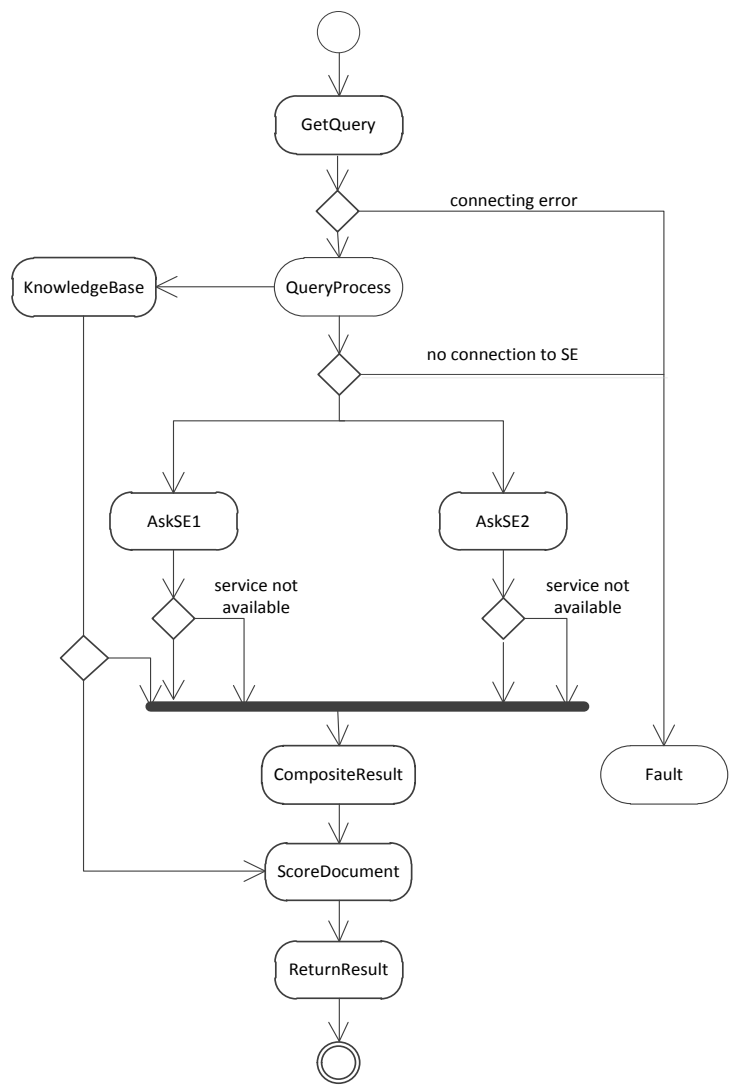

- **GetQuery:**
  User feeds the query/keyword. If hitting the search button produces no effect, a connection error is displayed. Or else query is forwarded to query processor.

- **QueryProcess:**
  User keyword is redirected to Search Engine and Knowledge Base. If HTTP connection is not set up, it changed to fault state.

**Figure 5: Activity diagram of iral Meta search Web service**

- **KnowledgeBase:** It processes the search query, no result if tail term detected. Or send result to score document.

- **ASKSE**
  MSE create thread for both search engines, if service not available/ available, in both state goes to composite result.

- **CompositeResult**
  All retrieved results from search engines are merged and duplicates are removed. Send new list of results to ScoreDocument.

- **ScoreDocument**
  New list of merged results are ranked according to set parameters.

- **ReturnResult**
  Ranked search results are forwarded to the user for review.

- **Fault**
  The comprehending error message is returned.

### 3.3.2 Detail level design

This section includes the Use case diagram and Sequence diagram for iral Meta search engine.

#### 3.3.2.1 Use Case diagram

**Input Query:** User input search query. It can be text, integers or special characters.

**Query Processing:** System will check the query and provide a connection to external search engines.

**Search Results:** System process the query and get result from external search engines. It will organize and display results to user.

**Browse Results:** User will browse through the results provided by system.

**Select Result:** User will select the best matching result from the system.

**Open Webpage:** System will connect to web domain and open webpage for result selected by the user.

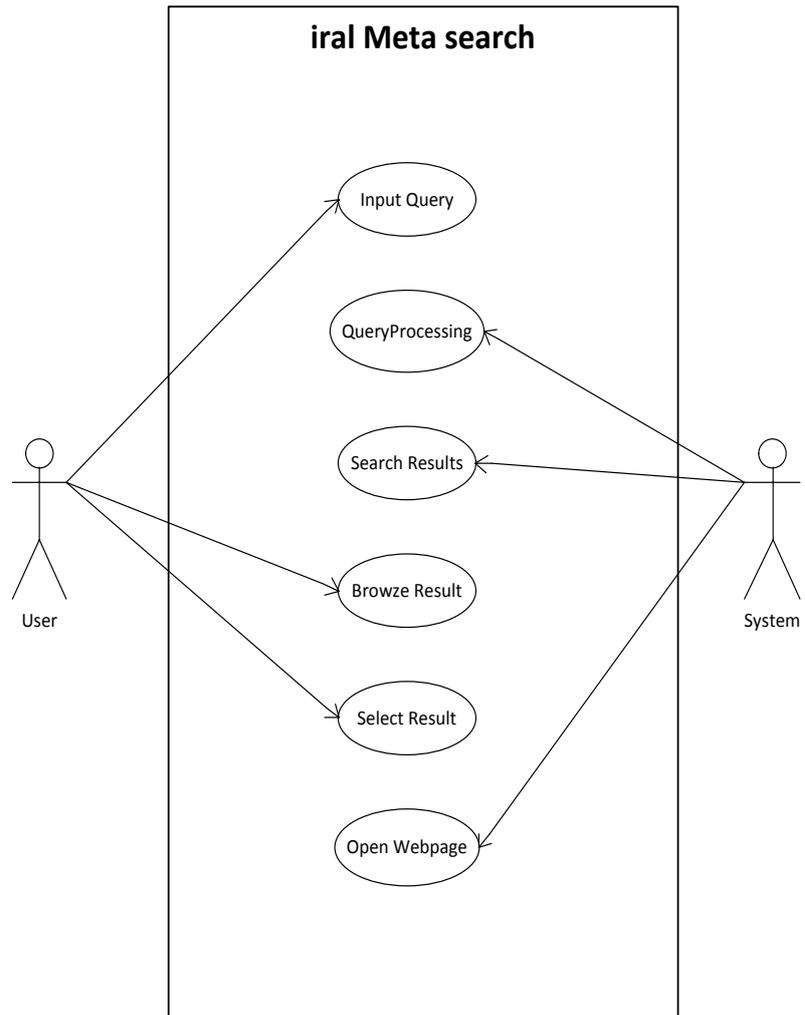

Figure 6: Use case Diagram- iral Meta search engine

### 3.3.2.2 Sequence diagram

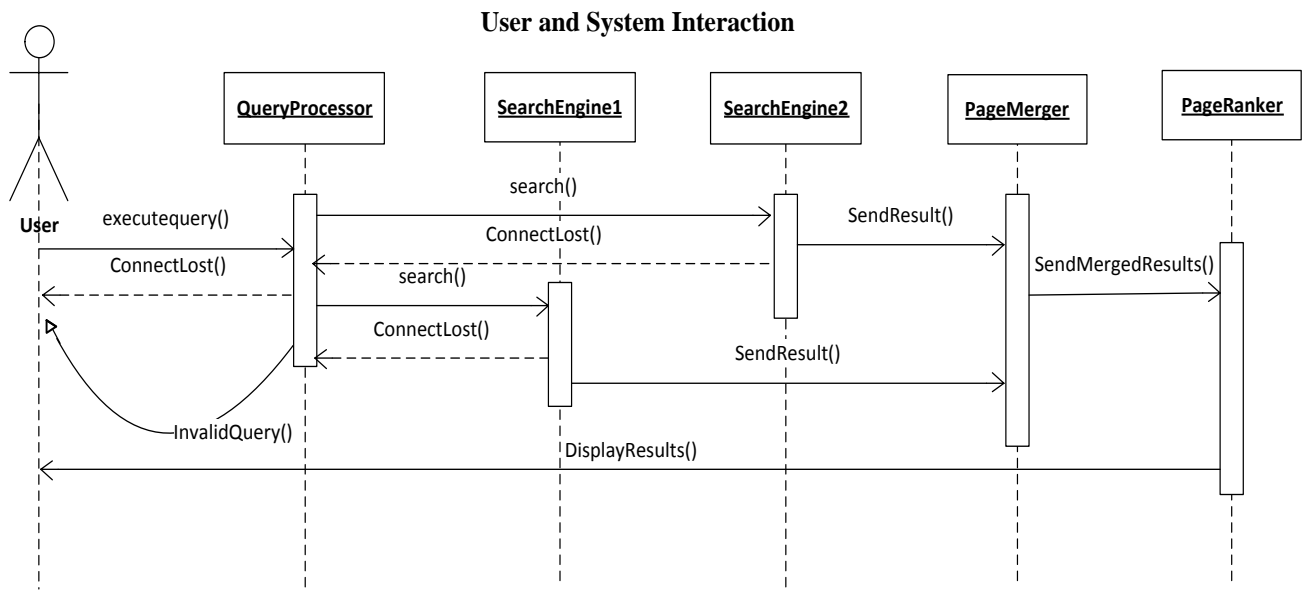

**User and System Interaction**

## 3.3.3 Non-Functional Requirements

Non-Functional requirements of system are 'how perfectly system's functional requirements are fulfilled. ' For users, some characteristics of system can make an effect of 'how well' system has performed.

The system is a web-based application which has a GUI for users. From users perceptive requirements of the system are:

**Usage:** The Meta search engine's GUI is based on same provided by Google. As Google is most used search engine, so users will be experienced in using it.

**Robustness:** The system is robust as it perform task easily without any failure.

**Precision:** System aims to generate the best results for any given query.

**Recall time:** The recall timing of this system is not been calculate precisely but is decent.

**Performance:** System is developed in Hypertext pre-processor language. It has to functionalities as can be easily modified and consumes less bandwidth hence increasing performance.

**Reusability:** System is easy to modify since been developed in php. The system is reusable and can be used for further developments.

### 3.3.4  Data Source

Meta search engines works by combining results from various search engines. Selection of search engines is based on current search engine market share. According to survey conducted by (Net Market Share, 2012) Google is most preferred search engine with a market of 82% searches worldwide. The current implementation of iral supports following search engines: Google and Bing.

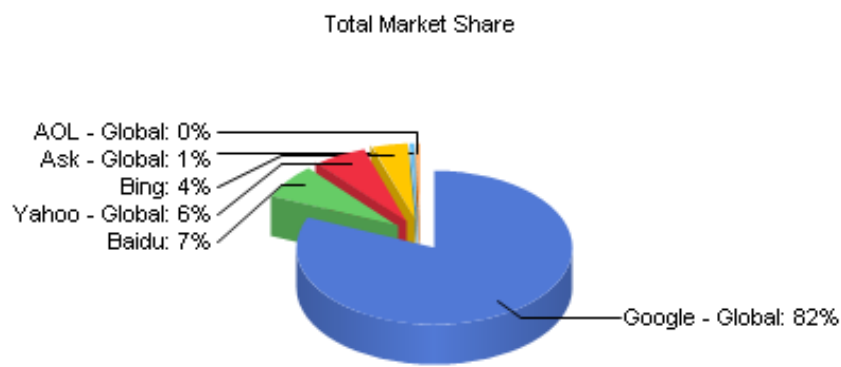

Figure 8: Search Engine Global Market Share 2012

For generating the results for search query, researcher has implemented Google and Bing API into the system. In earlier model of Meta search engine, Yahoo search and MSN was included. This has to be removed from the system because Bing is now the official search provider for Yahoo and MSN, generating same results for given query. The use of API's is important as this research focuses on information embedded in HTML tags of web-pages which can easily be gathered from SRR's.

### 3.3.4.1  Google Search API

Using Google API helps software developers to write their own programs to query the web. The use of Google SOAP Search API is discontinued to use. Instead the researcher has used Google Web Search API. This API is useful for generating results from Google search engine. Number of research has shown the use of Google API in their Met search engine. This API is available in various languages and platform like- Java, Perl, PHP. Only limitation of this API is that unregistered/ free users can search 100 queries per day. Anything above that terminates the connection. More details and code are in Appendix [2]

### 3.3.4.2  Bing Search API

Bing is now the biggest search provider after Google[8]. Since Bing is now the official search provider to Yahoo and MSN, this research hence uses Bing search results. Bing search SDK provides a platform to send query to its search engine and retrieve the information. The received information can be modified and displayed according to the need of developer. A brief use and code is available at Appendix [2]

---

[8] http://www.forbes.com/sites/greatspeculations/2012/02/15/bing-yahoo-lose-net-u-s-search-market-share-in-january/

### 3.3.5 Alternate Design Approach

#### 3.3.5.1 Approach 1:

**Overview:** This research alternatively can use web based crawlers which extract information from websites. These applications can be helpful for extracting HTML content of webpages. As mentioned in (Shetty *et al.,* 2012) HTML content can be converted into XML and parsed with Xpath[9]. New web based crawlers are efficient for capturing specified data (Agarwal *et al.*, 2012).

**Design:** The web crawler can be programmed to extract HTML tag information from the web pages. Aim of this research is find important SEO parameters for search optimization; crawler can be built to extract information from web-page based on these parameters. Mercator (Heydon and Najork, 1999) is a web crawler used for web mining, whose parameters can be easily altered for extracting information. Various researchers (Shkapenyuk, 2002; Boldi *et al.,* 2004; Castillo, 2005; Edwards *et al.,* 2007) has developed and used crawler based on their requirements and hypothesis.

**Limitations:**

**Time-** Using web crawler limits the scope of study. World Wide Web is large pool of knowledge and indexing even a single part of it requires time. In related experiment conducted by (Brin and Page, 1998) it took 9 days for indexing 26 million pages at an average 48.5 pages/second and using 3 crawlers at one time. Developing and training of web crawler take long time.

**Indexing-** For indexing this large amount of data, major data structure is needed. Problem like disk seek and is common in large databases. Normalization of data units is must as crawler can crawl same content twice. Sorting and removing duplicate is another major concern.

---

[9] A type of query language to select nodes from XML documents.

**Spam-** Web crawlers often index spam web-pages which results in duplicate unwanted data. Indexing of this data is both hard and time consuming. In order to remove this problem the use of external search engines in this research is necessary.

### 3.3.5.2 Approach 2

**Overview:** Using specialized search engines such as Google for general query, Torrentntz for files and Sweetsearch for Educational can help in better search result analysis. The present system is good for text based search but may obtain video or image result due to the use of Google and Bing API. On the other hand, dynamic engine selector can be developed for selecting search engines of particular type (Modi *et al.*, 2011). This approach helps to better understand the query and SEO effects on web-pages.

**Design:** Meta crawler can include a dynamic search selector for processing search query. Meta search engines like Helios (Gulli, 2005) and Tadpole (Mahabhashyam and Singitham, 2004) used dynamic search selector. In present model a dynamic search selector can be integrated after query processor, to send query to appropriate search engine.

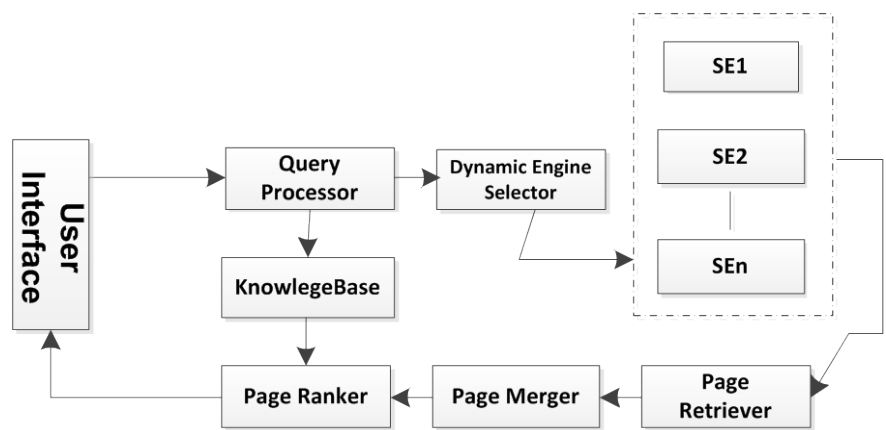

Figure 9: Alternate diagram for Meta search engine

**Limitation:** Using dynamic engine selector requires training particular set of data which requires additional time. The classification of text can be done by using Naïve Bayesian Classifier Algorithm.

### 3.3.6 Ranking Strategy

Ranking of web-pages generated from search engines is an important factor for popularity of search engines. There have been various researches in this field, categorizing important data elements for ranking web-pages. In PageRank[10] (Brin and Page, 1998) web-page ranking is based on its link structure and link text. Other noted ranking methodologies for Meta search engines are Broda's Positional Method (Fagin *et al.*, 2004), Weighted Broada-Fuse (Dorn and Naz, 2008), Broda count method (Akritidis, Katsaros and Bozanis, 2008), TopD (Lu *et al.,* 2005) and SRR Rank (Dwork *et al.,* 2001). The details of these ranking parameters and ranking strategies are given in Appendix [3]. In this section PageRank algorithm and proposed page ranking algorithm for iral Meta search engine is described.

### 3.3.6.1 PageRank

The concept of PageRank (Brin and Page, 1998) in Google has enormous effects on web optimization techniques and the way webpages are constructed for better ranking. According to (Golliher, 2008) PageRank lies in assumption that link in the web page is an on topic link and unbiased link to other website. Web spammers have used several methods to manipulate the ranking algorithms, like using; 'link farm', blog spam, paid page linking and massive link exchanges. These assumptions in PageRank highly reduce the effectiveness of result page ranking.

Basic page rank formula is given by:

$$R(B) = \delta / n + (1-\delta) \sum_{hyperlink(A,B)} R(A) / outlinks(A)$$

In this equation, δ is set to 0.85 (Boldi, no date), A and B are different webpages, n is number of pages collected in the calculation and out links of A are hyperlinks directing toward page A. From this equation value page rank of B is determined by total number of page collected (more pages collected will decrease page rank), and webpage like A will reduced its pagerank by number of pages it refer (Golliher, 2008). From recent studies and news it's obvious that search engines keep their algorithms safe and update them frequently. Algorithms are now more sophisticated and improved from the past. A detailed ranking and verification is presented in Appendix [3]

---

[10] Trademark of Google

### 3.3.6.2 Modified iral Ranking Method

The ranking parameter of page ranking algorithm is modified according to need of this experiment. Using combinations of ranking metrics, algorithms explained above and combing them with other variables, set of new parameter are laid for ranking web-pages. These new parameters for ranking are given below:

For every parameter same weightage is assigned. Important search engine optimizing (SEO) factors [11] are considered and assigned the weight.

These parameters have same weight in order to judge their importance in other search engines. Every search engine has its own parameters and ranking strategies, if we have to learn the important parameters for ranking we must dig all the existing and known parameters.

| SEO Parameters | Description |
|---|---|
| Title Tag | If query is a term in webpage title |
| Meta Description | If query term is present in description |
| Meta Keyword | If query matches keyword |
| Snippet | Number of times word appears in summary |
| Meta expires | Recent pages to be given higher rank |
| Meta content | help to show compatible pages to users (UTF-8, ASCII or ISO) |
| Image attribute | use of Image alt attribute helps in higher weightage as it provides description to images |
| Sitemap | use of sitemap increase visibility for dynamic web-pages |
| Links present | number of inbound links to the page |

**Table 1: Parameters for ranking**

Ranking should be based on the SRRs retrieved from search engines. The SRR contains useful information like-page title, snippet, meta-contents, keywords etc.

---

[11] Only white hat techniques are used for

The ranking is calculated as below:

In this equation: $v_i$ is the value of parameter

$$PageRank(P) = \sum v_i(P) * w_i(v_i)$$

$w_i$ is weight of the parameter $v_i$

**Assumptions:** Intuitively, a more reasonable search engine will retrieve better results for given query.

### 3.3.7  Design Constrains

#### 3.3.7.1  Hardware requirements

| Component | Minimum Requirement | |
|-----------|---------|--------|
| | **Client** | **Server** |
| Processor | 64 bit | 64 bit |
| Hard Disk | 8MB | 500MB |
| RAM | 512MB | 1GB |

Table 2: Hardware requirements for Meta search engine

#### 3.3.7.2  Software requirements

**Client**

- Browser: Firefox, Google Chrome, Internet Explorer 5.5 and later, Safari[12]

**Server**

- Apache Tomcat Server
- Java Servlet Development Kit (JDSK 2.0)
- Operating system
- PHP version 5.2.3 or later
- MySQL database 4.0 or later

### 3.3.8  Deployment Details

iral Meta search engine is been deployed  for general public use and gathering data on one.com servers.

---

[12] For Macintosh Operating System

## 3.4  Testing

### 3.4.1  Overview

This section gives the useful insight in testing iral Meta search engine. Software testing is a regress process designed to approve software's purpose of development. Software has to be predictable and reliable before handing over to users (Glenford *et al.,* 2011 p.3). Software testing has been an important stage in SDLC (software development life cycle), but is more important today due to complex software designed for broader audiences (Glenford *et al.,* 2011 p.5).    There are various techniques and tools for debugging and testing software depending on their programming language and development environment. There are basically three methods of testing described in (Glenford *et al.,* 2011 p.52) White Box testing, Black Box testing and Grey box testing.

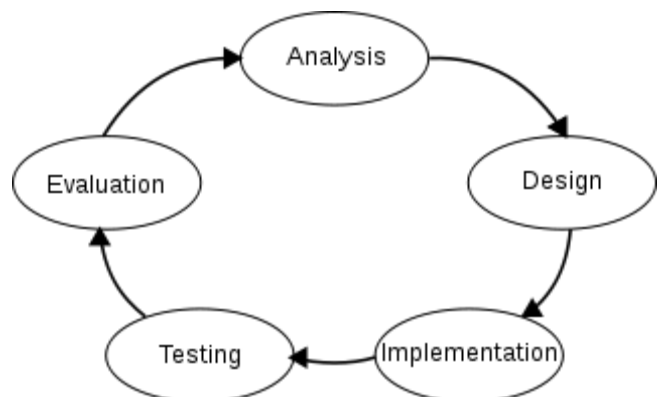

Figure 10: Software Development Life Cycle

White Box testing methods are applied while testing iral Meta search engine. As mentioned in (Glenford *et al.,* 2011 pp.52-54) white box testing is performed when tester has access to code, data structure and algorithm of the system.

### 3.4.2  Test Cases

Test cases are scripts or set of conditions or variables assigned by tester to test and validate working of software programme as designed to do so (Glenford *et al.,* 2011 pp.72-77). Finding and designing proper test cases are an important factor in testing. In order to carry out testing, researcher has used three approaches namely, code validation, unit testing and Graphical user interface testing.

### 3.4.2.1 Code Test

Present system is a web based application developed and programmed in Hypertext Preprocessor (PHP). Testing is done by validating source code of the system. There are two promising methods for checking validity of source code: Web based methods and System software methods.

### *a) System Software Methods:*

There are numerous software applications which generate validation for scripts written in different programming language. The most common applications for validating php scripts are given below:

- **Adobe Dreamweaver**

  Most of the coding and development of software is done using Adobe Dreamweaver CS5.5 Version 11.5 Build 5344. The build in W3C markup validator is used as standard for validating web documents written using markup languages.

- **PHP Unit Test**

  It is popular de-facto standard for testing PHP developments. PHPUnit 3.6 is used for testing which requires PHP 5.4.0 or later. For installing PHPUnit, PHP Extension and Application Repository (PEAR) package for code coverage is used.

Some important test cases results are in next page:

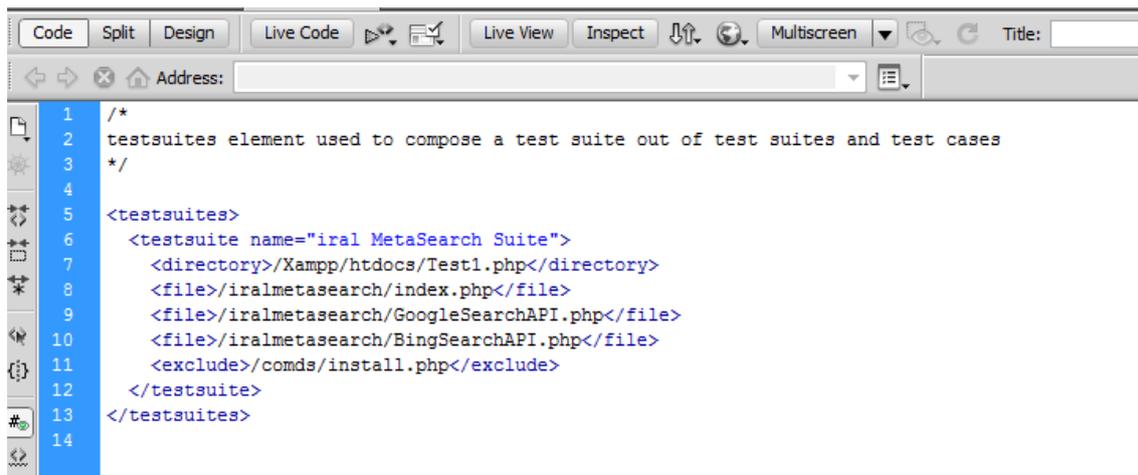

**Figure 11: Test Case**

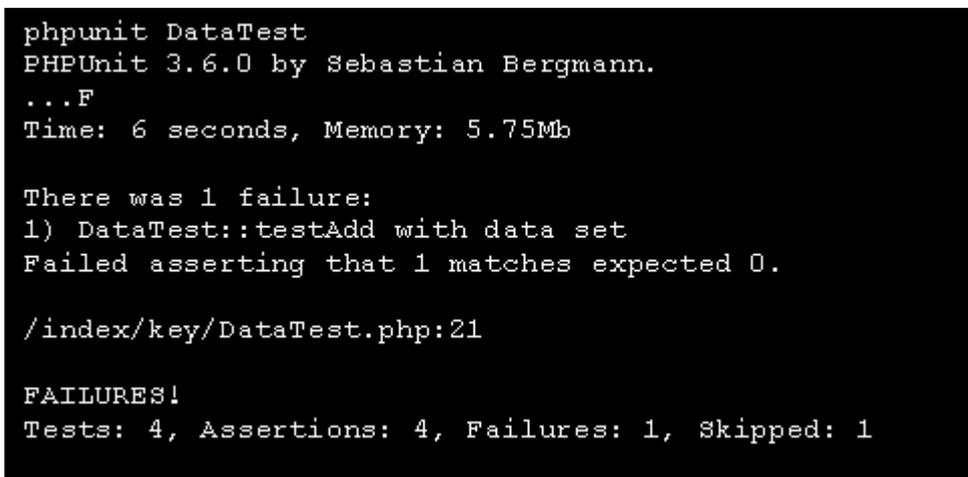

**Figure 12: Data Test case report**

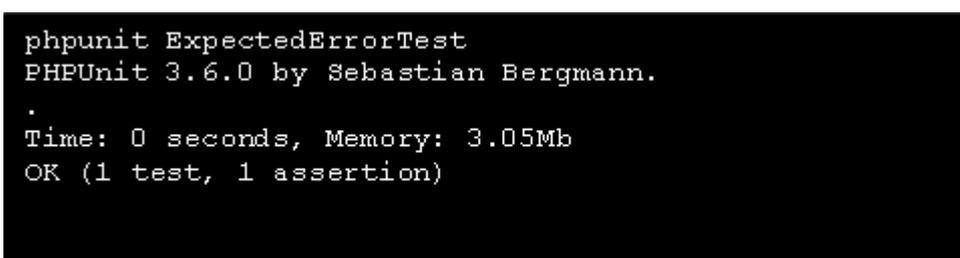

**Figure 14: Expected error test case report**

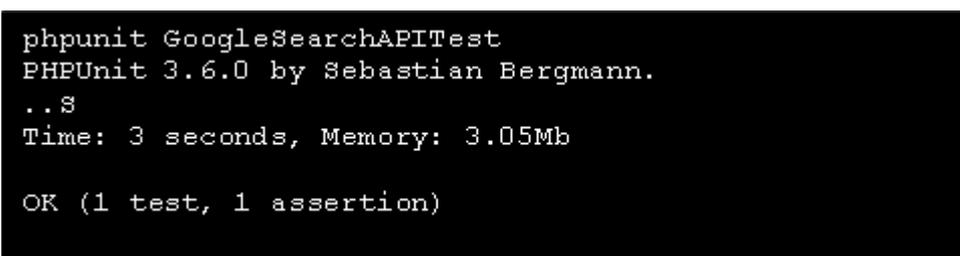

**Figure 13: Google Search API test case report**

```
phpunit BingSearchAPITest
PHPUnit 3.6.0 by Sebastian Bergmann.
..S
Time: 3 seconds, Memory: 3.05Mb

OK (1 test, 1 assertion)
```

**Figure 15: Bing Search API test case report**

*b)* *Online Validation*

Online validation of the system is done by using validator available at W3C website.

| | | | |
|---|---|---|---|
| Errors found while checking this document as HTML 4.01 Transitional! | | | |
| Result : | 4 Errors, 5 warning(s) | | |
| Address : | http://iralmetasearch.com/search.php | | |
| Modified : | (undefined) | | |
| Server : | Apache | | |
| Size : | 254 | | |
| Content-Type : | text/html | | |
| Encoding : | utf-8 | (detect automatically) | |
| Doctype : | HTML 4.01 Transitional | (detect automatically) | |
| Root Element : | div | | |

## 3.4.2.2  GUI Test

Testing GUI is an important testing process to ensure products Graphical User Interface meets its programming specification. Main elements to test in GUI are listed next page:

- *Test 1: Search Button*

| Test Case: Search Button | |
|---|---|
| **Test case ID** | **1.1** |
| Test case Name | Blank Search |
| Test case Description | Check text button response with no value in text box |
| Test case Procedure | Manually reordered testing |
| Expected Result | No Response |
| Actual Result | No Response |
| Remarks | Pass |

Table 3: Search Button test case

| Test Case: Search Button | |
|---|---|
| **Test case ID** | **1.2** |
| Test case Name | Filled Search |
| Test case Description | Check text button response with value in text box |
| Test case Procedure | Manually reordered testing |
| Expected Result | Display result in same page |
| Actual Result | Display result in same page |
| Remarks | Pass |

Table 4: Filled search test case report

- *Test 2: Search Text Box*

| Test Case: Search Text box | |
|---|---|
| **Test case ID** | **2.1** |
| Test case Name | Keyword Test |
| Test case Description | Check maximum length of words in text box |
| Test Data | Numeric and Alphabets |
| Test case Procedure | Manually reordered testing |
| Expected Result | Unlimited |
| Actual Result | Unlimited |
| Remarks | Pass |

Table 5: Keyword test case report

| Test Case: Search Text box | |
|---|---|
| **Test case ID** | **2.2** |
| Test case Name | Special character text |
| Test case Description | Check if text box can input special character set |
| Test Data | @ £ $ % & * ( ) < > ? / ; ' '' # |
| Test case Procedure | Manually reordered testing |
| Expected Result | Can Input special characters |
| Actual Result | Can Input special characters |
| Remarks | Pass |

Table 6: Special character test case report

### 3.4.2.3  Load Test

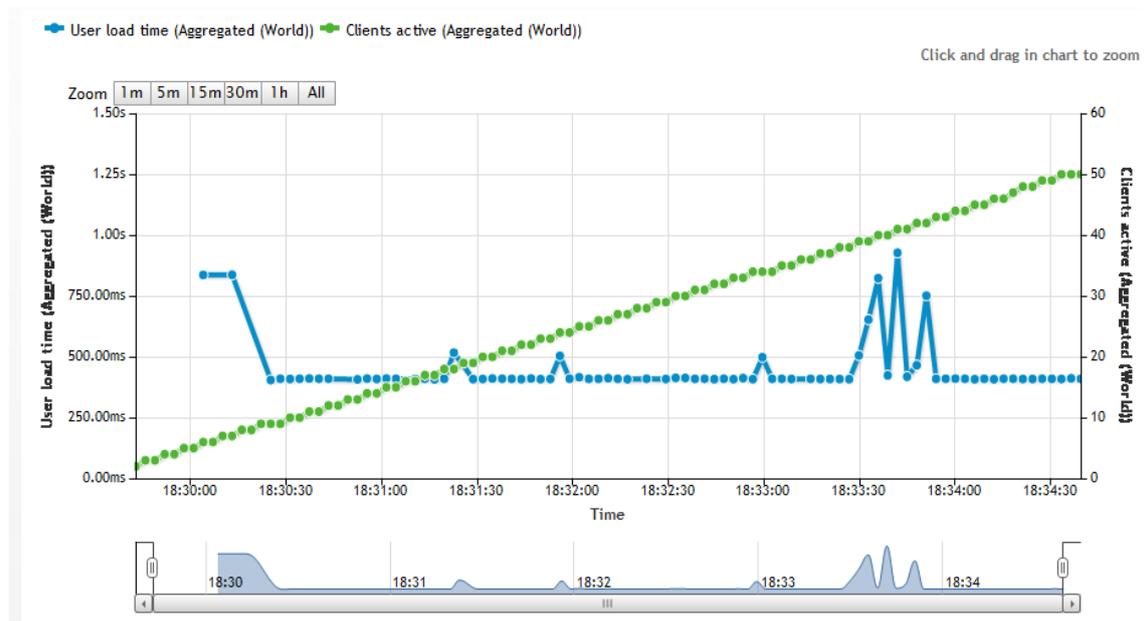

### 3.4.3  Summary

Testing is done to find errors in the programme in executable mode. Testing of search engines is a difficult and time consuming job. The above white hat testing techniques have validated that system is operational and bugs reported were checked and cleared. Validation of written code is done by using online and offline methods. PHP Test Unit 3.6.0 is used for creating and testing test cases. For validation of code W3C validator is used online and Adobe Dreamweaver CS5.5 Version 11.5 offline. GUI testing is performed manually as tools are expensive and hard to find. Load testing is done using web based software, Pingdom which is widely used by testers.  Due to time constrain only important test cases were designed but overall system performance in testing has qualified it for user interaction.

# 4   Results, Analysis and Evaluation

This chapter contains results for experiments carried out to find performance for the product created. The Meta search engine, iral, is created for better optimizing search results generated from search engines Google and Bing. This search results were ranked according to SEO parameters. There are three parameters on which current system is analysed and evaluated. First section uses web mining techniques to find SEO parameters used by search engines in indexing and ranking webpages. Second section is to calculate performance of iral Meta search engine which optimizes search results (by using SEO parameters) of Google and Bing search engines. Performance is calculated in terms of precession and recall timing and results based on user's interaction with the system.

## 4.1   SEO Techniques

Objective of this research focuses on highlighting best optimizing techniques for website. These SEO techniques are important for search engines to index and categorize web-pages. As mentioned in (Gupta and Aggarwal, 2012) there has been not many research done in finding those SEO techniques which are used by search engines for ranking web-pages. Although in Google (Larry and Page, 1998) techniques used for categorizing web-pages is based on title of web-page, but due to expansion of WWW and presence of spammers has changed the way search engine categorize web-pages (Yu *et al.,* 2012).

In order to find important SEO techniques, results generated from above experiments were analysed and evaluated by web mining methods explained in literature review. For full access to web-mining documents see Appendix [4]

### 4.1.1   Results

The below list is used to determine and relate SEO techniques are useful for search engine crawler. This result shows the importance of SEO techniques and there weighting parameters for ranking. Evaluation of only 9 SERPS[13] is done for this research due to limited time. Wikipedia is ranked first in every search engine as it's considered as web knowledgebase, hence excluded from this study.

| | | | | | | | | | | | |
|---|---|---|---|---|---|---|---|---|---|---|---|
| **GOOGLE RESULT 'Alcoholism'** | | | | | | | | | | | |
| **Ra nk** | **Website** | **Key word in Title tag** | **Meta descri ption** | **Meta keyw ord** | **Snip pet** | **Met a Expi res** | **Met a cont ent** | **Imag e Attri bute** | **Site map** | **Lin ks pres ent** |
| 1 | alcoholism.about.com | yes | 13 | 203 | 6 | yes | 29 | 9 | yes | 142 |
| 2 | alcoholism.about.com/od/about/a/symptoms.htm | yes | 11 | 338 | 3 | yes | 23 | 18 | yes | 280 |
| 3 | patient.co.uk/health/Alcoholism-and-Problem-Drinking.htm | yes | 6 | 178 | 5 | yes | 21 | 18 | yes | 280 |
| 4 | blackwellpublishing.com/journal.asp?ref=0145-6008 | yes | 9 | 212 | 3 | yes | 31 | 4 | yes | 74 |
| 5 | netdoctor.co.uk/health_advice/facts/alcoholism.htm | yes | 3 | 123 | 3 | yes | 25 | 9 | yes | 201 |
| 6 | http://www.medicinenet.com/alcohol_abuse_and_alcoholism/article.htm | yes | 10 | 150 | 6 | yes | 31 | 18 | yes | 370 |
| 7 | alcoholics-anonymous.org.uk/newcomers/?PageID=69 | yes | 5 | 63 | 3 | yes | 16 | 3 | yes | 26 |
| 8 | tandf.co.uk/journals/titles/07347324.asp | yes | 7 | 110 | 2 | yes | 12 | 10 | yes | 60 |
| 9 | nhs.uk/news/2012/03march/Pages/lsd-acid-alcoholism-treatment.aspx | yes | 5 | 79 | 4 | yes | 10 | 19 | yes | 209 |

Table 7: SEO Parameters present in Google SERP 'Alcoholism'

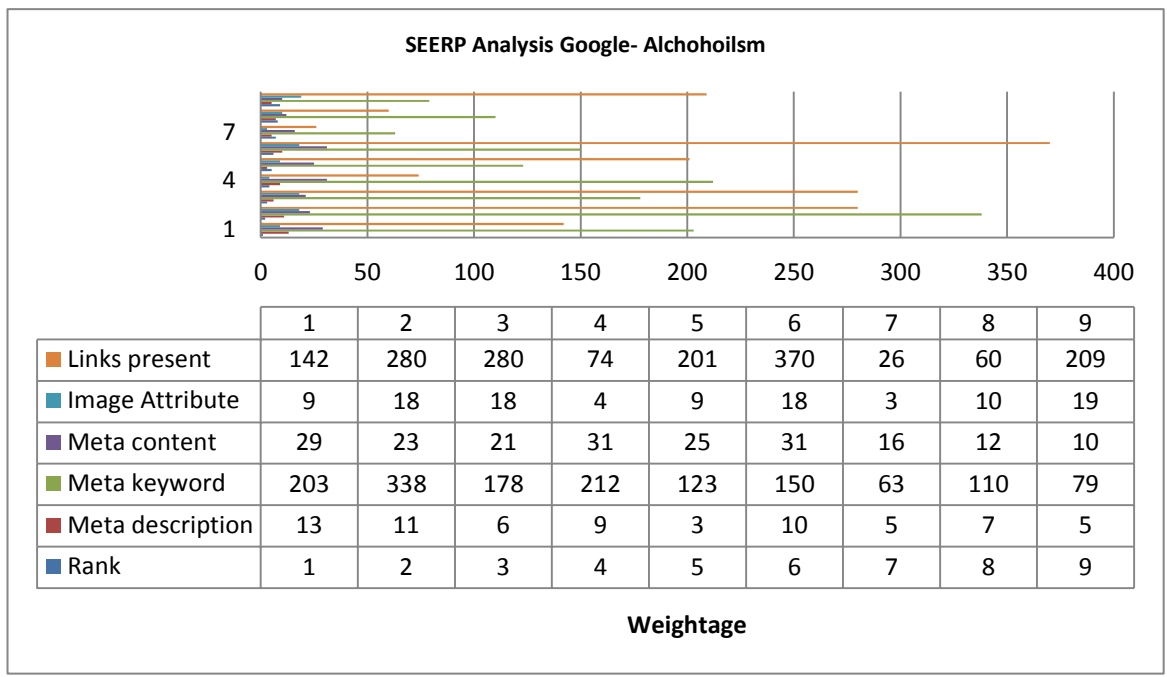

Figure 16: SEERP Analysis Google- Alcoholism

---

[13] Search engine Result Pages

| BING RESULT 'Alcoholism' | | | | | | | | | | |
|---|---|---|---|---|---|---|---|---|---|---|
| Ra nk | Website | Keyw ord inTitl e tag | Meta descrip tion | Meta keyw ord | Snip pet | Meta Expi res | Meta cont ent | Image Attrib ute | Sitem ap | Link s pres ent |
| 1 | alcoholism.about.com | yes | 13 | 203 | 6 | yes | 29 | 9 | yes | 142 |
| 2 | medical-dictionary.thefreedictionary.com/alc oholism | yes | 7 | 112 | 4 | yes | 21 | 15 | yes | 213 |
| 3 | alcoholism.about.com/od/about/a/sy mptoms.htm | yes | 11 | 338 | 3 | yes | 23 | 18 | yes | 143 |
| 4 | ncbi.nlm.nih.gov/pubmedhealth/PM H0001940/ | yes | 21 | 122 | 2 | yes | 12 | 2 | yes | 146 |
| 5 | nlm.nih.gov/medlineplus/alcoholism .html | yes | 23 | 211 | 1 | yes | 21 | 3 | yes | 128 |
| 6 | encyclopedia2.thefreedictionary.co m/alcoholism | yes | 15 | 200 | 1 | yes | 32 | 7 | yes | 21 |
| 7 | adam.about.net/reports/Alcoholism. htm | yes | 9 | 142 | 5 | yes | 21 | 12 | yes | 123 |
| 8 | netdoctor.co.uk/health_advice/facts/ alcoholism.htm | yes | 12 | 132 | 1 | yes | 13 | 2 | yes | 21 |
| 9 | patient.co.uk/health/Alcoholism-and-Problem-Drinking.htm | yes | 6 | 178 | 5 | yes | 21 | 18 | yes | 280 |

**Table 8: Parameters present in Bing SERP 'Alcoholism'**

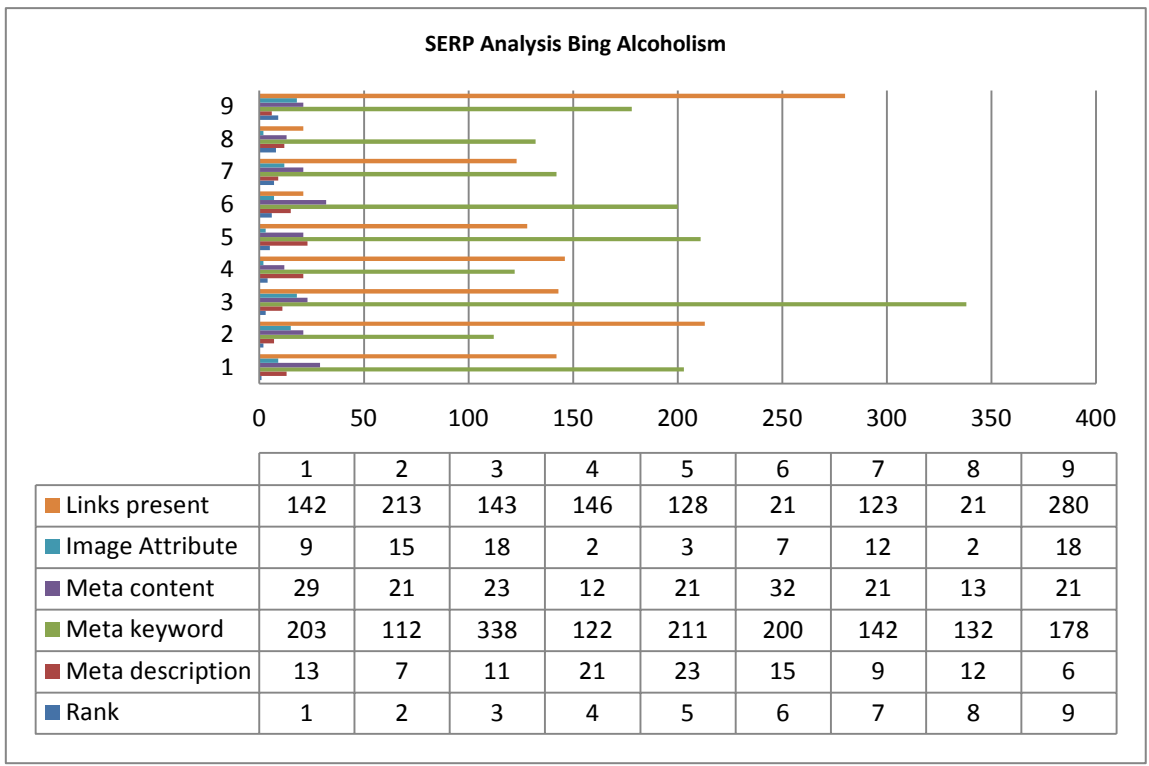

**Figure 17: SEERP Analysis Bing- Alcoholism**

| GOOGLE RESULT 'local computer shop' | | | | | | | | | | | |
|---|---|---|---|---|---|---|---|---|---|---|---|
| **Rank** | **Website** | **Keyword inTitle tag** | **Meta description** | **Meta keyword** | **Snippet** | **Meta Expires** | **Meta content** | **Image Attribute** | **Site map** | **Links present** |
| 1 | http://uk.local.yahoo.com/United_Kingdom/Computer_Shops/uk100000082-s-23424975.html | yes | 21 | 5 | 2 | yes | 12 | 3 | yes | 170 |
| 2 | pzcomputers.com/pzc/index.html | yes | 12 | 4 | 1 | yes | 10 | 4 | yes | 78 |
| 3 | heapsofpcs.com | yes | 11 | 4 | 1 | yes | 12 | 0 | yes | 7 |
| 4 | tricomputers.co.uk/Index.php | yes | 12 | 3 | 2 | yes | 5 | 0 | yes | 7 |
| 5 | which.co.uk/technology/computing/guides/computer-repair-top-tips/ | yes | 10 | 3 | 1 | yes | 9 | 0 | yes | 12 |
| 6 | pcadvisor.co.uk/forums/2/consumerwatch/125700/local-computer-shop-and-innocent-customer/ | yes | 10 | 3 | 2 | yes | 2 | 0 | yes | 79 |
| 7 | localpcs.co.uk/ | yes | 9 | 3 | 1 | yes | 6 | 0 | yes | 322 |
| 8 | yourlocalcomputerguy.co.uk/ | yes | 23 | 3 | 1 | yes | 2 | 2 | yes | 20 |
| 9 | www.abiko.co.uk/ | yes | 11 | 2 | 1 | yes | 20 | 1 | yes | 66 |
| 10 | mumsnet.com/Talk/geeky_stuff/591841-so-my-local-computer-shop-has-my-laptop-for-repair/AllOnOnePage | yes | 1 | 2 | 2 | yes | 14 | 21 | yes | 241 |

**Table 9: Parameters present in Google SERP 'local computer shop'**

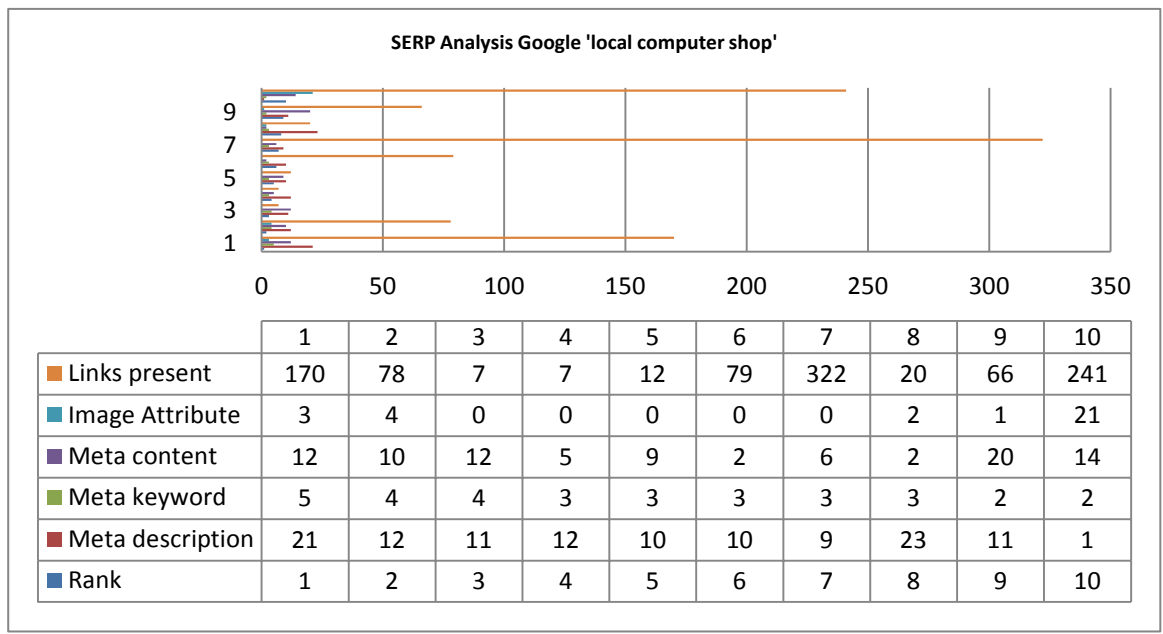

**Figure 18: SEERP Analysis Google- local computer shop**

| BING RESULT 'local computer shop' | | | | | | | | | | | |
|---|---|---|---|---|---|---|---|---|---|---|---|
| Rank | Website | Keyword inTitle tag | Meta description | Meta keyword | Snippet | Meta Expires | Meta content | Image Attribute | Sitemap | Links present |
| 1 | http://uk.local.yahoo.com/United_Kingdom/Computer_Shops/uk100000082-s-23424975.html | yes | 21 | 5 | 2 | yes | 12 | 3 | yes | 170 |
| 2 | http://www.localpcs.co.uk/ | yes | 15 | 7 | 1 | yes | 11 | 2 | yes | 1 |
| 3 | http://yourlocalcomputerstore.com/ | yes | 20 | 4 | 2 | yes | 4 | 1 | yes | 1 |
| 4 | http://www.mylocalcomputershop.com/ | yes | 11 | 3 | 1 | yes | 5 | 1 | yes | 5 |
| 5 | pzcomputers.com/pzc/index.html | yes | 16 | 5 | 3 | yes | 11 | 1 | yes | 7 |
| 6 | ipatter.com/anglianinternet/local-computer-shops-in-norfolk-13808 | yes | 32 | 5 | 2 | yes | 4 | 3 | yes | 142 |
| 7 | javea-computer-club.wikidot.com/suppliers | yes | 11 | 3 | 1 | yes | 9 | 1 | yes | 153 |
| 8 | local.yahoo.com/info-33045872-local-computer-shop-grapevine | yes | 21 | 2 | 2 | yes | 14 | 1 | yes | 67 |
| 9 | dtechcomputers.co.uk/ | yes | 12 | 1 | 1 | yes | 2 | 1 | yes | 7 |
| 10 | accessplace.com/computer-system/west-midlands/dudley.htm | yes | 17 | 1 | 1 | yes | 12 | 1 | yes | 78 |

Table 10: Parameters present in Bing SERP 'local computer shop'

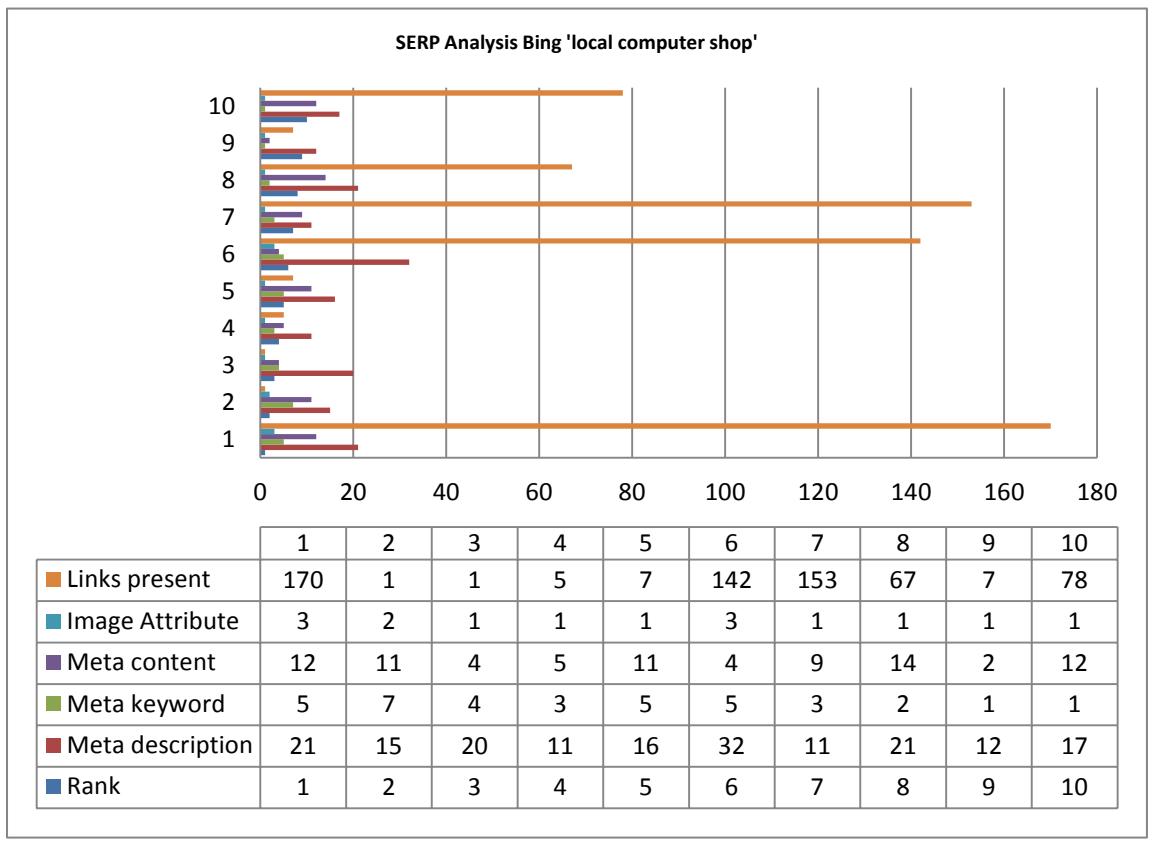

Figure 19: SEERP Analysis Bing- local computer shop

## 4.1.2  SEO SERP Evaluation

From above results, researcher was able to find relevant SEO techniques which help websites to rank higher in search engines result page. The below data is useful for webmaster providing SEO consultancy to website owners.

From literature review it was concluded that Google prioritize link building for ranking it result pages. Above experiments proved this theory and highlighted other SEO parameters and their effects in ranking.

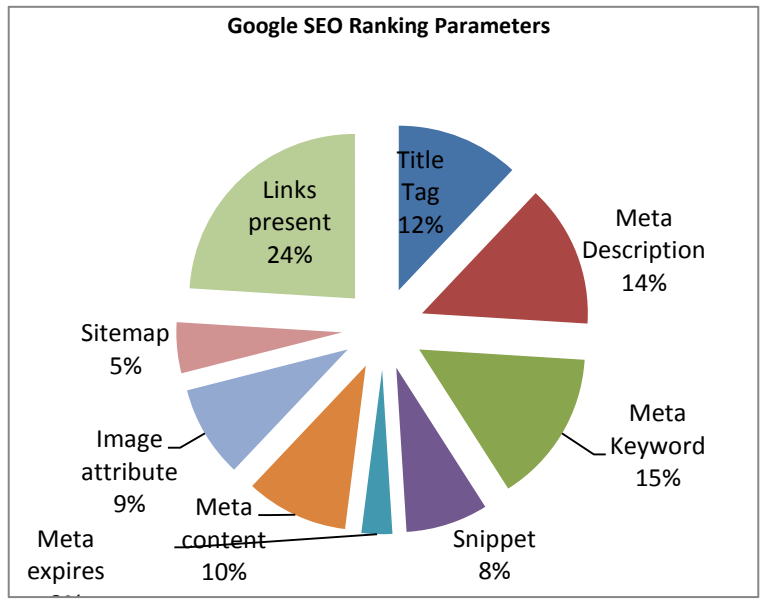

Figure 20: Google SEO Ranking Parameters

Experiments conducted shows that Bing gives highest preference to number of inbound-links present in website. Meta keyword is another important factor to be considered while optimizing website. Other factors and percentage in ranking are given in figure.

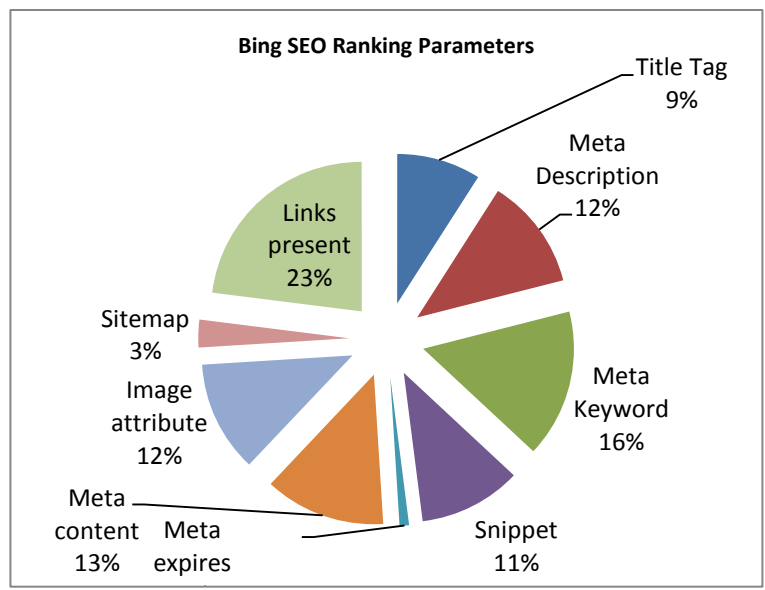

Figure 21: Bing SEO Ranking Parameters

Chart below is comparing Bing and Google SEO parameters and their contribution in ranking web-pages.

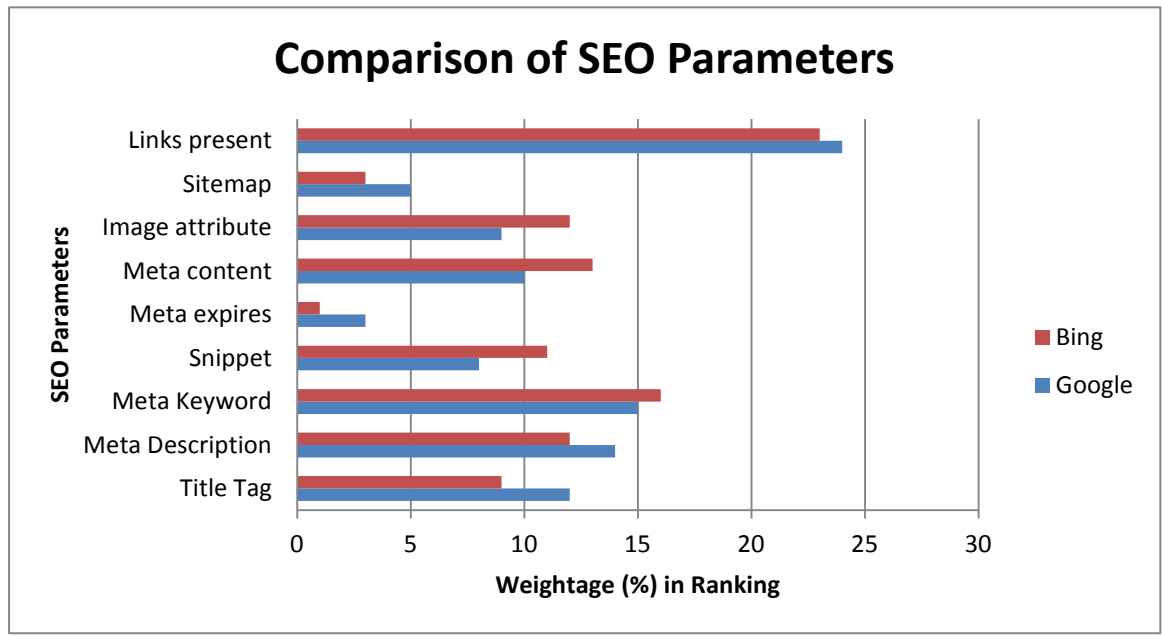

### 4.1.3  Summary

There are different types of Search engine optimization techniques; Black-hat and White-hat, discussed in literature review. For search engines, white-hat techniques are those which enhance web-pages quality, content and user satisfaction. Not much research is conducted to find these techniques and there significance in ranking. In this research important white-hat techniques are highlighted. These techniques will help webmasters and website developers for better optimizing their website. Google and Bing are two major search providers with different set of algorithms, rankling strategy, indexing and categorizing. For same query they produce different results. The above results has compared there SEO ranking parameters and draws conclusion that Bing and Google both uses SEO parameters for categorizing but weightage they assign to these factors are different. Hence webmasters should use above mentioned SEO factors which are important in both search engines.

## 4.2  iral Meta Search Engine

### 4.2.1  Overview

Experiment was carried out to find performance of search engines namely: Google, Bing and iral Meta search engine. These experiments are limited due to vast number of search results generated from each search engine. For any given query, search engine retrieve millions of pages. Evaluating and analysing them is a difficult and time consuming job. In order to carry out experiment, researcher thus used and compared first 50 results from every search engine. Extensive web mining techniques are used for evaluating results. List of retrieved documents and results are given in Appendix [4]

### 4.2.2  Precession and Recall

In evaluating performance of search engines, precession time and recall time are important factors to be considered. In earlier research (Dreilinger and Howe, 1997; Meng *et al.*, 2002; Mohamed, 2004; Chen and Liu, 2005; Moghaddam and Parirokh, 2006; Zheng 2010; Ganzha *et al.*, 2010; Jadidoleslamy, 2011) uses precision and recall times for evaluating performance of their Meta search engines. Precision of search results are more important than recall as the results from search engines are far more than a user can look at.

***Precession***

As suggested in Meng *et al.*, (2002) precision is a factor in information retrieval used to decide the quality of results produce by search engines.  It is given by;

$$precision = \frac{no.\,of\ relevant\ documents\ retrived}{total\ no.\,of\ retrived\ documents}$$

Below are search engines performance table and figures:

| Google Precision Rate | | | | | | |
|---|---|---|---|---|---|---|
| Search Query | Total no. of sites retrieved | No. of sites evaluated | More Relevant | Less Relevant | Irrelevant | Precision |
| Alcoholism | 34,100,000 | 50 | 22 | 19 | 9 | 0.44 |
| local computer shop | 266,000,000 | 50 | 19 | 16 | 15 | 0.40 |

Table 11: Google Precision Rate

| Bing Precision Rate | | | | | | |
|---|---|---|---|---|---|---|
| Search Query | Total no. of sites retrieved | No. of sites evaluated | More Relevant | Less Relevant | Irrelevant | Precision |
| Alcoholism | 33,100,000 | 50 | 16 | 22 | 12 | 0.31 |
| local computer shop | 304,000,000 | 50 | 12 | 23 | 15 | 0.24 |

Table 12: Bing Precision Rate

| iral Meta search engine Precision Rate | | | | | | |
|---|---|---|---|---|---|---|
| Search Query | Total no. of sites retrieved | No. of sites evaluated | More Relevant | Less Relevant | Irrelevant | Precision |
| Alcoholism | 50 | 50 | 24 | 20 | 6 | 0.48 |
| local computer shop | 50 | 50 | 18 | 19 | 13 | 0.37 |

Table 13: iral Meta search Precision Rate

| Search Engine | Avg. Precision for Alcoholism |
|---|---|
| Google | 0.44 |
| Bing | 0.31 |
| iral Metasearch | 0.48 |

Table 14: Avg. Precision for 'Alcohol'

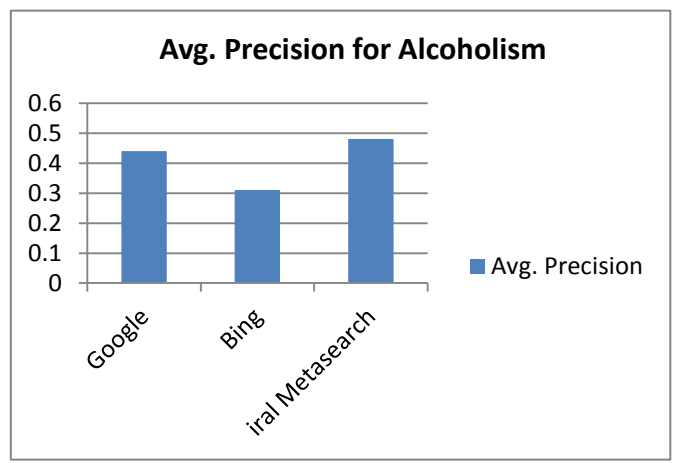

Figure 23: Avg. Precision for 'Alcohol'

| Search Engine | Avg. Precision for local computer shop |
|---|---|
| Google | 0.40 |
| Bing | 0.24 |
| iral Metasearch | 0.37 |

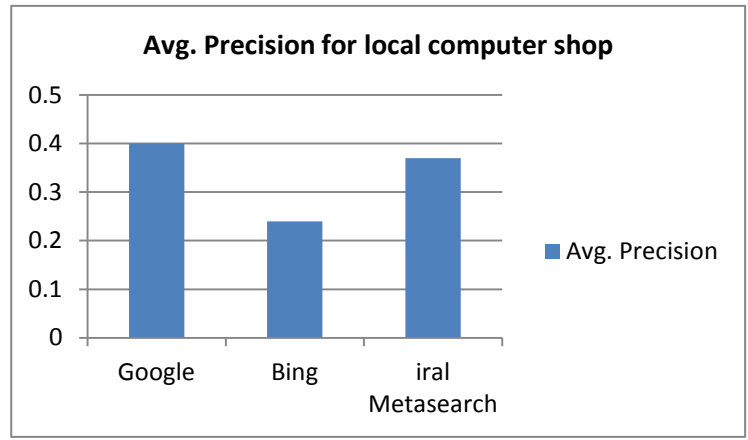

Table 15: Avg. Precision for 'local computer shop'

Figure 25: Avg. Precision for 'local computer shop'

| Search Engine | Mean Precision |
|---|---|
| Google | 0.42 |
| Bing | 0.28 |
| iral Meta search | 0.43 |

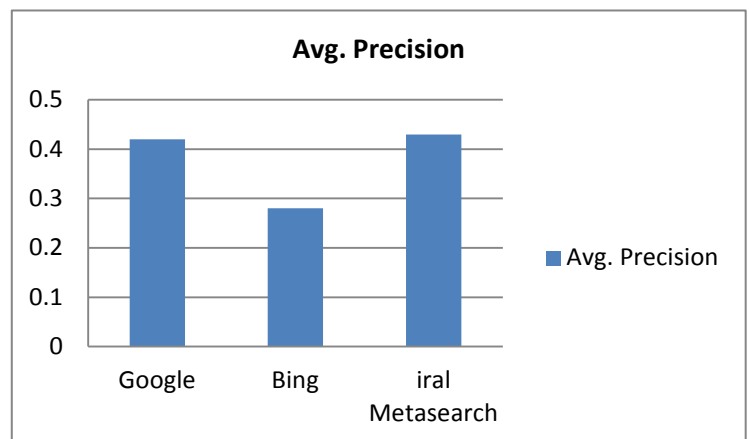

Table 16: Mean Precision of Search engines

Figure 24: Mean Precision of search engines

***Relative Recall***

According to (Meng *et al.*, 2002) relative recall is calculated for evaluating the dominating search engine among all those search engines used in research. Formula for calculating recall is given by;

$$\text{relative recall} = \frac{Total\ no.\ of\ relevant\ documents\ retrived\ by\ search\ engine}{total\ no.\ of\ relevant\ documents\ retrived\ by\ all\ search\ engine}$$

| Search Query | Google | | Bing | | iral meta search engine | |
|---|---|---|---|---|---|---|
| | Total no. of sites | Relative Recall | Total no. of sites | Relative Recall | Total no. of sites | Relative Recall |
| Alcoholism | 34,100,000 | 0.51 | 33,100,000 | 0.49 | 50 | 0.00000074 |
| local computer shop | 266,000,000 | 0.466667 | 304,000,000 | 0.533333 | 50 | 0.000000008 |

**Table 17: Relative recall for Search Engines**

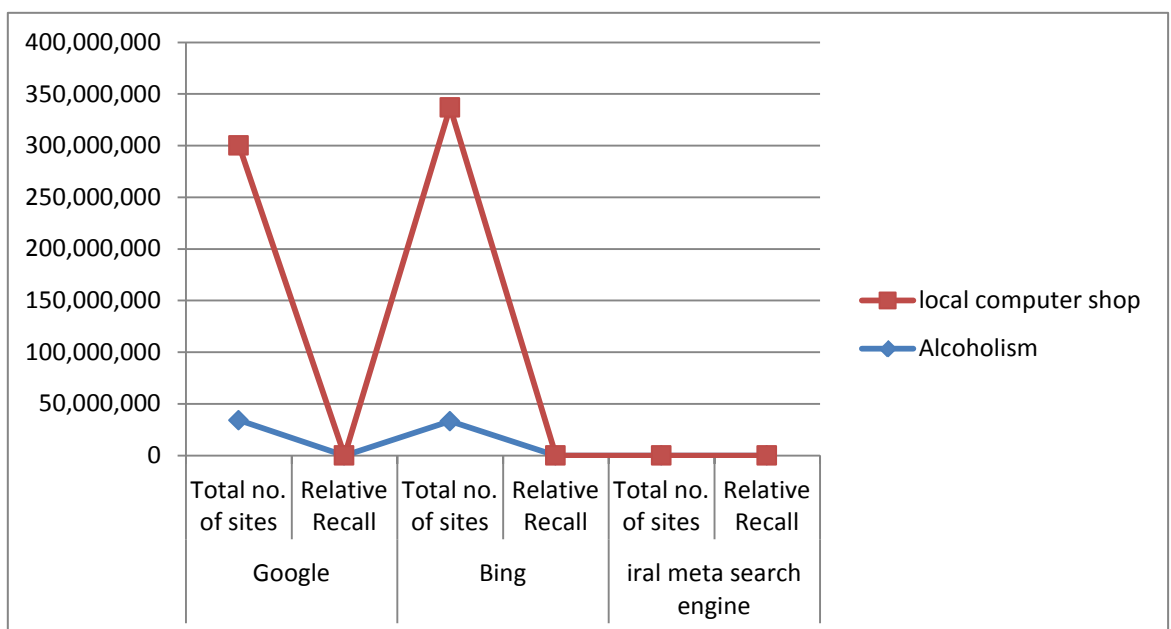

**Figure 26: Relative recall for Search Engines**

### 4.2.3 Manual Pick Performance

This experiment was carried out to find relevancy of results generated from iral Meta search engine for given query. Two queries were considered due to limitation of time. To see the effects of web-optimization techniques in search results and their relevance to the user query, the following online experiment was performed. First batch of experiment was performed using 2 types of keywords: head keywords (they consist of single word) and tail keywords (they are longer in length 3 or more words). Keywords used are: '*alcoholism*' and '*local computer shop*'. This batch included 50 students randomly selected from Northumbria University library. They were asked to use below keywords and rate search engines in a scale of 5, where 1 is the least preferred and 5 is most preferred. This forms a quantitative approach for evaluating search engine performance. Researcher like (Mockus and Weiss, 2001) argued that software is developed for general users and their response to system shows actual success of it.

| Search Engines | Google | Bing | iral |
|---|---|---|---|
| Keyword | Alcoholism | Alcoholism | Alcoholism |
| No. of pages retrieved | 34,100,000 | 33,100,000 | 50 |
| Time to process | 0.23 sec | n/a | n/a |
| Pages User Evaluated | 10 | 10 | 10 |
| User rating | 4 | 3 | 4.5 |

**Table 18: Comparison data for Alcoholism**

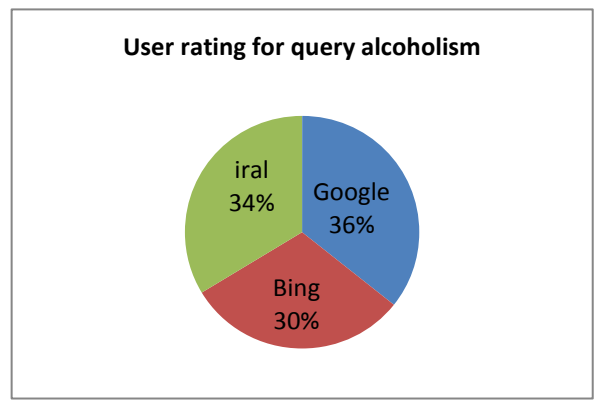

**Figure 27: User rating for Alcoholism**

| Search Engines | Google | Bing | iral |
|---|---|---|---|
| Keyword | local computer shop | local computer shop | local computer shop |
| No. of pages retrieved | 266,000,000 | 304,000,000 | 50 |
| Time to process | 0.25 sec | n/a | n/a |
| Pages User Evaluated | 10 | 10 | 10 |
| User rating | 3.5 | 3 | 3.3 |

**Table 19: Comparison data for local computer shop**

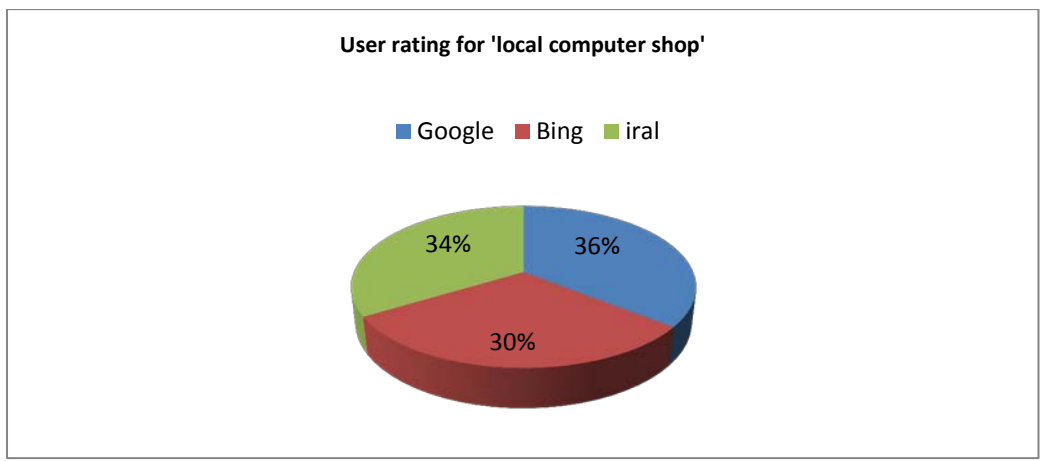

### 4.2.4  Summary

In experiment conducted above, there were promising results from iral Meta search engine. For one word query iral Meta search outperformed Google and Bing. Results retrieved from iral were more related to given query than Google and Bing search engines. In tail term query, Google's performance was better than other two search engine.  This is due to the fact that iral Meta search engine retrieve results from Google and Bing. If precession rate of Bing falls it has adverse effect on iral Meta search performance. Recall of iral is low due to number of results it retrieve is limited to 50. From user's perceptive, iral Meta search has outperformed Google and Bing in retrieving relevant results for search query.

## 4.3  Summary

This chapter has covered results, analysis and its evaluation. Experimental results have proved that iral Meta search engine has outperformed Google and Bing search. Results have shown some new finding in web-optimizing area. Two major findings achieved from results are; First- iral Meta search has potential for further research; results generated from this search engine are unbiased and are according to its design. Second- There are many hypothesis based on SEO parameters and their relevancy to search engine ranking. Results from this dissertation has confirmed literature hypothesis to the experimental results.

# 5  Project Evaluation

## 5.1  Introduction

This chapter provides listing for evaluating this research. The purpose of this chapter is to evaluate whole dissertation which include literature review, practical work undertaken, results generated from experiments, area for further improvements and limitations. This chapter also includes an evaluation chart which aims to give synopsis of this chapter.

## 5.2  Evaluation of Literature Review

The research conducted prior to this dissertation highlighted many inadequacies in search engine optimization techniques and there effects on web search. In literature review it was noted that search engine optimization is a broad branch of science with little research conducted. Many deficiencies were highlighted such as; how to use SEO in effective manner, what parameters are important in optimizing, relation between SEO and information retrieval. Most of the literature gives a vague idea of SEO techniques and fails to relate its importance to search engines. This art of SEO came to existence after Google rolled out from Stratford University lab and became a market leader in search business. Larry and Page (1998) uses new methods and approaches to retrieve information from clustered web. This was the beginning of new area of science in information retrieval namely SEO.

Main aim of SEO is to help search crawler for finding relevant information on the web. This helps in generic indexing and categorizing of web-pages which helps user to find relevant information using search engines. Other researcher used methods like developing Meta search engine for better performance of user query. Different Meta search engines are developed to this date for generating results based on user preferences, location and using click through data. On the other hand research on SEO only focuses on ranking web pages higher in search engines. They failed to relate the importance of SEO in information retrieval for search engines.

In above research, researcher developed useful insight about working of search engines, way they index and categorize data from WWW and presents data to user's query using information retrieval methods. Experiments produced in this research show the importance of SEO for not only marketing purpose but its effectiveness in information retrieval.

## 5.3 Evaluation of Experiments and Results

Main objective of conducting experiment was to highlight the importance of search engine optimizing techniques in search engines. SEO is always considered as an art for ranking websites higher in search engines result page. Experiments conducted in this research have shown its importance not only in ranking but its significance in categorizing web-pages.

iral Meta search engine is developed to extract information from existing search engines, Google and Bing and display results based on search engine optimizing parameters. Results from this Meta search engine when compared to users query shows significant findings. It was noted that web-pages those were optimized precisely were indexed by both search engines and categorized according to searched term meaning. These web-pages contain useful information about user's search query. Analysis of this results shows that iral Meta search engine outperformed Google and Bing with high precession rate of 0.43.

The above experiment also provides important SEO techniques used by search engines. In a comparison it was seen both search engine has its own priorities for SEO factors. This factors affects in indexing and ranking of websites thus for same search query search engines displays different results. From analysis of results it can be rightly said that SEO factors  (mostly meta data, keywords) are not only important for ranking of web-pages but also plays important role in categorizing web data.

## 5.4  Evaluation of Dissertation Objectives

All the objectives of this dissertation are successfully met. Objectives include finding best techniques for search engine optimization, their effectiveness in indexing and categorizing web-data, position in ranking strategies and their importance in user search query to results displayed by different search engines. It also include highlighting important SEO factors and implementation methods for better optimizing websites.

## 5.5  Achievements

This research not only helps in better understanding given topic but in small time frame has achieved milestones:

| Project Process | Product | Impact on Target Group |
|---|---|---|
| Development | iral Meta search engine | 1. New business for Abster-it 2. Helps in future research |
| Web Mining | Collection of SERP Data | Useful as research data |
| Results | SEO Parameters | Webmasters, Website developers and SEO Consultant |

Table 20: List of Project Achievements

## 5.6  Limitations

World Wide Web is ever changing and constantly expanding in nature. There were certain limitations in this dissertation. The design of iral Meta search engine includes Google and Bing API for generating results. It has limitations of 100 queries per day. Due to limitation of time, only 10 results from Google and Bing search were analysed. Precision and Recall of iral Meta search is high based on first 10 results analysis. Only two query terms were included in experiments. For manual results a batch of 50 students was selected and asked to rank search engines. Every individual has its own preferences and understanding. Results may vary if a large user base is involved. The set of best SEO parameters is designed after analysing small fraction of World Wide Web.

## 5.7  Area of Improvement

There are certain limitations in this research mentioned above. To start with, design of Meta search engine can be more efficient and effective if written in object oriented programming language. For finding results more number of keywords should be included. For analysing data and training for search results parameters more time should be allocated in this section.

# 6  Conclusion and Recommendation

## 6.1  Conclusion

The initial objective of this dissertation was to identify important search engine optimizing techniques used by webmasters to increase ranking of websites in search engine result page. The research question for this dissertation was raised in a study by (Pringle *et al.*, 1998) on the importance of SEO techniques in ranking search engines results. Another study by (Berman and Katona, 2011) uses black hat SEO methods and suggests it does increase page rank of websites and users satisfaction. There are two types of SEO techniques: Black hat methods and White hat methods. This research is based on the study of white hat SEO techniques.

In order to answer the research question and breaking new finding, number of objectives and milestone were acknowledged making sure the research remains on right track. Initial steps include interviewing website developers and using web-mining techniques to identify search engine optimizing techniques.

As part of this dissertation, an intelligent Meta search engine *(namely, iral)* is developed which aggregate results from two search engines, Google and Bing. To understand the importance of SEO factors, webpages form this Meta search engine are ranked according to certain SEO parameters using a new ranking algorithm.

Experimental results have proved that iral Meta search engine with average precision of 0.48 has outperformed Google (0.44) and Bing (0.31) search engines. Initial results were promising and findings revealed that search engine optimization is useful in categorizing web data effectively and ranking based on these parameters can achieve results more beneficial for user query thus increasing user satisfaction.

The iral Meta search engine is online for users and there is interesting possibility for learning lifetime. The system will improve over the time and interactive feedbacks from user will help in future changes.

Interestingly, it was observed that high level of web-page optimization is useful when search algorithms of search engines are less accurate.

This research has also addressed important SEO techniques and their weightage in indexing and ranking webpages by Google and Bing search engines. This finding will help webmasters and developers for better optimizing their websites for search engines, ultimately increasing their ranking. Results also provides important recommendation to webmasters that optimization to certain level helps website ranking and indexing but using black hat SEO techniques may cause webpage been banned from search engines.

## 6.2  Recommendation and Future Work

It was observed in the research that SEO techniques are not only important in website ranking but it plays a vital role in its indexing and categorizing web data. The use of SEO can be helpful in field of information retrieval. Although this research has given a vague idea of using SEO for information retrieval, new research can aim for working in this field. A web crawler can be designed which can extract information from websites and rank them according to optimizing factors discussed in this research.

On the other note the design of iral Meta search engine can be improved by connecting more search engines to it. Design can include a dynamic engine selector which can aim to select search engines specific to user query. Such as for video it can select YouTube and Blinkz, for file search Torrent and Google. This will help in generating more specific results to user queries.  iral Meta search engine can motivate the design of other Meta search engines and can develop them using new techniques (like changing retrieval function) by relating what have been achieved in this study.

Other research can include optimizing websites using SEO factors presented in this research. This will help to understand the ever-changing ranking strategies of search engines.

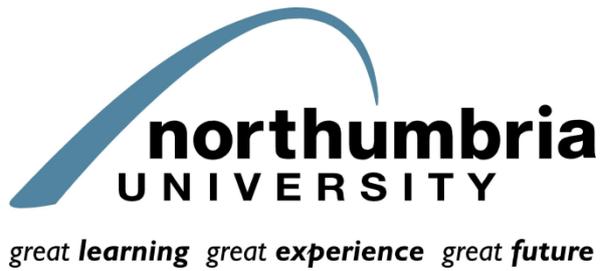
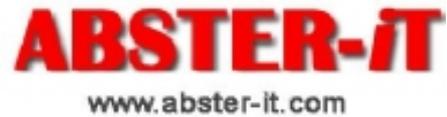

# A Proposal to Develop Intelligent Meta Search Engine

***Abstract:*** *This report presents a plan to develop an intelligent Meta search engine, iral. This Meta search engine is aimed to provide comprehensive, efficient and relevant search results for given queries by combining results from different search engines. This will provide users to cover a larger database of World Wide Web for their queries and to get more relevant search results. On the other hand this Meta search engine will be creating a larger user base and attracting online marketing for Abster-iT. This will provide a new business set-up for Abster-iT.*

Submitted By:

**Jai Manral**
**Northumbria University**
**Newcastle Upon Tyne, NE1 8ST**

Submitted To:

**Adam**
**Business Owner and Project Engineer**
**Abster-iT, Newcastle upon Tyne**
**NE5 5EE England**

Table of Contents

# 1. System Request Form:

<div align="right">Date: March 1<sup>st</sup>, 2012</div>

**Project Name:**  iral Meta search engine

**Is it a New System:** Yes

**Development Time Limit**: 1 Months 2 weeks

**Proposer/Customer:**  Abster-it, Newcastle upon Tyne, UK.

**Project Supervisor**: Prof. Alamgir Hossain, Professor of Computer Science, Northumbria University, UK

**Project Developer:** Jai Manral, Northumbria University, Newcastle, UK

**Business need:** Web search engines have gained the title of most useful and high profile resources available on the internet. The market capital of web search engines has risen to billions of dollars.

**Products Details**: A web based Meta search engine which aims to provide better searching experience to users and capture the web search market in United Kingdom.

**Main Functionality:** The proposed meta-search engine, iral, solves the problem of manually searching different web based search engines for finding relevant data by providing a central place and interface where users can search several search engines at once. This benefits users by saving them time from having to individually learn and visit multiple search engines. By means of this model, the meta-searcher can map users' queries into specific sources more accurately, and it can achieve good precision and recall. Moreover, it will benefit the selection of target source and computing priority. Because new search engines emerge frequently and old ones are updated when their function and content change, the data model needs good adaptivity and scalability to keep in step with the rapidly developing World Wide Web.

Other key features:

- Ability to access a cross section of results from several search engines.
- The ability to access multiple databases provides the most up-to-date results.
- No need of indexing the web.
- No need of maintaining a database.

**Expected Value:**

*Tangible Benefit*

- Provide 24/7 access to the system from any location using internet.
- Provide easy access to users for searching the web.
- Provide relevant results from multiple search engines.
- Enhance the user experience of searching the web.
- Financial gain for Abster-it and Developer (Jai Manral)

*Intangible Benefit*

- Increase university's reputation and can help in developing new ideas and projects.

- Allow the developers to increase their development skills
- Helps in future research in areas like meta search engine, information retrieval, page ranking and SEO.

**Special issues or constraints:**

- Must finish by 20$^{th}$ April, 2012
- Failure in meeting deadline may causes discontinuation of the project and funding.

**Deliverables:**

1. Requirement, design and quality assurance documents.
2. Working solution with all components completed and tested.

**Timeline:**

- 7-10 days for planning
- 1 weeks for analysis
- 2 weeks for design
- 2 weeks for implementation

**IPR**: Abster-iT, Newcastle Upon Tyne, UK

**Delivery date of 1st phase for testing**: 16$^{th}$ April, 2012

## 2. Problem Description:

The extraordinary growth of the internet has made it difficult, if not impossible, for search engines to keep up with its immense size and pace. Thus far, the major search engines have only been able to index a fraction of all the data that is available on the internet. Therefore, chances of finding relevant search results occasionally fails if users rely on only one search engine. Thus, the key to effective internet searching is not to rely on one, but rather on multiple search engines.

However, it can be time consuming and tedious to individually visit and perform a search on multiple search engines.

## 3. System Objectives and Advantages:

### 3.1 System Objectives:

A Meta search engine (MSE) is a search engine that collects results from other search engines, and then presents a summary of that information as the results of a search. The MSE iral is a Web service that offers such functionality. In practice, one can adopt different approaches to achieve the same goal; one could build a unique Web service that fulfils all tasks and does everything. With a smaller granularity, iral Meta search engine is a service for each search engine and an extra one that will use all the other to retrieve the results and compile them.

**Usability:** The system is a cloud based application which can be easily accessible from any location just using the internet connection. The application can be used using the smartphones and other electronic pads.

**Functionality:** The system is easy to use. Users can go directly to systems web-domain, there is no login id or password required for searching. The user interface is simple; a text box is displayed for users to enter their quires.

**Secure:** Security of the data is an important aspect and this system will be capable of keeping the data safe and secure from the unauthorized personals.

**Transparency:** This system provides transparency of

# 8.1  3.2 System Advantages:

Meta search engine iral accept users query, and send it out to multiple search engines in parallel. It is quite fast, using private "backdoor" servers made available by the search engines they query. There are several advantages to using iral Meta search engine; most obvious advantage is that user can get results from multiple search engines without having to visit each in turn. Apart from the time savings, it gives search a broader scope, since each individual search engine's index differs from all others.

- Querying multiple engines simultaneously.
- Broader overview of a topic.
- Helps in optimizing the time spent on searching the web for information.
- It eliminates user chances of missing out the information on various other regular search engines when user is accessing only one of them.
- The Meta search gives a quicker view of the high ranking results in search engines.
- Eliminating duplicate results also saves time.

## 4. Use-Case diagram

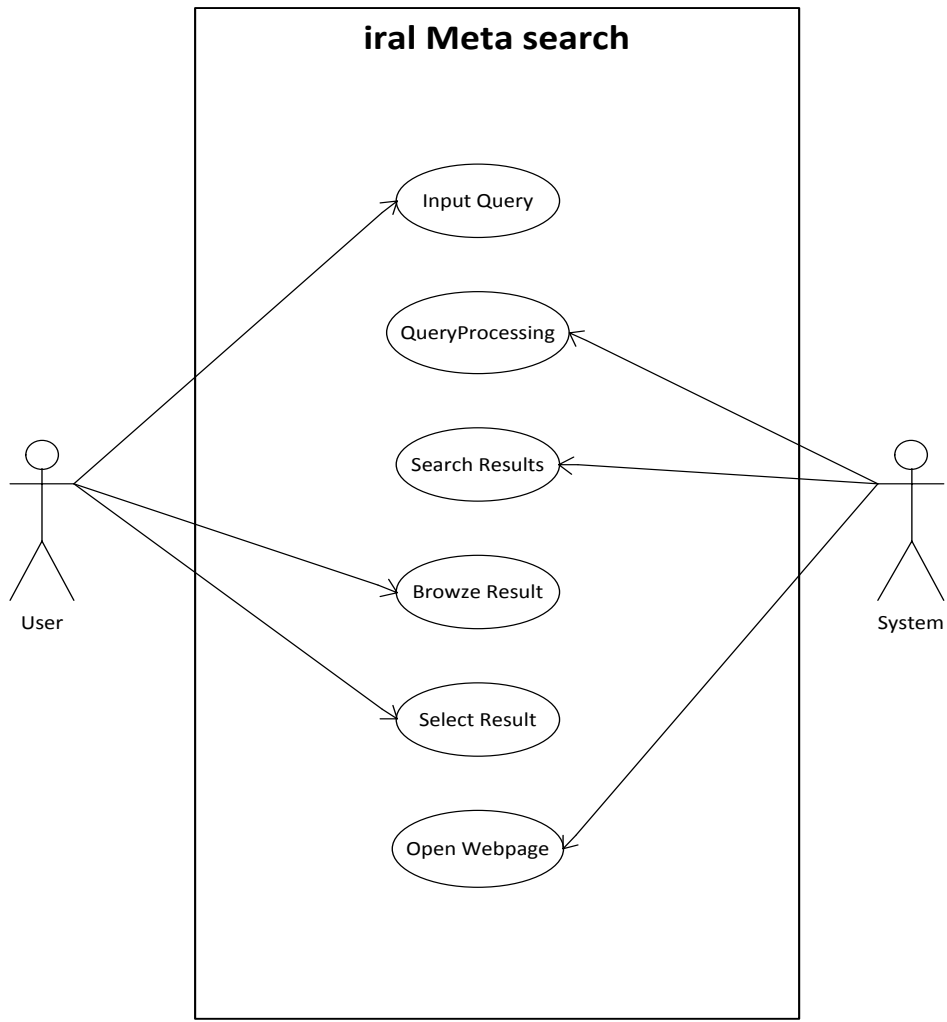

## 8.2 4.1 Advantages

1. **Input Query:** User input search query. It can be text, integers or special characters.
2. **Query Processing:** System will check the query and provide a connection to external search engines.
3. **Search Results:** System process the query and get result from external search engines. It will organize and display results to user.
4. **Browse Results:** User will browse through the results provided by system.
5. **Select Result:** User will select the best matching result from the system.
6. **Open Webpage:** System will connect to web domain and open webpage for result selected by the user.

## 8.3 4.2 Disadvantages

1. **Input Query:** Query cannot be any mathematical formula or equations. System does not support it.
2. **Query Processing:** Sometimes connections are lost due to unavailability of the embedded search engines.

## 5. Feasibility

## 8.4

### 8.4.1 5.1 Technical Feasibility

The technical feasibility is divided into four steps. In this particular software there are no such technical problems which can be faced while development or after deployment. There are very basic and not gruesome requirements in order to finish this project.

**5.1.1 Familiarity with the application:** The application is a new idea but the design and the concepts are based on some of the applications which are currently popular in the World Wide Web. The examples are metacrawlers, dogpile, MetaLib and SiteStep.

**5.1.2 Familiarity with the technology:** The application will not have any new technological requirements which can create any technical risk. This application will be using easy and popular technology such as the usage of the programming languages are PHP, java script, CSS and Fuzzy Logic.

**5.1.3 Project Size:** The project size of this application is estimated around 2 months. This is a demanding and hard application to build, but doesn't have high risk factor. The project can be easily finished in the proposed time limit.

**5.1.4 Compatibility:** As this application is web based and no working will be done using old technology thus the compatibility the proposed project is highly reliable. The project is a web-based application and thus it does not require handling with the existing systems.

## 8.5  5.2 Legal Feasibility

As the business would be based in the UK (though the servers where the software will be executed and utilised by the consumer may be located elsewhere) there are number of UK laws which will have to be adhered to. There is no database for this system and no functionality of user's loggings or ids, hence the Data Protection Act 1998, is not applied to it. Additionally as services will be selling over the internet, thus it is needed to satisfy the stipulations covered by the Distance Selling Regulations Act 2000. As long as we abide by Laws the proposed business venture should be legal.

## 8.6  5.3 Economic Feasibility

In order to get the acute feasibility of the system, various research activities have been done in the current market. Use of search engines such as Google and Yahoo has shown some of the famous systems such as dogpile and metasearchengine. These systems are biggest competitor of iral Meta search in terms of market presence.

| RANK | Provider | Searches (000) | YOY Growth | % of all Searches |
|------|----------|----------------|------------|-------------------|
| | All Search | 8,608,488 | 4.40% | 100.00% |
| 1 | Google Search | 5,510,366 | 7.80% | 64.0% |
| 2 | Yahoo! Search | 1,406,416 | -2.80% | 16.3% |
| 3 | MSN/Windows Live Search | 852,998 | 7.20% | 9.9% |
| 4 | AOL Search | 321,205 | -8.80% | 3.7% |
| 5 | Ask.com Search | 181,617 | 5.90% | 2.1% |
| 6 | My Web Search Search | 59,110 | 3.60% | 0.70% |
| 7 | Comcast Search | 45,338 | -1.80% | 0.50% |
| 8 | Yellow Pages Search | 37,160 | N/A* | 0.40% |
| 9 | NexTag Search | 22,845 | 3.90% | 0.30% |
| 10 | **Dogpile.com Search** | 17,010 | 3.10% | 0.20% |
| Source: Nielsen MegaView Search | | | | |

In order to develop the system the financial requirements are stated below:

| | Development Cost | |
|---|---|---|
| | **Month 1** | **Month 2** |
| **Hardware Costs** | | |
| Desktop System | £395.00 | £0.00 |
| Domain Name | £10.00 | £0.00 |
| Dedicated Server Hosting | £20.00 | £20.00 |
| **Software Costs** | | |
| Microsoft Office Professional License | £70.00 | £0.00 |
| Google API | £62.00 | |
| Bing API | £10.00 | |
| **Staff Costs** | | |
| Jai Manral | £0.00 | £0.00 |
| **Miscellenous Costs** | £10.00 | £10.00 |
| | | |
| **Total Expenditure** | £577.00 | £30.00 |
| | | |
| | **Total Development Cost** | **£607.00** |
| | **Finance Requested** | **£620.00** |

The development cost of the project is low as there is no requirement of any kind of special expertise for developing the system. Though the system is new but the backend and the SDLC involved are in the company standards. The project is estimated to be finished in period of 2 months.

## 8.7  5.4 Organizational Feasibility

| | 01/04/12 | 01/04/13 | 01/04/14 | 01/04/15 | 01/04/16 | 01/04/17 | Total |
|---|---|---|---|---|---|---|---|
| **Organizational Feasiblity** | | | | | | | |
| **Benefit** | | | | | | | |
| Online Advertisement | £0.00 | £1,500.00 | £1,875.00 | £2,343.75 | £2,929.69 | £3,662.11 | |
| **Total Benefit** | **£0.00** | **£1,500.00** | **£1,875.00** | **£2,343.75** | **£2,929.69** | **£3,662.11** | |
| **Present Value Total Benefit** | **£0.00** | **£1,428.57** | **£1,700.68** | **£2,024.62** | **£2,410.26** | **£2,869.36** | **£10,433.49** |
| Development  Labor Cost | | | | | | | |
| Development Team Wages | £0.00 | £80.00 | £168.00 | £352.80 | £740.88 | £1,555.85 | |
| Maintainance Team Wages | | £120.00 | £252.00 | £529.20 | £1,111.32 | £2,333.77 | |
| Hardware Cost | | | | | | | |
| Desktop System | £395.00 | £0.00 | £0.00 | £0.00 | £0.00 | £0.00 | |
| Domain Name | £10.00 | £0.00 | £0.00 | £0.00 | £0.00 | £0.00 | |
| Dedicated Server Hosting | £20.00 | £21.20 | £22.47 | £23.82 | £25.25 | £26.76 | |
| Software Cost | | | | | | | |
| Microsoft Office Professional | £70.00 | £0.00 | £0.00 | £0.00 | £0.00 | £0.00 | |
| Google API | £62.00 | £62.00 | £62.00 | £62.00 | £62.00 | £62.00 | |
| Bing API | £10.00 | £10.00 | £10.00 | £10.00 | £10.00 | £10.00 | |
| **Total Development Cost** | **£567.00** | £0.00 | £0.00 | £0.00 | £0.00 | £0.00 | |
| **Operational Cost** | | | | | | | |
| Hardware (Server hosting) | | £21.20 | £22.05 | £22.93 | £23.85 | £24.80 | |
| Software | | £120.00 | £127.20 | £134.83 | £142.92 | £151.50 | |
| Operational Labor | | £200.00 | £420.00 | £882.00 | £1,852.20 | £3,889.62 | |
| Miscellanous Cost | | £10.00 | £10.10 | £10.20 | £10.30 | £10.41 | |
| **Total Operational Cost** | | **£351.20** | **£579.35** | **£1,049.96** | **£2,029.27** | **£4,076.32** | |
| **Total Costs** | **£567.00** | **£351.20** | **£579.35** | **£1,049.96** | **£2,029.27** | **£4,076.32** | |
| **Present Value Total Costs** | **£567.00** | **£334.48** | **£525.49** | **£907.00** | **£1,669.49** | **£3,193.91** | **£7,197.35** |

| NPV (PV Total Benefit-PV Total Cost) | | | | | | | £3,236.14 |
|---|---|---|---|---|---|---|---|

The system is economically feasible to develop. The Breakeven point is calculated at **2.49 years**. The calculation of the benefits is estimated by studying the present systems such as metacrawler.com and Dogpile. The system can earn the revenue from online advertisement. Online advertisements can be provided by the Google on pay per click basis.

System is feasible and will start making the money from the launch date. The Return on Investment (ROI) is calculated at 200% considering the value of pound which is needed to build the system and the rate of change in its value after a period of 5 years.

The NPV, net present value is calculated at **£ 3,236.14** at the end of the 5 years. It is been calculated by deducting the total benefits earned from the system to the total cost which is needed to develop the system. From an organizational perspective, this project has low risk. Thus the investment to this project will be a beneficiary one.

## 6. Work Plan

### 6.1   Function Point

| Description | | Complexity | | | |
|---|---|---|---|---|---|
| | Total Number | Low | Medium | High | Total |
| Inputs | 1 | 0*0 | 0*0 | 1*6 | 6 |
| Outputs | 3 | 0*4 | 2*5 | 1*7 | 17 |
| Queries | 5 | 1*3 | 3*4 | 1*6 | 21 |
| Program Interfaces | 2 | 0*5 | 1*7 | 1*10 | 17 |
| Files | 0 | 0*7 | 0*10 | 0*15 | 0 |
| **Total Unadjusted Function Points (TUFP)** | | | | | **61** |

The total unadjusted function points in the system are calculated to be **61.**

| | | |
|---|---|---|
| Data Communication | 1 | On-line data collection or TP (teleprocessing) front-end to a batch process or query system. |
| Heavy use configuration | 1 | Stated operation restrictions require special constraints on the application in the central processor or a dedicated processor |
| Transaction rate | 4 | High transaction rates stated by the user in the application requirements or service level agreements are high enough to require performance analysis tasks in the design phase. |
| End-User efficiency | 1 | Navigational aids (for example, function keys, jumps, dynamically generated menus). |
| Complex Processing | 0 | None |
| Installation Ease | 1 | No special considerations were stated by user and no special set-up required for installation. |
| Multiple Sites | 3 | Needs of multiple sites were considered in the design and the application is designed to operate under different hardware and/or software environments. |
| Performance | 3 | Response time or throughput is critical during all business hours. No special design for CPU utilization was required. Processing deadline requirements with interfacing systems are constraining. |
| Distributed Functions | 3 | Distributed processing and data transfer are on-line and in both directions. |
| Online data Entry | 0 | 16 percent to 23 percent of transactions are interactive data entry. |
| Online updates | 4 | On-line update of four or more control files. Protection against data lost is essential and has been specially designed and programmed in the system. |
| Reusability | 5 | The application was specifically packaged and/or documented to ease re-use, and application is customized by user at source code level. |
| Operation ease | 3 | No special operational considerations other than the normal back-up procedures were stated by the user. |
| Extensibility | 3 | Can be easily extended for future usage |
| **Total Processing Complexity (PC)** | **32** | |

**Adjusted Project Complexity (APC) = 0.65+ (0.01* 32) = 0.97**

**Total Adjusted Function Points (TAFP) = 0.97(APC) * 61 (TUFP) =60 (TAFP)**

**Lines of Code = 60 (Total Adjusted Function Point) * 15 (PHP) = 900**

## 6.2    COCOMO Model

It is used to calculate the effort needed to build the system. The detailed model uses different efforts multipliers for each cost drivers attribute these Phase Sensitive effort multipliers are each to determine the amount of effort required to complete each phase Chiang, R. (2009).

Below is calculated the Efforts in person months.

**Effort (in person-months)** = 1.4* 0.9 = 1.26 **person-*month***

The time estimated for the project is approx. 1 months but having a working force of 1 person per month can increase the time by 1 month 1 week. Hence the total amount of time needed to complete the task is **38 days**. Below is the Gantt chart for the project.

### 6.3    Gantt chart

| WBS | Task Name | Duration | Start | Finish | Predecessors | Resource Names |
|---|---|---|---|---|---|---|
| **1** | **iral Meta Search Engine** | **38 days** | **Mon 05/03/12** | **Wed 25/04/12** | | |
| **1.1** | **Planning Phase** | **3.75 days** | **Mon 05/03/12** | **Thu 08/03/12** | | |
| 1.1.1 | Develop System Request Form | 0.5 days | Mon 05/03/12 | Mon 05/03/12 | | Jai Manral |
| 1.1.2 | Problem Description | 0.5 days | Mon 05/03/12 | Mon 05/03/12 | 3 | Jai Manral |
| 1.1.3 | Objectives | 0.25 days | Tue 06/03/12 | Tue 06/03/12 | 4 | Jai Manral |
| 1.1.4 | Use Case Diagrams | 0.5 days | Tue 06/03/12 | Tue 06/03/12 | 5 | Jai Manral |
| 1.1.5 | Feasibilty Analysis | 1 day | Tue 06/03/12 | Wed 07/03/12 | 6 | Jai Manral |
| 1.1.6 | Work Plan | 1 day | Wed 07/03/12 | Thu 08/03/12 | 7 | Jai Manral |
| **1.2** | **Analysis Phase** | **7 days** | **Thu 08/03/12** | **Mon 19/03/12** | | |
| 1.2.1 | Evaluate Alternatives | 3 days | Thu 08/03/12 | Tue 13/03/12 | 8 | Jai Manral |
| 1.2.2 | Feasibilty Analysis | 2 days | Tue 13/03/12 | Thu 15/03/12 | 10 | Jai Manral |
| 1.2.3 | System Requirement | 1 day | Thu 15/03/12 | Fri 16/03/12 | 11 | Jai Manral |
| 1.2.4 | Requirement Analysis | 1 day | Fri 16/03/12 | Mon 19/03/12 | 12 | Jai Manral |
| **1.3** | **Design Phase** | **5 days** | **Mon 19/03/12** | **Mon 26/03/12** | | |
| 1.3.1 | Use Case Diagrams | 1 day | Mon 19/03/12 | Tue 20/03/12 | 13 | Jai Manral |

| 1.3.2 | Sequence Diagram | 1 day | Tue 20/03/12 | Wed 21/03/12 | 15 | Jai Manral |
|---|---|---|---|---|---|---|
| 1.3.3 | Class Diagram | 1 day | Wed 21/03/12 | Thu 22/03/12 | 16 | Jai Manral |
| 1.3.4 | State Diagram | 1 day | Thu 22/03/12 | Fri 23/03/12 | 17 | Jai Manral |
| 1.3.5 | Detail Specification | 1 day | Fri 23/03/12 | Mon 26/03/12 | 18 | Jai Manral |
| **1.4** | **Implementation Phase** | **22 days** | **Mon 26/03/12** | **Tue 24/04/12** | | |
| **1.4.1** | **Development Phase** | **12.5 days** | **Mon 26/03/12** | **Wed 11/04/12** | | |
| 1.4.1.1 | Design UI | 0.5 days | Mon 26/03/12 | Tue 27/03/12 | 19 | Jai Manral |
| 1.4.1.2 | Integration of SE's | 0.5 days | Mon 26/03/12 | Mon 26/03/12 | | Jai Manral |
| 1.4.1.3 | Integration of Knowledge Base | 4 days | Mon 26/03/12 | Fri 30/03/12 | 23 | Jai Manral |
| 1.4.1.4 | Ranking Functionality | 6 days | Fri 30/03/12 | Mon 09/04/12 | 23,24 | Jai Manral |
| 1.4.1.5 | Developer Debugging | 2 days | Mon 09/04/12 | Wed 11/04/12 | 25 | Jai Manral |
| **1.4.2** | **Testing Phase** | **2 days** | **Wed 11/04/12** | **Fri 13/04/12** | | |
| 1.4.2.1 | Develop Unit Test Plan | 1 day | Wed 11/04/12 | Thu 12/04/12 | 26 | Jai Manral |
| 1.4.2.2 | Develop Test Use Case | 1 day | Thu 12/04/12 | Fri 13/04/12 | 28 | Jai Manral |
| **1.4.3** | **Product Testing** | **3 days?** | **Fri 13/04/12** | **Wed 18/04/12** | | |
| 1.4.3.1 | Test Product | 1 day? | Fri 13/04/12 | Mon 16/04/12 | 29 | Jai Manral |
| 1.4.3.2 | Repair Defects | 1 day? | Mon 16/04/12 | Tue 17/04/12 | 31 | Jai Manral |
| 1.4.3.3 | Regression Testing | 1 day? | Tue 17/04/12 | Wed 18/04/12 | 32 | Jai Manral |
| **1.4.4** | **Deployment Phase** | **3.75 days** | **Wed 18/04/12** | **Tue 24/04/12** | | |
| 1.4.4.1 | Deploy to Live Environment | 6 hrs | Wed 18/04/12 | Thu 19/04/12 | 33 | Jai Manral |
| 1.4.4.2 | Customer Testing | 1 day | Thu 19/04/12 | Fri 20/04/12 | 35 | Abster-iT |
| 1.4.4.3 | Customer Sign off | 1 day | Fri 20/04/12 | Mon 23/04/12 | 36 | Abster-iT |
| 1.4.4.4 | Contract Close out | 1 day | Mon 23/04/12 | Tue 24/04/12 | 37 | Abster-iT |
| 1.5 | **Project Close Out** | 0.25 days | Wed 25/04/12 | Wed 25/04/12 | | Jai Manral and Abster-iT |

## ANALYSIS AND DESIGN PHASE

# 1. Information Gathering

**INTERVIEW QUESTIONS FROM USERS**

1. Is it possible to examine all the documents on the Web, how do you search for query?
2. Which search engine is "the biggest"? According to you.
3. Which search engine you 'mostly' use?
4. How long does it take you for getting useful search result?
5. How often do you use search engine?
6. How many pages of search results you browse for fining results?
7. Are you happy with search engines performances?
8. Have you tried semantic or smart search engines?
9. Have you ever used 'advance tab' in search engines for specifying you query?
10. How useful is search engine to find information on the web?
11. Have you heard of Meta Search engines?
12. Will it be helpful if some software will search your query in 2 or more search engine and extract common result to you?

## 2. Sequence Diagrams

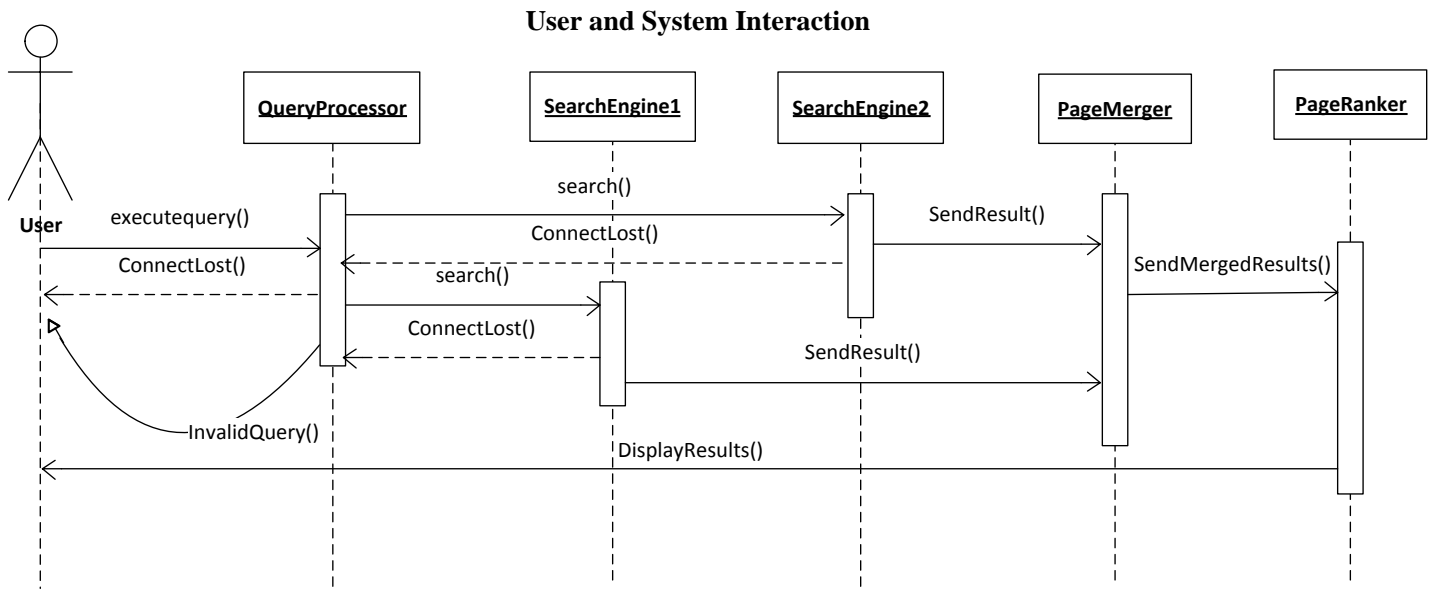

User and System Interaction

## 3. Class Structure Diagram

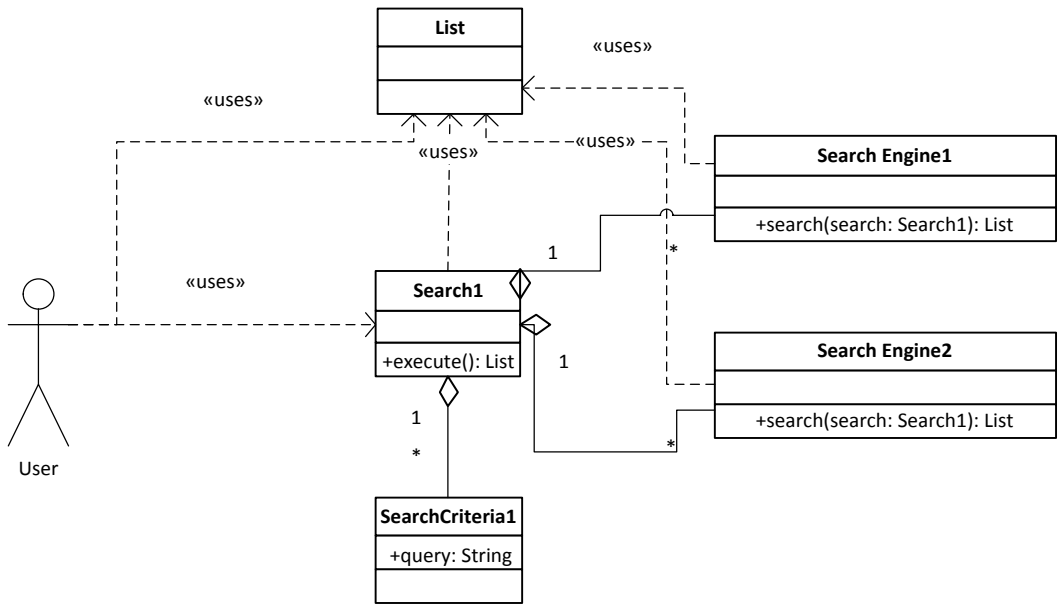

## 4. State Diagrams

**SEARCH STATE**

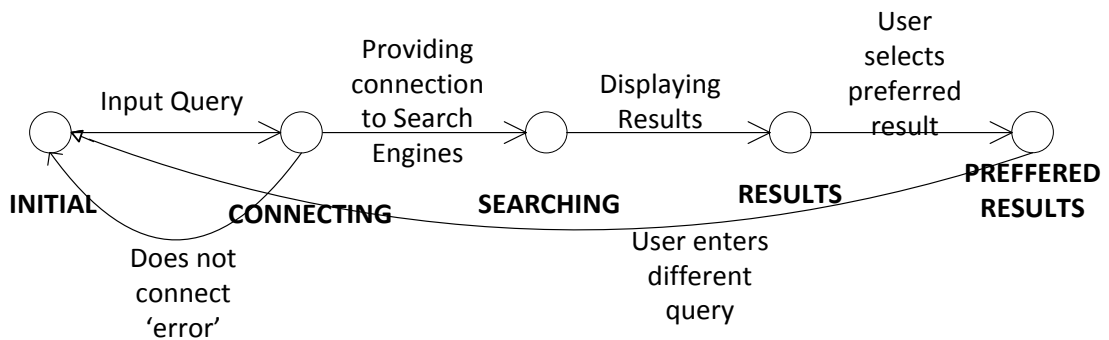

## 5. User Interface

### a. Window/Frame design

**Frame Design:**

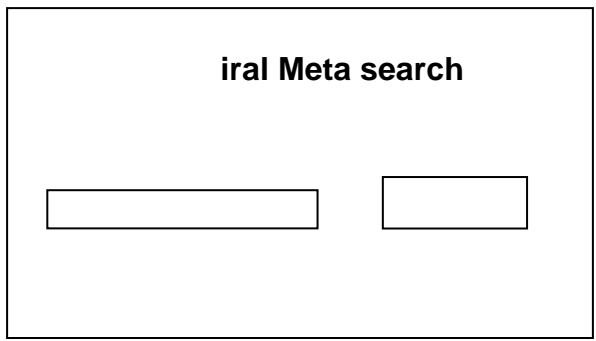

**State Table:**

| Component | Title for source code | Visibility | Active | Text |
|-----------|----------------------|------------|--------|------|
| Text box | Txtbx1 | Yes | Yes | None |
| Button | btmember | Yes | No | Search |

**State table:**

| Component | Title for source code | Visibility | Active | Text |
|---|---|---|---|---|
| Text box | Txtbx1 | Yes | Yes | None |
| Button1 | Btmember1 | Yes | Yes | Search |
| Text Result 1 | Txtbx2 | Yes | Yes | Result 1 |
| Text Result 2 | Txtbx3 | Yes | Yes | Result 2 |
| Text Result 3 | Txtbx4 | Yes | Yes | Result 3 |
| Text Result 4 | Txtbx5 | Yes | Yes | Result 4 |
| Text Result 5 | Txtbx6 | Yes | Yes | Result 5 |
| Text Result 6 | Txtbx7 | Yes | Yes | Result 6 |
| Text Result 7 | Txtbx8 | Yes | Yes | Result 7 |
| Text Result 8 | Txtbx9 | Yes | Yes | Result 8 |
| Text Result 9 | Txtbx10 | Yes | Yes | Result 9 |
| Text Result 10 | Txtbx11 | Yes | Yes | Result 10 |
| Button2 | Btmember2 | Yes | Yes | 1 |
| Button3 | Btmember3 | Yes | Yes | 2 |
| Button4 | Btmember4 | Yes | Yes | 3 |

## 6. Program Specification Form

Module: **Index Page**
Name: Search Task
Purpose: To search
Programmer: Jai Manral
Date Due:

• **PHP**   ☐ PowerScript   ☐ COBOL   ☐ VISUAL BASIC

Events
User input the text for search

| Input Name: | Type: | Provided by: | Notes: |
| --- | --- | --- | --- |
| | | | |
| SearchId | Varchar() | Window GUI | |
| | | | |
| | | | |

| Output Name: | Type: | Used by: | Notes: |
| --- | --- | --- | --- |
| | | | |
| | | | |
| | | | |

**Pseudo code**
If getQuery ()=true then
        Send Query
        Exits
    Else
        Return Connection Lost

Module: **Result Page**
Name: Search Result
Purpose: To display search results
Programmer: Jai Manral
Date Due:

☐ C          • PHP          ☐ COBOL          ☐ JSP

Events ─────────────────────────────────────────
For user query, results to be displayed numbering 10 including 5 pages

| Input Name: | Type: | Provided by: | Notes: |
|---|---|---|---|
| SearchID | Varchar () | Window GUI | |
| | | | |
| | | | |
| | | | |

| Output Name: | Type: | Used by: | Notes: |
|---|---|---|---|
| | | | |
| ResultID | Varchar () | Result Page GUI | |
| | | | |
| | | | |

**Pseudo code**
If getUserSearchID()=true then
      Generate ResultsID
        Exits
    Else
      Return Connection Lost

## 7. Conclusion

Meta-search engines, also known as multiple search engines, metasearchers, or metacrawlers, are special search tools that present the results by accessing multiple search engines and web directories. This way, they allow users to quickly receive combined results that are merged in one place at once. Thus, web users neither need to type the query several times nor have to access every single search engine by themselves. This job will be done for the users by meta-search engines, which might additionally suggest engines that the user had not considered before.

By performing a search query, meta-search engines transmit the typed terms simultaneously to multiple individual search engines. Multi-search engines don't do the crawling or maintain their own database like single search engines, but usually filter the results they found instead. Based on a specific algorithm, they eliminate duplicates and rank the results from their sources into a list. The list of collection will be displayed on the SERP, very similar to the search engines' results page that relies on the indices of other search engines.

In dependence on studies of Dogpile, findings show that within the four search engines Google, Yahoo Search, Bing and Ask overlap across their first results page was only 0.6% for a given query. On the contrary, each engine still finds a large amount of unique results search engines which means that they have a lack of duplication and each of them mostly don't see pages as equally important. In other words, the imputation that search engines are the same is a myth. Meta-search engines collect and thus, cover the best results of the sources, including overlap. This is crucial, since overlapped documents tend to be more relevant.

The economic feasibility given in this reports forecasts the financial benefits for Abster-iT in near future. It will be another service domain for the business which has chances for further development, can accrue a larger user base and attract online                                              marketers.

Appendix beginning from 2- 9

# Appendix 2: Application Codes

## Google API

```
<script type="text/javascript">

  var _gaq = _gaq || [];
  _gaq.push(['_setAccount', 'UA-31712426-1']);
  _gaq.push(['_trackPageview']);

  (function() {
    var ga = document.createElement('script'); ga.type =
'text/javascript'; ga.async = true;
    ga.src = ('https:' == document.location.protocol ?
'https://ssl' : 'http://www') + '.google-analytics.com/ga.js';
    var s = document.getElementsByTagName('script')[0];
s.parentNode.insertBefore(ga, s);
  })();

</script>

<?php
error_reporting(1);
//URL of targeted site
//$url="http://api.bing.net/json.aspx?AppId=DF8EEA8B1242C4510F05
D9C4E87EC1B85AA9DD79&Version=2.2&Market=en-
IN&Query=".urlencode($_REQUEST['query'])."&Sources=web+spell&Web
.Count=40";
//$url_g="https://www.googleapis.com/customsearch/v1?key=AIzaSyD
Zt7EPYtO10OEpiScPsR_1sz48fhkgjOQ&cx=002075151492928372411:3ohn-
d250gm&q=".$_REQUEST['query']."&alt=json";
$url_g1="https://www.googleapis.com/customsearch/v1?q=".urlencod
e($_REQUEST['query'])."&cref=http%3A%2F%2Finspireweb.co.in%2FGoo
gle.html&cx=002075151492928372411%3A3ohn-
d250gm&gl=IN&googlehost=https%3A%2F%2Fwww.google.co.in&num=10&st
art=1&pp=1&pp=1&key=AIzaSyDZt7EPYtO10OEpiScPsR_1sz48fhkgjOQ";
$url_g2="https://www.googleapis.com/customsearch/v1?q=".urlencod
e($_REQUEST['query'])."&cref=http%3A%2F%2Finspireweb.co.in%2FGoo
gle.html&cx=002075151492928372411%3A3ohn-
d250gm&gl=IN&googlehost=https%3A%2F%2Fwww.google.co.in&num=10&st
art=11&pp=1&pp=1&key=AIzaSyDZt7EPYtO10OEpiScPsR_1sz48fhkgjOQ";
$url_g3="https://www.googleapis.com/customsearch/v1?q=".urlencod
e($_REQUEST['query'])."&cref=http%3A%2F%2Finspireweb.co.in%2FGoo
gle.html&cx=002075151492928372411%3A3ohn-
d250gm&gl=IN&googlehost=https%3A%2F%2Fwww.google.co.in&num=10&st
art=21&pp=1&pp=1&key=AIzaSyDZt7EPYtO10OEpiScPsR_1sz48fhkgjOQ";
$url_g4="https://www.googleapis.com/customsearch/v1?q=".urlencod
e($_REQUEST['query'])."&cref=http%3A%2F%2Finspireweb.co.in%2FGoo
gle.html&cx=002075151492928372411%3A3ohn-
d250gm&gl=IN&googlehost=https%3A%2F%2Fwww.google.co.in&num=10&st
art=31&pp=1&pp=1&key=AIzaSyDZt7EPYtO10OEpiScPsR_1sz48fhkgjOQ";
```

```php
//google
$ch = curl_init();

// set URL and other appropriate options
curl_setopt($ch, CURLOPT_SSL_VERIFYPEER, false);
curl_setopt($ch, CURLOPT_URL, $url_g1);
curl_setopt($ch, CURLOPT_HEADER, 0);
curl_setopt($ch, CURLOPT_RETURNTRANSFER, true);

// grab URL and pass it to the browser
$output_g1 = curl_exec($ch);
curl_setopt($ch, CURLOPT_URL, $url_g2);
$output_g2 = curl_exec($ch);
curl_setopt($ch, CURLOPT_URL, $url_g3);
$output_g3 = curl_exec($ch);
curl_setopt($ch, CURLOPT_URL, $url_g4);
$output_g4 = curl_exec($ch);
curl_close($ch);
$output_ar=json_decode ($output,true,512);
$output_ar_g1=json_decode ($output_g1,true,512);
$output_ar_g2=json_decode ($output_g2,true,512);
$output_ar_g3=json_decode ($output_g3,true,512);
$output_ar_g4=json_decode ($output_g4,true,512);
$search_results_ar_g1=$output_ar_g1['items'];
$search_results_ar_g2=$output_ar_g2['items'];
$search_results_ar_g3=$output_ar_g3['items'];
$search_results_ar_g4=$output_ar_g4['items'];
$search_results_ar_g=array_merge($search_results_ar_g1,$search_r
esults_ar_g2,$search_results_ar_g3,$search_results_ar_g4);
$i=0;
$final_result_ar=array();
foreach($search_results_ar_g as $key=>$value)
{
      //echo "==<br>result no:".$key."<br>Title is:";
      $final_result_ar[$i]['title']=$value['title'];
      $final_result_ar[$i]['link']=$value['link'];
      //link on the title
      $final_result_ar[$i]['description']=$value['htmlSnippet'];
                //description to show
      $final_result_ar[$i]['displaylink']=$value['displayLink'];
                //link displayed below description
      $i=$i+2;
}
```

### Bing API

```php
<?php
error_reporting(1);
//URL of targeted site
$url="http://api.bing.net/json.aspx?AppId=DF8EEA8B1242C4510F05D9
C4E87EC1B85AA9DD79&Version=2.2&Market=en-
IN&Query=".urlencode($_REQUEST['query'])."&Sources=web+spell&Web
.Count=40";

//bing
$ch = curl_init();

// set URL and other appropriate options
curl_setopt($ch, CURLOPT_URL, $url);
curl_setopt($ch, CURLOPT_HEADER, 0);
curl_setopt($ch, CURLOPT_RETURNTRANSFER, true);

// grab URL and pass it to the browser

$output = curl_exec($ch);
curl_close($ch);

$output_ar=json_decode ($output,true,512);

$search_results_ar=$output_ar['SearchResponse']['Web']['Results'
];
$i=1;
foreach($search_results_ar as $key=>$value)
{
      //echo "<br>Result no ".$i."<br>";
      $final_result_ar[$i]['title']=$value['Title'];
      $final_result_ar[$i]['link']=$value['Url'];
            //link on the title
      $final_result_ar[$i]['description']=$value['Description'];
            //description to show
      $final_result_ar[$i]['displaylink']=$value['DisplayUrl'];
            //link displayed below description
      $i=$i+2;
}
foreach($final_result_ar as $key=>$value)
{
      $new_key=$value['displaylink'];
      $final2[$new_key]['title']=$value['title'];
      $final2[$new_key]['link']=$value['link'];
      $final2[$new_key]['description']=$value['description'];
      $final2[$new_key]['displaylink']=$value['displaylink'];
}
```

```
?>
```

## Pagination

```
<!DOCTYPE html PUBLIC "-//W3C//DTD XHTML 1.0 Transitional//EN"
"http://www.w3.org/TR/xhtml1/DTD/xhtml1-transitional.dtd">
<html xmlns="http://www.w3.org/1999/xhtml">
<head>
<meta http-equiv="Content-Type" content="text/html; charset=utf-
8" />
<title>pagination</title>
</head>
<script type="text/javascript">
function showpage(page_no)
{
     if(page_no==1)
     {

     document.getElementById("page1").style.display="block";
          document.getElementById("page2").style.display="none";
          document.getElementById("page3").style.display="none";
          document.getElementById("page4").style.display="none";
          document.getElementById("page5").style.display="none";
     }
     else if(page_no==2)
     {
          document.getElementById("page1").style.display="none";

     document.getElementById("page2").style.display="block";
          document.getElementById("page3").style.display="none";
          document.getElementById("page4").style.display="none";
          document.getElementById("page5").style.display="none";
     }
     else if(page_no==3)
     {
          document.getElementById("page1").style.display="none";
          document.getElementById("page2").style.display="none";

     document.getElementById("page3").style.display="block";
          document.getElementById("page4").style.display="none";
          document.getElementById("page5").style.display="none";
     }
     else if(page_no==4)
     {
          document.getElementById("page1").style.display="none";
          document.getElementById("page2").style.display="none";
          document.getElementById("page3").style.display="none";

     document.getElementById("page4").style.display="block";
          document.getElementById("page5").style.display="none";
     }
     else if(page_no==5)
     {
          document.getElementById("page1").style.display="none";
```

```
            document.getElementById("page2").style.display="none";
            document.getElementById("page3").style.display="none";
            document.getElementById("page4").style.display="none";

      document.getElementById("page5").style.display="block";
      }
}
</script>

<?php
$final_result_ar=array();
foreach($search_results_ar_g as $key=>$value)
{
      //echo "==<br>result no:".$key."<br>Title is:";
      $final_result_ar[$i]['title']=$value['title'];
      $final_result_ar[$i]['link']=$value['link'];
      //link on the title
      $final_result_ar[$i]['description']=$value['htmlSnippet'];
      //description to show
      $final_result_ar[$i]['displaylink']=$value['displayLink'];
      //link displayed below description
      $i=$i+2;
}

$search_results_ar=$output_ar['SearchResponse']['Web']['Results'
];
$i=1;
foreach($search_results_ar as $key=>$value)
{
      //echo "<br>Result no ".$i."<br>";
      $final_result_ar[$i]['title']=$value['Title'];
      $final_result_ar[$i]['link']=$value['Url'];
      //link on the title
      $final_result_ar[$i]['description']=$value['Description'];
      //description to show
      $final_result_ar[$i]['displaylink']=$value['DisplayUrl'];
      //link displayed below description
      $i=$i+2;
}
foreach($final_result_ar as $key=>$value)
{
      $new_key=$value['displaylink'];
      $final2[$new_key]['title']=$value['title'];
      $final2[$new_key]['link']=$value['link'];
      $final2[$new_key]['description']=$value['description'];
      $final2[$new_key]['displaylink']=$value['displaylink'];
}
echo "Search Results:<br>";
$df=0;
foreach($final2 as $key=>$value)
{
      if($df==50)
```

```
        break;

        if($df==0)
                echo "<div id='page1'>";
        else if($df==10)
                echo "<div style='display:none;' id='page2'>";
        else if($df==20)
                echo "<div style='display:none;' id='page3'>";
        else if($df==30)
                echo "<div style='display:none;' id='page4'>";
        else if($df==40)
                echo "<div style='display:none;' id='page5'>";
                echo                                        "<br><a
href='".$final2[$key]['link']."'>".$final2[$key]['title']."</a><
br>";
                echo $final2[$key]['description']."<br>";
                echo                                        "<span
class='resulturl'>".$final2[$key]['displaylink']."</span><br>";

        if($df==9)
                echo "</div>";
        else if($df==19)
                echo "</div>";
        else if($df==29)
                echo "</div>";
        else if($df==39)
                echo "</div>";
        else if($df==49)
                echo "</div>";
        $df++;
}
?>
```

# Appendix 3: Algorithms Classification

**Algorithm: iral calculator**

Input: Page P, in link and Out link Weights of All back links of P, Query Q, d (damping factor).

Output: Rank score

Step 1: Relevance calculation:

Find all meaningful word strings of Q (say N)

Find whether the N strings are occurring in P or not?

Z= Sum of frequencies of all N strings.

S= Set of the maximum possible strings occurring in P.

X= Sum of frequencies of strings in S.

Content Weight (CW) = X/Z

C= No. of query terms in P

D= No. of all query terms of Q while ignoring stop words.

Probability Weight (PW) = C/D

Step 2: Rank calculation:

Find all back links of P (say set B).

PR (P) = (1-d) +d

**Output PR (P) i.e. the Rank score**

**Google page rank calculation**

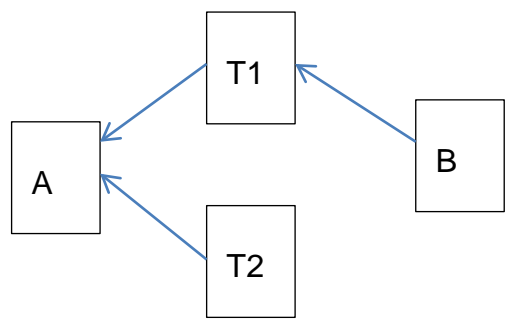

If   A, T1, T2 and B are webpages

Suppose T1 and T2 link page A and B link to Page T1.

For computing rank of A we have,

PR(A)= (1-d) +d $\left(\frac{PR(T1)}{C(T1)} + \frac{PR(T2)}{C(T2)}\right)$ ……..eq. 1

Where, d= damping factor (Due to the fact that user may randomly select or navigate webpages backling)

C(A) is defined as number of links going out of page.

Value of d= 0.85 (can be from 0 to 1)

For PR(A),

From eq. 1 we have,

PR(B)= (1-0.85)+.85 (0) = 0.15

PR(T1)= (1-0.85)+0.85(0.15/1)= 0.27

PR(T2)= (1-0.85)+0.85(0)= 0.15

Hence replacing the value in eq. 1 we get,

PR(A)= (1-0.85)+0.85(0.27+0.15)= 0.5133

Proof,

As stated by (Brin and Page, 1998) sum of all the links is equal to 1

PR(A)+ PR(T1)+ PR(T2)+ PR(B)= 0.5133+0.27+0.15+0.15 = 1

# Appendix 4: Web Mining Data

## 1. Google SERP Analysis

Keyword: alcoholism

1. http://alcoholism.about.com/

| Meta tags report for: http://alcoholism.about.com/ | | |
|---|---|---|
| **meta tag** | **length** | **value** |
| Title: | 24 | The Alcoholism Home Page |
| Description: | 149 | The starting place to find information, resources and the latest news about alcohol, alcoholism, substance abuse and recovery issues on the Internet. |
| Keywords: | 203 | alcoholism alcohol effects problem stop treatment centres 12 step meeting recovery families support college binge drinking teens drunk driving women alcohol withdrawal symptoms treatment help drug abuse |
| Robots: | 5 | NOODP |
| **Headers returned from: http://alcoholism.about.com/** | | |
| Related Keyword found (number of times and density): | | **alcoholism** - 19 - 2.53%<br>**alcohol** - 6 - 0.80%<br>**alcoholism search** - 1 - 0.13%<br>**alcoholism symptoms** - 3 - 0.40%<br>**alcoholism help** - 2 - 0.27%<br>**health alcoholism** - 2 - 0.27%<br>**alcoholism 101** - 2 - 0.27%<br>**free alcoholism** - 2 - 0.27%<br>**alcoholism withdrawal** - 2 - 0.27%<br>**alcohol do** - 2 - 0.27%<br>**alcoholism newsletter** - 2 - 0.27%<br>**10 alcohol** - 1 - 0.13%<br>**alcoholism most** - 1 - 0.13%<br>**alcohol drugs** - 1 - 0.13%<br>**alcoholism must** - 1 - 0.13%<br>**alcohol alcoholism** - 1 - 0.13%<br>**alcoholism search** - 1 - 0.13%<br>**explore alcoholism** - 1 - 0.13%<br>**alcoholism forum** - 1 - 0.13%<br>**category alcoholism** - 1 - 0.13%<br>**alcoholism screening** - 1 - 0.13%<br>**search alcoholism** - 1 - 0.13%<br>**popular alcoholism** - 1 - 0.13%<br>**alcoholism about.com** - 1 - 0.13%<br>**quiz alcohol** - 1 - 0.13%<br>**discussion alcoholism** - 1 - 0.13%<br>**topic alcoholism** - 1 - 0.13%<br>**alcohol withdrawal** - 1 - 0.13%<br>**alcoholism guide** - 1 - 0.13%<br>**alcoholism spotlight** - 1 - 0.13%<br>**free alcoholism newsletter** - 2 - 0.27%<br>**alcoholism withdrawal symptoms** - 2 - 0.27%<br>**alcoholism newsletter sign** - 2 - 0.27%<br>**alcohol withdrawal symptom** - 1 - 0.13% |

| | |
|---|---|
| | **about.com health alcoholism** - 1 - 0.13%<br>**alcohol alcoholism about.com** - 1 - 0.13%<br>**alcoholism must reads** - 1 - 0.13%<br>**most popular alcoholism** - 1 - 0.13%<br>**alcoholism search alcoholism** - 1 - 0.13%<br>**symptom quiz alcohol** - 1 - 0.13%<br>**alcoholism symptoms diagnosis** - 1 - 0.13%<br>**asked questions alcoholism** - 1 - 0.13%<br>**explore alcoholism must** - 1 - 0.13%<br>**alcoholism most popular** - 1 - 0.13%<br>**alcoholism forum photo** - 1 - 0.13%<br>**alcoholism spotlight 10** - 1 - 0.13%<br>**more free alcoholism** - 1 - 0.13%<br>**alcoholism about.com health** - 1 - 0.13%<br>**health alcoholism most** - 1 - 0.13%<br>**topic alcoholism 101** - 1 - 0.13%<br>**alcoholism screening quiz** - 1 - 0.13%<br>**marijuana free alcoholism** - 1 - 0.13%<br>**questions alcoholism spotlight** - 1 - 0.13%<br>**search alcoholism symptoms** - 1 - 0.13%<br>**quiz alcohol withdrawal** - 1 - 0.13%<br>**videos explore alcoholism** - 1 - 0.13%<br>**popular alcoholism screening** - 1 - 0.13%<br>**health alcoholism search** - 1 - 0.13%<br>**alcoholism guide sign** - 1 - 0.13%<br>**discussion alcoholism forum** - 1 - 0.13%<br>**spotlight 10 alcohol** - 1 - 0.13%<br>**category alcoholism 101** - 1 - 0.13% |
| x-meta-robots: | NOODP |
| connection: | Keep-Alive |
| cache-control: | max-age=-3600 |
| set-cookie: | TMog=C54IFt0h20SA0[Ee; domain=.about.com; path=/; expires=Sat, 10-Aug-13 17:22:36 GMT |
| date: | Fri, 04 May 2012 18:15:57 GMT |
| x-meta-keywords: | alcoholism alcohol effects problem stop treatment centers 12 step meeting recovery families support college binge drinking teens drunk driving women alcohol withdrawal symptoms treatment help drug abuse |
| vary: | * |
| client-peer: | 207.126.123.20:80 |
| client-date: | Fri, 04 May 2012 18:15:54 GMT |
| content-type: | text/html |
| pragma: | no-cache |
| client-transfer-encoding: | chunked |
| server: | Apache |
| x-meta-description: | The starting place to find information, resources and the latest news about alcohol, alcoholism, substance abuse and recovery issues on the Internet. |

| link: | ; rel="icon" |
|---|---|
| pics-label: | (pics-1.1 "http://www.icra.org/pics/vocabularyv03/" l gen true for "http://alcoholism.about.com" r (n 0 s 0 v 0 l 0 oa 0 ob 0 oc 0 od 0 oe 0 of 0 og 0 oh 0 c 0) gen true for "http://alcoholism.about.com" r (n 0 s 0 v 0 l 0 oa 0 ob 0 oc 0 od 0 oe 0 of 0 og 0 oh 0 c 0)) |
| keep-alive: | timeout=15, max=23 |
| client-response-num: | 1 |
| x-meta-pd: | Friday, 04-May-2012 10:04:12 UTC |
| x-meta-docset: | 6 |
| p3p: | CP="IDC DSP COR DEVa TAIa OUR BUS UNI" |
| x-ua-compatible: | IE=edge,chrome=1 |
| title: | The Alcoholism Home Page |
| expires: | Fri, 04 May 2012 17:15:57 GMT |

**Keywords found in the Anchor tags: Keyword - Times Found.**
These are text links on web page (include the 'alt' text from images in the links).

**alcoholism - 4**

**symptoms of alcoholism - 2**

**alcoholism 101 - 2**

**effects of alcohol - 2**

**treatment of alcoholism - 2**

**women and alcohol - 2**

**what is alcoholism - 2**

**alcohol and drugs in the news - 1**

**alcoholism forum - 1**

**alcoholism screening quiz – 1**

**Keywords found in the IMG Alt tags: Keyword - Times Found.**
This is text found in the 'alt' tag from the images. For web pages with a lot of images those tags are important (for best results try to name them after your primary keywords).

**about.com - 2**

photoxpress.com - 1

this website is accredited by health on the net foundation click to verify - 1

bigstockphoto.com - 1

what to expect at a first aa meeting - 1

facts about drug addiction - 1

causes and effects of cirrhosis of the liver - 1

buddy t - 1

| URLs found in the page: | URL - Times | Found |
|---|---|---|
| Found 142 urls from where 95 unique. | | |
| http://alcoholism.about.com/ | - | 5 |
| http://alcoholism.about.com/od/drugs/Commonly_Abused_Drugs.htm | - | 3 |
| http://alcoholism.about.com/cs/drugs/a/aa030426a.htm | - | 3 |
| http://alcoholism.about.com/bio/Buddy-T-37.htm | - | 3 |
| http://video.about.com/alcoholism/Drug-Addiction.htm | - | 3 |
| http://www.about.com/ | - | 3 |
| http://video.about.com/alcoholism/What-to-Expect-at-AA-Meetings.htm | - | 3 |
| http://video.about.com/alcoholism/Cirrhosis-of-Liver.htm | - | 3 |
| http://alcoholism.about.com/od/college/College_Binge_Drinking_Issues.htm | - | 2 |
| http://alcoholism.about.com/od/women/Women_and_Substance_Abuse.htm | - | 2 |
| http://alcoholism.about.com/od/support/How_to_Stop_Alcohol_and_Drugs.htm | - | 2 |
| http://alcoholism.about.com/od/problem/Do_I_Have_A_Problem_.htm | - | 2 |
| http://alcoholism.about.com/od/ecstasy/ig/ecstasy/index.htm | - | 2 |
| http://alcoholism.about.com/od/meetings/Find_Support_Group_Meetings.htm | - | 2 |
| http://alcoholism.about.com/od/about/a/alcoholism.htm | - | 2 |
| http://alcoholism.about.com/od/coke/ig/cocaine/index.htm | - | 2 |
| http://alcoholism.about.com/od/pro/Alcohol_and_Drug_Treatment_and_Rehab_Centers.htm | - | 2 |
| http://alcoholism.about.com/od/about/Alcoholism_101.htm | - | 2 |
| http://alcoholism.about.com/od/teens/Teens_and_Substance_Abuse.htm | - | 2 |
| http://alcoholism.about.com/gi/pages/stay.htm | - | 2 |
| http://alcoholism.about.com/od/withdraw/a/withdrawal_fear.htm | - | 2 |
| http://alcoholism.about.com/od/about/a/symptoms.htm | - | 2 |
| http://alcoholism.about.com/cs/info2/a/blfam.htm | - | 2 |
| http://www.hon.ch/HONcode/Conduct.html?HONConduct276829 | - | 2 |
| http://alcoholism.about.com/od/effect/The_Effects_of_Alcohol_and_Drugs.htm | - | 2 |
| http://alcoholism.about.com/od/fam/Support_for_Families_and_Friends.htm | - | 2 |
| http://alcoholism.about.com/od/about/a/treatment.htm | - | 2 |
| http://alcoholism.about.com/od/sa/Drug_Abuse_Information.htm | - | 2 |
| http://alcoholism.about.com/od/dui/Information_about_Drunk_Driving_DUI_DWI_and_OWI.htm | - | 2 |
| http://alcoholism.about.com/b/2012/05/03/teen-prescription-drug-use-unchanged.htm | - | 2 |
| http://alcoholism.about.com/od/etc/12_Step_Recovery.htm | - | 2 |

| URL | | |
|---|:-:|:-:|
| http://alcoholism.about.com/od/binge/Binge_Drinking.htm | - | 2 |
| http://www.about.com/health/ | - | 2 |
| http://alcoholism.about.com/b/2012/05/01/safety-warning-issued-for-pain-patches.htm | - | 2 |
| http://alcoholism.about.com/b/ | - | 2 |
| http://alcoholism.about.com/cs/news/a/drugnews.htm | - | 2 |
| http://alcoholism.about.com/b/2012/05/02/heavy-marijuana-use-by-teens-surges.htm | - | 2 |
| http://0.tqn.com/6/g/alcoholism/b/rss2.xml | | 1 |
| http://www.advertiseonabout.com/ | - | 1 |
| http://www.about.com/gi/pages/hc.htm | - | 1 |
| http://caloriecount.about.com/ | | 1 |
| http://alcoholism.about.com/b/2012/05/04/alcohol-abuse-in-the-news.htm#gB3 | - | 1 |
| http://alcoholism.about.com/cs/faq/a/bldrugfaq.htm | | 1 |
| http://video.about.com/ | - | 1 |
| http://alcoholism.about.com/b/2012/05/04/alcohol-abuse-in-the-news.htm | | 1 |
| http://alcoholism.about.com/gi/pages/shareurl.htm?PG=http%3a%2f%2falcoholism%2eabout%2ecom%2fb%2f2012%2f05%2f02%2fheavy%2dmarijuana%2duse%2dby%2dteens%2dsurges%2ehtm&zItl=%27Heavy%27%20Marijuana%20Use%20by%20Teens%20Surges | - | 1 |
| http://alcoholism.about.com/gi/pages/shareurl.htm?PG=http%3a%2f%2falcoholism%2eabout%2ecom%2fb%2f2012%2f05%2f04%2falcohol%2dabuse%2din%2dthe%2dnews%2ehtm&zItl=Alcohol%20and%20Drugs%20in%20the%20News | - | 1 |
| http://alcoholism.about.com/od/sa/a/blnida041129.htm | - | 1 |
| http://alcoholism.about.com/popular.htm | - | 1 |
| http://alcoholism.about.com/od/tests/l/blquiz_alcohol.htm | - | 1 |
| http://alcoholism.about.com/od/support/u/help.htm | - | 1 |
| http://www.about.com/gi/pages/patent.htm | - | 1 |
| http://jobs.about.com/ | - | 1 |
| http://www.advertiseonabout.com/category/press-releases/ | - | 1 |
| http://alcoholism.about.com/library/blwithdrawalquiz.htm | - | 1 |
| http://0.tqn.com/0/dc/s118.css | - | 1 |
| http://0.tqn.com/f/a08.ico | - | 1 |
| http://www.about.com/gi/pages/mprivacy.htm | - | 1 |
| http://spiderbites.about.com/sitemap.htm | - | 1 |
| http://www.advertiseonabout.com/about-us/ | - | 1 |
| http://www.hon.ch/HONcode/Conduct.html | - | 1 |
| http://alcoholism.about.com/library/blmarijuanaquiz.htm | - | 1 |
| http://forums.about.com/n/pfx/forum.aspx?nav=messages&webtag=ab-alcoholism | - | 1 |
| http://alcoholism.about.com/od/drugs/a/drug_pictures.htm | - | 1 |
| http://forums.about.com/ab-alcoholism/start/ | - | 1 |
| http://alcoholism.about.com/od/effect/u/Risks.htm | - | 1 |
| http://0.tqn.com/4/o/os.xml | - | 1 |
| http://alcoholism.about.com/gi/pages/stay.htm#rs | - | 1 |
| http://alcoholism.about.com/od/pot/a/marijuana_test.htm | - | 1 |
| http://alcoholism.about.com/gi/pages/shareurl.htm?PG=http%3a%2f%2falcoholism%2eabout%2ecom%2fb%2f2012%2f05%2f03%2fteen%2dprescription%2ddrug%2duse%2dunchanged%2ehtm&zItl=Teen%20Prescription%20Drug%20Use%20Unchanged | - | 1 |
| http://www.about.com/gi/pages/ethics.htm | - | 1 |
| http://www.about.com/gi/pages/uagree.htm | - | 1 |
| http://alcoholism.about.com/od/pot/a/effects.-Lya.htm | - | 1 |
| http://alcoholism.about.com/b/2012/05/01/safety-warning-issued-for-pain-patches.htm#gB3 | - | 1 |

http://www.nytco.com/ - 1
http://www.about.com/gi/pages/mprivacy.htm#adchoices - 1
http://alcoholism.about.com/gi/pages/shareurl.htm?PG=http%3a%2f%2falcoholism%2eabout%2ecom%2fb%2f2012%2f05%2f01%2fsafety%2dwarning%2dissued%2dfor%2dpain%2dpatches%2ehtm&zItl=Safety%20Warning%20Issued%20for%20Pain%20Patches - 1
http://0.tqn.com/8/dc/rdie.css - 1
http://ad.doubleclick.net/jump/abt.health/health_alcoholism;kw=;site=alcoholism;chan=health;pos=lb;sz=728x90;ord=1C54IFt0h20SA0[Ee - 1
http://alcoholism.about.com/od/drugs/a/nsduh_drugs.htm - 1
http://alcoholism.about.com/b/2012/05/02/heavy-marijuana-use-by-teens-surges.htm#gB3 - 1
http://alcoholism.about.com/updated.htm - 1
http://azlist.about.com/a.htm - 1
http://alcoholism.about.com/b/2012/05/03/teen-prescription-drug-use-unchanged.htm#gB3 - 1
http://alcoholism.about.com/cs/drugs/a/aa030425a.htm - 1
http://beaguide.about.com/ - 1
http://www.about.com/gi/pages/printrequests.html - 1
http://forums.about.com/ab-alcoholism/start/?lgnF=y - 1
http://alcoholism.about.com/od/about/u/symptoms.htm - 1
http://video.about.com/health.htm - 1
http://alcoholism.about.com/z/js/o'+(p?p:'')+'.htm'?k='+zUriS(t.toLowerCase())+(this.zobr?zobr:'')+'&d='+zUriS(t)+'&r='+zUriS(zWl)+' - 1
http://alcoholism.about.com/gi/pages/shareurl.htm?PG=http%3a%2f%2falcoholism%2eabout%2ecom%2f&zItl=The%20Alcoholism%20Home%26%23160%3bPage - 1
http://www.about.com/health/review.htm - 1
http://alcoholism.about.com/cs/drugs/f/drug_faq01.htm - 1
http://0.tqn.com/4/o/ws.htm?sdn=alcoholism&ws=landingpage_spotlight - 1

2. http://alcoholism.about.com/od/about/a/symptoms.htm

| Meta tags report for: http://alcoholism.about.com/od/about/a/symptoms.htm | | |
|---|---|---|
| meta tag | length | value |
| Title: | 79 | Alcoholism Symptoms - Signs and Symptoms of Alcoholism - Alcohol Abuse Symptoms |
| Description: | 161 | There are many signs and symptoms of drinking problems, but there are seven major symptoms of the most severe drinking problem, alcohol dependence or alcoholism. |
| Keywords: | 118 | alcoholism alcohol dependence severe drinking problems symptoms signs severe alcohol abuse alcoholism symptoms effects |
| Robots: | 5 | NOODP |

| **Headers returned from: http://alcoholism.about.com/od/about/a/symptoms.htm** | |
|---|---|
| Related Keyword found<br><br>(number of times and density): | alcohol - 54 - 6.19%<br>alcoholism - 39 - 4.47%<br>alcohol abuse - 17 - 1.95%<br>alcohol dependence - 10 - 1.15%<br>alcohol withdrawal - 7 - 0.80%<br>alcoholism symptoms - 6 - 0.69%<br>alcoholism screening - 4 - 0.46%<br>abuse alcohol - 3 - 0.34%<br>symptoms alcohol - 3 - 0.34%<br>alcoholism alcoholism - 3 - 0.34%<br>quiz alcohol - 3 - 0.34%<br>drinking alcohol - 3 - 0.34%<br>dependence alcohol - 3 - 0.34%<br>alcohol problems - 2 - 0.23%<br>alcohol symptoms - 2 - 0.23%<br>health alcoholism - 2 - 0.23%<br>alcoholism include - 2 - 0.23%<br>alcoholism 101 - 2 - 0.23%<br>problems alcoholism - 2 - 0.23%<br>alcohol do - 2 - 0.23%<br>alcoholism screening quiz - 4 - 0.46%<br>alcohol abuse alcohol - 3 - 0.34%<br>alcohol withdrawal symptoms - 3 - 0.34%<br>quiz alcohol withdrawal - 3 - 0.34%<br>alcohol dependence alcohol - 3 - 0.34%<br>alcohol withdrawal symptom - 2 - 0.23%<br>dependence alcohol withdrawal - 2 - 0.23%<br>symptoms alcohol abuse - 2 - 0.23%<br>alcoholism symptoms alcohol - 2 - 0.23%<br>abuse alcohol dependence - 2 - 0.23%<br>alcoholism symptoms diagnosis - 2 - 0.23%<br>alcohol abuse diagnosis - 2 - 0.23%<br>screening quiz alcohol - 2 - 0.23%<br>alcoholism symptoms signs - 1 - 0.11%<br>alcoholism alcohol abuse - 1 - 0.11%<br>words alcohol abuse - 1 - 0.11%<br>health alcoholism alcoholism - 1 - 0.11%<br>alcoholics - 3 - 0.34%<br>alcoholism 101 alcoholism - 1 - 0.11%<br>alcohol abuse vs - 1 - 0.11%<br>of alcoholism - 1 - 0.11%<br>alcoholic - 1 - 0.11%<br>vs alcohol - 1 - 0.11%<br>alcoholism international - 1 - 0.11% |
| x-meta-robots: | NOODP |
| connection: | Keep-Alive |
| cache-control: | max-age=-3600 |
| set-cookie: | TMog=C54Icu0C20SA1[IM; domain=.about.com; path=/; expires=Sat, 10-Aug-13 17:47:37 GMT |
| date: | Fri, 04 May 2012 18:40:58 GMT |
| x-meta-keywords: | alcoholism alcohol dependence severe drinking problems symptoms signs severe alcohol abuse alcoholism symptoms effects |
| vary: | * |
| client-peer: | 207.126.123.20:80 |
| client-date: | Fri, 04 May 2012 18:40:55 GMT |
| content-type: | text/html |
| pragma: | no-cache |
| client-transfer-encoding: | chunked |
| server: | Apache |
| x-meta-description: | There are many signs and symptoms of drinking problems, but there are seven major symptoms of the most severe drinking problem, alcohol dependence or alcoholism. |
| link: | ; rel="icon" |
| pics-label: | (pics-1.1 "http://www.icra.org/pics/vocabularyv03/" l gen true for "http://alcoholism.about.com" r (n 0 s 0 v 0 1 0 oa 0 ob 0 oc 0 od 0 oe 0 of 0 og 0 oh 0 c 0) gen true for "http://alcoholism.about.com" r (n 0 s 0 v 0 1 0 oa 0 ob 0 oc 0 od 0 oe 0 of 0 og 0 oh 0 c 0)) |
| keep-alive: | timeout=15, max=88 |
| client-response-num: | 1 |
| x-meta-pd: | Friday, 13-Apr-2012 00:23:44 UTC |
| p3p: | CP="IDC DSP COR DEVa TAIa OUR BUS UNI" |
| x-ua-compatible: | IE=edge,chrome=1 |
| title: | Alcoholism Symptoms - Signs and Symptoms of Alcoholism - Alcohol Abuse Symptoms |
| expires: | Fri, 04 May 2012 17:40:58 GMT |

**Keywords found in the Anchor tags: Keyword - Times Found.** These are text links on your web page (include the 'alt' text from images in the links). These become more important by many search engines (for best results try to name them after your primary keywords).

**alcoholism** - 4

**alcoholism screening quiz** - 4

**alcoholism 101** - 2

**how to talk to your teen about alcohol** - 2

**treatment of alcoholism** - 2

**alcohol abuse** - 2

alcoholism symptoms - 2

alcohol withdrawal - 2

what is alcoholism - 2

alcohol dependence - 2

treatment for alcohol withdrawal symptoms - 1

top myths about drinking alcohol - 1

symptoms of alcoholism - 1

do only alcoholics experience problems from alcohol - 1

seven symptoms of alcoholism - 1

effects of alcohol - 1

alcohol withdrawal symptoms - 1

signs of alcoholism - 1

what do we mean by alcoholism - 1

types of alcohol problems - 1

effects of drinking alcohol - 1

symptoms of alcohol abuse - 1

about drug abuse - 1

frequently asked questions about alcohol abuse and alcoholism - 1

diagnosis of alcohol dependence - 1

alcohol getting the facts about alcohol and alcohol abuse - 1

alcoholism what is alcoholism what is alcohol dependence what is alco - 1

alcohol withdrawal symptom quiz - 1

what is alcohol abuse - 1

alcoholism symptoms diagnosis - 1

alcohol abuse and alcohol dependence what is the difference between alco - 1

women and alcohol - 1

alcohol abuse vs alcohol dependence - 1

early signs of alcoholism - 1

alcohol withdrawal symptoms quiz - 1

diagnosis of alcoholism - 1

**Keywords found in the IMG Alt tags:** Keyword - Times Found.
This is text found in the 'alt' tag from the images. For web pages with a lot of images those tags are important (for best results try to name them after your primary keywords).

**about.com** - 2

**causes and effects of cirrhosis of the liver** - 1

**how to talk to your teen about alcohol** - 1

**buddy t** - 1

**this website is accredited by health on the net foundation click to verify** - 1

**what to expect at a first aa meeting** - 1

| URLs found in the page: URL - Times Found | | |
|---|---|---|
| Found 280 urls from where 225 unique. | | |

| URL | | Times Found |
|---|---|---|
| http://www.patient.co.uk/forums/ | - | 23 |
| http://www.patient.co.uk/health/Alcoholism-and-Problem-Drinking/professional | - | 4 |
| http://www.patient.co.uk/health/Alcoholism-and-Problem-Drinking/resources | - | 4 |
| http://www.patient.co.uk/patientplus.asp | - | 3 |
| http://www.patient.co.uk/support-directory.asp | - | 3 |
| http://www.patient.co.uk/pils.asp | - | 3 |
| http://www.patient.co.uk/mobile.asp | - | 3 |
| http://www.patient.co.uk/ | - | 3 |
| - | | 2 |
| http://www.patient.co.uk/forums | - | 2 |
| http://www.patient.co.uk/news.asp | - | 2 |
| http://www.patient.co.uk/showdoc/22 | - | 2 |
| http://www.patient.co.uk/patient-access.asp | - | 2 |
| http://www.patient.co.uk/editorial_disclaimer.asp | - | 2 |
| http://www.patient.co.uk/textbooks.asp | - | 2 |
| http://www.patient.co.uk/pdf/pilsL580.pdf | - | 2 |
| http://www.patient.co.uk/health/Alcohol-and-Liver-Disease.htm | - | 2 |
| http://www.patient.co.uk/health/http%3A%2F%2Fwww%2Epatient%2Eco%2Euk%2Fhealth%2FAlcoholism%2Dand%2DProblem%2DDrinking%2Ehtm&send=false&layout=button_count&width=80&show_faces=false&action=like&colorscheme=light&font&height=21&appId=157980420978393" | | 2 |
| http://www.patient.co.uk/doctor/Acute-Alcohol-Withdrawal-and-Delirium-Tremens.htm | - | 2 |
| http://www.patient.co.uk/health/Alcohol-and-Sensible-Drinking.htm | - | 2 |
| http://www.patient.co.uk/printer.asp?doc=27000115 | - | 2 |
| http://www.patient.co.uk/feedback.asp | - | 2 |
| http://www.patient.co.uk/disclaimer.asp | - | 2 |
| http://www.patient.co.uk/doctor/Alcohol-Related-Problems.htm | - | 2 |
| http://www.patient.co.uk/dils.asp | - | 2 |
| http://www.patient.co.uk/displayCategory/16777279 | - | 1 |
| http://www.patient.co.uk/books.asp | - | 1 |
| http://www.patient.co.uk/displayCategory/16777422 | - | 1 |
| http://www.emis-online.com/about-us/careers/ | - | 1 |
| http://www.patient.co.uk/offers.asp | - | 1 |

| URL | | |
|---|---|---|
| http://www.ncbi.nlm.nih.gov/entrez/query.fcgi?cmd=Retrieve&db=PubMed&dopt=Abstract&list_uids=19141482 | - | 1 |
| http://www.patient.co.uk/displayCategory/16777231 | | 1 |
| http://www.ncbi.nlm.nih.gov/entrez/query.fcgi?cmd=Retrieve&db=PubMed&dopt=Abstract&list_uids=19168210 | - | 1 |
| http://www.patient.co.uk/displayCategory/16777226 | - | 1 |
| http://www.patient.co.uk/displayCategory/16777274 | - | 1 |
| http://www.patient.co.uk/drugs/i.htm | - | 1 |
| http://www.patient.co.uk/fancybox/jquery.fancybox.css | - | 1 |
| http://www.patient.co.uk/health/Vaginal-Thrush.htm | - | 1 |
| http://www.patient.co.uk/health/Alcoholism-and-Problem-Drinking.htm#section_5 | - | 1 |
| http://www.patient.co.uk/patientplus/p.htm | - | 1 |
| http://www.patient.co.uk/drugs/a.htm | - | 1 |
| http://www.patient.co.uk/health/http%3A%2F%2Fwww.facebook.com%2Fpatientuk&width=140&height=612&colorscheme=light&show_faces=true&border_color&stream=false&header=true&appId=157980420978393" | - | 1 |
| http://www.patient.co.uk/showdoc/50000004 | - | 1 |
| http://www.patient.co.uk/illness/x.htm | - | 1 |
| http://www.patient.co.uk/health/blogs/sarah/ | - | 1 |
| http://www.patient.co.uk/illness/a.htm | - | 1 |
| http://www.patient.co.uk/drugs/r.htm | - | 1 |
| http://www.patient.co.uk/patientplus/g.htm | - | 1 |
| http://www.patient.co.uk/css/print.css | - | 1 |
| http://www.patient.co.uk/showdoc/50000001 | - | 1 |
| http://www.patient.co.uk/patientplus/k.htm | - | 1 |
| http://www.patient.co.uk/drugs/d.htm | - | 1 |
| http://guidance.nice.org.uk/CG100 | - | 1 |
| http://www.nacoa.org.uk | - | 1 |
| http://www.patient.co.uk/patientplus/m.htm | - | 1 |
| http://www.patient.co.uk/displayCategory/16777230 | - | 1 |
| http://www.patient.co.uk/patientplus/c.htm | - | 1 |
| http://www.patient.co.uk/patientplus/b.htm | - | 1 |
| http://www.patient.co.uk/patientplus/x.htm | - | 1 |
| http://www.patient.co.uk/displayCategory/16777234 | - | 1 |
| http://www.patient.co.uk/health/Irritable-Bowel-Syndrome.htm | - | 1 |
| http://www.patient.co.uk/displayCategory/16777458 | - | 1 |
| http://www.patient.co.uk/press.asp | - | 1 |
| http://fluidsurveys.com | - | 1 |
| http://www.patient.co.uk/drugs/x.htm | - | 1 |
| https://www.healthonnet.org/HONcode/Conduct.html?HONConduct950113 | - | 1 |
| http://www.patient.co.uk/displayCategory/16777223 | - | 1 |
| http://www.drinkaware.co.uk | - | 1 |
| http://www.patient.co.uk/showdoc/50000011 | - | 1 |
| http://www.patient.co.uk/showdoc/50000007 | - | 1 |
| http://www.patient.co.uk/displayCategory/16777460 | - | 1 |
| http://www.patient.co.uk/patientplus/i.htm | - | 1 |
| http://www.patient.co.uk/patientplus/d.htm | - | 1 |
| http://www.patient.co.uk/health/Alcoholism-and-Problem-Drinking.htm#section_2 | - | 1 |
| http://www.patient.co.uk/showdoc/50000019 | - | 1 |
| http://www.patient.co.uk/sitemap.asp | - | 1 |
| http://itunes.apple.com/gb/app/patient.co.uk/id492677470?mt=8&ls=1# | - | 1 |
| https://patient.emisaccess.co.uk | - | 1 |
| http://www.ncbi.nlm.nih.gov/entrez/query.fcgi?cmd=Retrieve&db=PubMed&dopt=Abstract&list_uids=15301399 | - | 1 |
| http://www.patient.co.uk/displayCategory/16777416 | - | 1 |
| http://www.patient.co.uk/showdoc/50000022 | - | 1 |
| http://www.al-anonuk.org.uk | - | 1 |
| http://www.patient.co.uk/awards.asp | - | 1 |
| http://www.patient.co.uk/showdoc/50000013 | - | 1 |
| http://www.patient.co.uk/displayCategory/16777233 | - | 1 |
| http://www.patient.co.uk/illness/m.htm | - | 1 |
| http://www.patient.co.uk/drugs/g.htm | - | 1 |
| http://www.patient.co.uk/health/Alcoholism-and-Problem-Drinking.htm#section_8 | - | 1 |
| http://www.patient.co.uk/showdoc/50000009 | - | 1 |
| http://www.patient.co.uk/illness/i.htm | - | 1 |
| http://www.patient.co.uk/displayCategory/16777278 | - | 1 |
| http://www.patient.co.uk/showdoc/50000010 | - | 1 |
| http://www.patient.co.uk/showdoc/50000016 | - | 1 |
| http://www.patient.co.uk/displayCategory/16777462 | - | 1 |
| http://www.patient.co.uk/drugs/o.htm | - | 1 |
| https://twitter.com/patientuk | - | 1 |
| http://www.patient.co.uk/illness/f.htm | - | 1 |
| http://www.patient.co.uk/patientplus/u.htm | - | 1 |
| http://www.patient.co.uk/displayCategory/16777225 | - | 1 |
| http://www.patient.co.uk/patientplus/h.htm | - | 1 |
| http://www.patient.co.uk/favicon.ico | - | 1 |
| http://www.patient.co.uk/drugs/j.htm | - | 1 |
| http://www.patient.co.uk/drugs/v.htm | - | 1 |
| http://www.patient.co.uk/health/http%3A%2F%2Fwww%2Epatient%2Eco%2Euk&send=false&layout=button_count&width=80&show_faces=false&action=like&colorscheme=light&font&height=21&appId=157980420978393" | - | 1 |
| http://www.patient.co.uk/patientplus/l.htm | - | 1 |
| http://www.patient.co.uk/showdoc/50000023 | - | 1 |
| http://www.patient.co.uk/illness/t.htm | - | 1 |
| http://www.patient.co.uk/illness/l.htm | - | 1 |
| http://www.patient.co.uk/health/Alcoholism-and-Problem-Drinking.htm | - | 1 |
| http://www.patient.co.uk/showdoc/19 | - | 1 |
| http://www.patient.co.uk/illness/k.htm | - | 1 |
| http://www.patient.co.uk/shop-index.asp | - | 1 |
| http://www.patient.co.uk/displayCategory/16777236 | - | 1 |
| http://www.patient.co.uk/showdoc/50000005 | - | 1 |
| http://www.patient.co.uk/displayCategory/16777275 | - | 1 |
| http://www.alcoholics-anonymous.org.uk | - | 1 |
| http://www.patient.co.uk/showdoc/50000020 | - | 1 |
| http://www.patient.co.uk/adhow.asp | - | 1 |
| http://www.patient.co.uk/showdoc/1097 | - | 1 |
| http://www.patient.co.uk/patientplus/n.htm | - | 1 |
| http://www.patient.co.uk/health/Haemorrhoids-(Piles).htm | - | 1 |
| http://www.patient.co.uk/drugs/y.htm | - | 1 |
| http://www.patient.co.uk/health/Alcoholism-and-Problem-Drinking.htm#refdis | - | 1 |
| http://www.patient.co.uk/health/Alcoholism-and-Problem-Drinking.htm#section_4 | - | 1 |
| http://www.patient.co.uk/css/ie6.css | - | 1 |
| http://www.hitwise.com/uk/resources/top-ten-award?action=confirm&market=uk&sDomain=www.patient.co.uk&iDate=2011&iCatnum=407&Cal=1&semi= | - | 1 |
| http://www.patient.co.uk/patientplus/y.htm | - | 1 |
| http://www.patient.co.uk/health/Vitamin-D-Deficiency.htm | - | 1 |
| http://www.patient.co.uk/showdoc/50000003 | - | 1 |
| http://www.patient.co.uk/showdoc/50000014 | - | 1 |
| http://www.patient.co.uk/css/ie7.css | - | 1 |
| http://www.patient.co.uk/pils.asp?tab=requested | - | 1 |
| http://www.patient.co.uk/drugs/n.htm | - | 1 |
| http://www.patient.co.uk/illness/o.htm | - | 1 |
| http://www.patient.co.uk/patientplus/s.htm | - | 1 |
| http://www.theinformationstandard.org/members | - | 1 |
| http://www.patient.co.uk/health/Alcoholism-and-Problem-Drinking.htm#to-content | - | 1 |
| http://www.patient.co.uk/drugs/l.htm | - | 1 |
| http://www.patient.co.uk/displayCategory/16777228 | - | 1 |
| http://www.patient.co.uk/illness/h.htm | - | 1 |
| http://www.patient.co.uk/patientplus/e.htm | - | 1 |
| http://www.ncbi.nlm.nih.gov/entrez/query.fcgi?cmd=Retrieve&db=PubMed&dopt=Abstract&list_uids=21414953 | - | 1 |
| http://www.patient.co.uk/showdoc/50000021 | - | 1 |
| http://www.patient.co.uk/illness/v.htm | - | 1 |
| http://www.patient.co.uk/patientplus/w.htm | - | 1 |
| http://www.patient.co.uk/displayCategory/16777224 | - | 1 |
| http://www.patient.co.uk/guidelines.asp | - | 1 |
| http://www.patient.co.uk/displayCategory/16777235 | - | 1 |
| http://www.patient.co.uk/drugs/m.htm | - | 1 |
| http://www.patient.co.uk/drugs/z.htm | - | 1 |
| http://www.patient.co.uk/health/Alcoholism-and-Problem-Drinking.htm#section_6 | - | 1 |
| http://www.patient.co.uk/drugs/h.htm | - | 1 |
| http://www.patient.co.uk/displayCategory/16777415 | - | 1 |
| http://www.patient.co.uk/drugs/u.htm | - | 1 |
| http://www.patient.co.uk/patientplus/t.htm | - | 1 |
| http://www.patient.co.uk/health/Shingles.htm | - | 1 |
| http://www.patient.co.uk/blogs/sarah-says/ | - | 1 |
| http://www.patient.co.uk/showdoc/50000018 | - | 1 |
| http://www.patient.co.uk/drugs/e.htm | - | 1 |
| http://www.patient.co.uk/patientplus/a.htm | - | 1 |
| http://www.dh.gov.uk/en/Publichealth/Healthimprovement/Alcoholmisuse/index.htm | - | 1 |
| http://www.patient.co.uk/illness/p.htm | - | 1 |
| http://www.patient.co.uk/illness/z.htm | - | 1 |
| http://www.patient.co.uk/about.asp | - | 1 |
| http://www.patient.co.uk/displayCategory/16777227 | - | 1 |
| http://www.patient.co.uk/displayCategory/16777276 | - | 1 |
| http://www.patient.co.uk/dils.asp?tab=requested | - | 1 |
| http://www.patient.co.uk/pharmacy.asp | - | 1 |
| http://www.patient.co.uk/health/Alcoholism-and-Problem-Drinking.htm#section_1 | - | 1 |
| http://www.patient.co.uk/showdoc/50000017 | - | 1 |
| http://www.patient.co.uk/PatientEquipment.asp | - | 1 |
| http://www.patient.co.uk/drugs/s.htm | - | 1 |
| http://guidance.nice.org.uk/CG115 | - | 1 |
| http://app.fluidsurveys.com/polls/patientuk/exercise-poll/ | - | 1 |
| http://www.patient.co.uk/blogs/sarah-says | - | 1 |
| http://www.patient.co.uk/css/style.css | - | 1 |
| http://www.patient.co.uk/displayCategory/16777277 | - | 1 |
| http://www.patient.co.uk/displayCategory/16777222 | - | 1 |
| http://www.patient.co.uk/drugs/t.htm | - | 1 |
| http://www.patient.co.uk/drugs/q.htm | - | 1 |
| http://www.patient.co.uk/patientplus/o.htm | - | 1 |
| http://www.patient.co.uk/drugs/b.htm | - | 1 |
| http://www.patient.co.uk/showdoc/50000006 | - | 1 |
| http://www.patient.co.uk/illness/d.htm | - | 1 |
| http://www.patient.co.uk/drugs/p.htm | - | 1 |
| http://www.patient.co.uk/showdoc/50000002 | - | 1 |
| http://www.patient.co.uk/patientplus/r.htm | - | 1 |
| http://www.patient.co.uk/health/Alcoholism-and-Problem-Drinking.htm#section_7 | - | 1 |
| http://www.patient.co.uk/illness/w.htm | - | 1 |
| http://www.patient.co.uk/illness/j.htm | - | 1 |
| http://www.patient.co.uk/drugs/f.htm | - | 1 |
| http://www.patient.co.uk/displayCategory/16777461 | - | 1 |
| http://www.patient.co.uk/drugs/w.htm | - | 1 |
| http://www.patient.co.uk/illness/c.htm | - | 1 |
| http://www.patient.co.uk/illness/e.htm | - | 1 |
| http://www.patient.co.uk/favicon.gif | - | 1 |
| http://www.patient.co.uk/showdoc/50000008 | - | 1 |
| http://www.patient.co.uk/displayCategory/16777319 | - | 1 |
| http://www.ncbi.nlm.nih.gov/entrez/query.fcgi?cmd=Retrieve&db=PubMed&dopt=Abstract&list_uids=20662805 | - | 1 |
| http://www.patient.co.uk/illness/r.htm | - | 1 |
| http://www.patient.co.uk/css/ie.css | - | 1 |
| http://www.patient.co.uk/search.asp | - | 1 |
| http://www.patient.co.uk/displayCategory/16777232 | - | 1 |
| http://www.patient.co.uk/illness/g.htm | - | 1 |
| http://www.patient.co.uk/blogs | - | 1 |
| http://www.patient.co.uk/displayCategory/16777229 | - | 1 |
| http://www.patient.co.uk/displayCategory/16777221 | - | 1 |
| http://www.patient.co.uk/health/Alcoholism-and-Problem-Drinking.htm# | - | 1 |
| http://www.patient.co.uk/drugs/c.htm | - | 1 |
| http://www.patient.co.uk/showdoc/50000015 | - | 1 |
| http://www.patient.co.uk/illness/y.htm | - | 1 |
| http://www.patient.co.uk/illness/u.htm | - | 1 |
| http://www.patient.co.uk/patientplus/f.htm | - | 1 |
| http://www.patient.co.uk/drugs/k.htm | - | 1 |
| http://www.patient.co.uk/patientplus/z.htm | - | 1 |
| http://www.patient.co.uk/showdoc/50000012 | - | 1 |
| http://www.patient.co.uk/illness/b.htm | - | 1 |
| http://www.patient.co.uk/patientplus/j.htm | - | 1 |
| http://www.patient.co.uk/patientplus/q.htm | - | 1 |

http://www.patient.co.uk/illness/s.htm                                    -                                    1
http://www.patient.co.uk/patientplus/v.htm                                    -                                    1
http://www.patient.co.uk/health/Alcoholism-and-Problem-Drinking.htm#section_3                    -                    1
http://www.patient.co.uk/illness/n.htm                                    -                                    1
http://www.patient.co.uk/cookies.asp                                    -                                    1
http://guidance.nice.org.uk/PH24 - 1

1. http://www.patient.co.uk/health/Alcoholism-and-Problem-Drinking.htm

| Meta tags report for: http://www.patient.co.uk/health/Alcoholism-and-Problem-Drinking.htm | | |
|---|---|---|
| meta tag | length | value |
| Title: | 53 | Alcoholism and Problem Drinking \| Health \| Patient UK |
| Description: | 150 | Alcoholism and Problem Drinking - Alcoholism is a word which many people use to mean alcohol dependence (alcohol addiction). Some people are proble... |
| Keywords: | 338 | alcohol problems,alcohol misuse,alcohol-related problems,problem drinking,problem drinker,national association for children of alcoholics,drinkline,drinkaware,alcohol related problems,alcohol problem,alcohol dependence,alcoholism,alcoholics anonymous,alcoholics,alcoholic,alcohol addiction,al-anon,sensible drinking,alcohol intake |

| Headers returned from: http://www.patient.co.uk/health/Alcoholism-and-Problem-Drinking.htm | |
|---|---|
| Related Keyword found (number of times and density): | **acute alcohol withdrawal and delirium tremens** - 2<br>**alcohol and liver disease** - 2<br>**alcohol and sensible drinking** - 2<br>**alcohol related problems** - 2 |
| x-meta-description: | Alcoholism and Problem Drinking - Alcoholism is a word which many people use to mean alcohol dependence (alcohol addiction). Some people are proble... |
| link: | ; /="/"; rel="canonical" |
| x-powered-by: | ASP.NET |
| client-response-num: | 1 |
| set-cookie: | ASPSESSIONIDAACSSQCR=NAHGDLPBKPFMPFNJEDCOAHAO; path=/ |
| cache-control: | no-cache,private |
| date: | Fri, 04 May 2012 18:55:20 GMT |
| x-meta-keywords: | alcohol problems,alcohol misuse,alcohol-related problems,problem drinking,problem drinker,national association for children of alcoholics,drinkline,drinkaware,alcohol related problems,alcohol problem,alcohol dependence,alcoholism,alcoholics anonymous,alcoholics,alcoholic,alcohol addiction,al-anon,sensible drinking,alcohol intake |
| client-peer: | 94.245.91.12:80 |
| content-length: | 69809 |
| client-date: | Fri, 04 May 2012 18:55:20 GMT |
| content-type: | text/html |
| title: | Alcoholism and Problem Drinking | Health | Patient UK |
| server: | Microsoft-IIS/7.5 |

**Keywords found in the IMG Alt tags: Keyword - Times Found.**
This is text found in the 'alt' tag from the images. For web pages with a lot of images those tags are important (for best results try to name them after your primary keywords).

patient information leaflet - 4

professional reference article - 4

hitwise top 10 website - 1

discuss - 1

the information standard certified member - 1

sign in - 1

patient app - 1

scan me qr code - 1

sarah-says-widget - 1

loading discussions - 1

hon code certified - 1

the information standard - 1

4   http://www.patient.co.uk/health/Alcoholism-and-Problem-Drinking.htm

**Meta tags report for: http://www.patient.co.uk/health/Alcoholism-and-Problem-Drinking.htm**

| meta tag | length | value |
|---|---|---|
| Title: | 53 | Alcoholism and Problem Drinking \| Health \| Patient UK |
| Description: | 150 | Alcoholism and Problem Drinking - Alcoholism is a word which many people use to mean alcohol dependence (alcohol addiction). Some people are proble... |
| Keywords: | 338 | alcohol problems,alcohol misuse,alcohol-related problems,problem drinking,problem drinker,national association for children of alcoholics,drinkline,drinkaware,alcohol related problems,alcohol problem,alcohol,alcohol dependence,alcoholism,alcoholics anonymous,alcoholics,alcoholic,alcohol addiction,al-anon,sensible drinking,alcohol intake |

**Headers returned from: http://www.patient.co.uk/health/Alcoholism-and-Problem-Drinking.htm**

Related Keyword found

(number of times and density):

**alcohol** - 95 - 2.89%
**alcohol-dependent** - 9 - 0.27%
**alcohol-related** - 7 - 0.21%
**alcoholics** - 6 - 0.18%
**alcoholism** - 4 - 0.12%
**alcohol-use** - 4 - 0.12%
**alcoholic** - 1 - 0.03%
**www.alcoholics-anonymous.org.uk** - 1 - 0.03%
**alcoholmisuse** - 1 - 0.03%
**low-alcohol** - 1 - 0.03%
**alcohol dependence** - 8 - 0.24%
**off alcohol** - 8 - 0.24%
**drinking alcohol** - 7 - 0.21%
**stop alcohol** - 5 - 0.15%
**alcohol-use disorders** - 4 - 0.12%
**alcohol completely** - 4 - 0.12%
**alcohol withdrawal** - 4 - 0.12%
**alcohol problems** - 3 - 0.09%
**alcohol related** - 3 - 0.09%
**alcohol misuse** - 3 - 0.09%
**alcohol unit** - 2 - 0.06%
**tremens alcohol** - 2 - 0.06%
**alcohol alcohol** - 2 - 0.06%
**alcoholics anonymous** - 2 - 0.06%
**alcohol drinking** - 2 - 0.06%
**alcohol such** - 2 - 0.06%
**more alcohol** - 2 - 0.06%
**alcohol-related condition** - 2 - 0.06%
**disease alcohol** - 2 - 0.06%
**national alcohol** - 2 - 0.06%
**alcohol-related problems** - 2 - 0.06%
**drink alcohol** - 2 - 0.06%
**acute alcohol** - 2 - 0.06%
**alcohol may** - 2 - 0.06%
**alcohol no** - 1 - 0.03%
**alcohol-related morbidity** - 1 - 0.03%
**called alcohol** - 1 - 0.03%
**stopping alcohol** - 1 - 0.03% **alcohol wears** - 1 - 0.03%
**alcohol-dependent especially** - 1 - 0.03%
**daily alcohol** - 1 - 0.03%
**also alcohol** - 1 - 0.03%
**drinking low-alcohol** - 1 - 0.03%
**alcohol craving** - 1 - 0.03%
**alcohol-related harm** - 1 - 0.03%
**ease alcohol** - 1 - 0.03%
**another alcohol-related** - 1 - 0.03%
**harmful alcohol** - 1 - 0.03%
**al alcohol-related** - 1 - 0.03%

**alcoholics through** - 1 - 0.03%
**alcohol helpline** - 1 - 0.03%
**minimise alcohol-related** - 1 - 0.03%
**alcohol intake** - 1 - 0.03%
**alcohol-dependent people** - 1 - 0.03%
**dependence alcohol** - 1 - 0.03%
**alcoholmisuse index.htm** - 1 - 0.03%
**ma alcohol-use** - 1 - 0.03%
**alcohol many** - 1 - 0.03%
**alcohol self-help** - 1 - 0.03%
**alcoholic drinks** - 1 - 0.03%
**heavily alcohol** - 1 - 0.03%
**alcohol often** - 1 - 0.03%
**alcohol sometimes** - 1 - 0.03%
**hospital alcohol** - 1 - 0.03%
**alcohol strategy** - 1 - 0.03%
**atient.co.uk alcohol** - 1 - 0.03%
**alcohol addiction** - 1 - 0.03%
**alcohol they** - 1 - 0.03%
**alcohol return** - 1 - 0.03%
**reduce alcohol** - 1 - 0.03%
**serious alcohol-related** - 1 - 0.03%
**having alcohol** - 1 - 0.03%
**mortality alcohol** - 1 - 0.03%
**mean alcohol** - 1 - 0.03%
**discuss alcoholism** - 1 - 0.03%
**alcohol-related diseases** - 1 - 0.03%
**conditions alcoholism** - 1 - 0.03%
**developing alcohol** - 1 - 0.03%
**alcohol however** - 1 - 0.03%
**alcohol therefore** - 1 - 0.03%
**alcohol even** - 1 - 0.03%
**alcohol res** - 1 - 0.03%
**references alcohol** - 1 - 0.03%
**2010 alcohol-use** - 1 - 0.03%
**randomized alcohol** - 1 - 0.03%
**web www.alcoholics-anonymous.org.uk** - 1 - 0.03%
**alcohol further** - 1 - 0.03%
**low-alcohol beers** - 1 - 0.03%
**alcohol suddenly** - 1 - 0.03%
**alcohol dependent** - 1 - 0.03%
**dependence alcohol-use** - 1 - 0.03%
**alcohol benzodiazepine** - 1 - 0.03%
**alcohol-dependent then** - 1 - 0.03%
**alcohol 2009** - 1 - 0.03%
**alcohol lorazepam** - 1 - 0.03%
**drinking alcoholism** - 1 - 0.03%
**drinking alcohol-use** - 1 - 0.03%
**alcohol regularly** - 1 - 0.03%
**alcoholics national** - 1 - 0.03%
**alcohol takes** - 1 - 0.03%
**alcoholics department** - 1 - 0.03%
**alcoholics po** - 1 - 0.03%
**out alcohol** - 1 - 0.03%
**alcohol disulfiram** - 1 - 0.03%
**alcohol do** - 1 - 0.03%
**illnesses alcohol** - 1 - 0.03%
**healthimprovement alcoholmisuse** - 1 - 0.03%
**help alcoholics** - 1 - 0.03%
**friend alcoholism** - 1 - 0.03%
**2011 alcohol** - 1 - 0.03%
**alcohol 2011** - 1 - 0.03%
**alcohol clin** - 1 - 0.03%
**stay off alcohol** - 6 - 0.18%
**stop drinking alcohol** - 4 - 0.12%
**stop alcohol completely** - 3 - 0.09%
**alcohol related problems** - 3 - 0.09%
**tremens alcohol related** - 2 - 0.06%
**staying off alcohol** - 2 - 0.06%
**liver disease alcohol** - 2 - 0.06%
**alcohol withdrawal symptoms** - 2 - 0.06%
**delirium tremens alcohol** - 2 - 0.06%

**alcohol-use disorders preventing** - 2 - 0.06%
**acute alcohol withdrawal** - 2 - 0.06%
**after stopping alcohol** - 1 - 0.03%
**alcoholics anonymous po** - 1 - 0.03%
**alcohol clin exp** - 1 - 0.03%
**alcohol drinking social** - 1 - 0.03%
**alcoholics po box** - 1 - 0.03%
**off alcohol many** - 1 - 0.03%
**alcohol-related harm al-anon** - 1 - 0.03%
**alcohol problems after** - 1 - 0.03%
**alcohol alcohol 2011** - 1 - 0.03%
**patientplus acute alcohol** - 1 - 0.03%
**feedback references alcohol** - 1 - 0.03%
**alcohol self-help groups** - 1 - 0.03%
**dependence alcohol-use disorders** - 1 - 0.03%
**consider drinking low-alcohol** - 1 - 0.03%
**alcohol misuse references** - 1 - 0.03%
**drin alcohol 2009 mar-apr** - 1 - 0.03%
**king alcohol-use disorders** - 1 - 0.03%
**alcohol-dependent especially during** - 1 - 0.03%
**resources discuss alcoholism** - 1 - 0.03%
**reducing daily alcohol** - 1 - 0.03%
**alcohol-related condition binge** - 1 - 0.03%
**detail acute alcohol** - 1 - 0.03%
**alcohol problems vitamins** - 1 - 0.03%
**drinking low-alcohol beers** - 1 - 0.03%
**stop alcohol suddenly** - 1 - 0.03%
**not having alcohol** - 1 - 0.03%
**alcohol-dependent people however** - 1 - 0.03%
**hospital alcohol alcohol** - 1 - 0.03%
**alcohol res health** - 1 - 0.03%
**national alcohol strategy** - 1 - 0.03%
**alcohol-use disorders lancet** - 1 - 0.03%
**alcohol misuse page** - 1 - 0.03%
**alcohol craving either** - 1 - 0.03%
**alcohol problems talking** - 1 - 0.03%
**need more alcohol** - 1 - 0.03%
**ease alcohol craving** - 1 - 0.03%
**alcohol alcohol alcohol** - 1 - 0.03%
**then stop alcohol** - 1 - 0.03%
**out alcohol completely** - 1 - 0.03%
**alcohol 2011 may-jun** - 1 - 0.03%
**drink heavily alcohol** - 1 - 0.03%
**drink alcohol regularly** - 1 - 0.03%
**dependence alcohol addiction** - 1 - 0.03%
**alcohol often makes** - 1 - 0.03%
**alcohol may persist** - 1 - 0.03%
**see also alcohol** - 1 - 0.03%
**alcohol lorazepam olanzapine** - 1 - 0.03%
**drinking alcohol no** - 1 - 0.03%
**problem drinking alcoholism** - 1 - 0.03%
**drinking alcohol benzodiazepine** - 1 - 0.03%
**harmful drinking alcohol-use** - 1 - 0.03%
**february 2011 alcohol** - 1 - 0.03%
**alcohol may seem** - 1 - 0.03%
**healthimprovement alcoholmisuse index.htm** - 1 - 0.03%
**off alcohol disulfiram** - 1 - 0.03%
**alcohol no one** - 1 - 0.03%
**off alcohol further** - 1 - 0.03%
**cut out alcohol** - 1 - 0.03%
**alcohol-related condition such** - 1 - 0.03%
**alcohol-dependent then detoxification** - 1 - 0.03%
**more alcohol withdrawal** - 1 - 0.03%
**alcohol helpline helpline** - 1 - 0.03%
**heavily alcohol dependence** - 1 - 0.03%
**alcohol alcohol 2009** - 1 - 0.03%
**drinking alcohol takes** - 1 - 0.03%
**another alcohol-related condition** - 1 - 0.03%
**alcoholics national association** - 1 - 0.03%
**illnesses alcohol may** - 1 - 0.03%
**alcohol dependence addiction** - 1 - 0.03%
**randomized alcohol clin** - 1 - 0.03%

| | |
|---|---|
| | **diseases conditions alcoholism** - 1 - 0.03%<br>**alcohol dependent then** - 1 - 0.03%<br>**alcohol they could** - 1 - 0.03%<br>**schuckit ma alcohol-use** - 1 - 0.03%<br>**alcohol however even** - 1 - 0.03%<br>**alcoholmisuse index.htm includes** - 1 - 0.03%<br>**male hospital alcohol** - 1 - 0.03%<br>**daily alcohol intake** - 1 - 0.03%<br>**developing alcohol dependence** - 1 - 0.03%<br>**alcohol further resources** - 1 - 0.03%<br>**ma alcohol-use disorders** - 1 - 0.03%<br>**mortality alcohol res** - 1 - 0.03%<br>**off alcohol lorazepam** - 1 - 0.03%<br>**drinkline national alcohol** - 1 - 0.03%<br>**al alcohol-related morbidity** - 1 - 0.03%<br>**2010 alcohol-use disorders** - 1 - 0.03%<br>**national alcohol helpline** - 1 - 0.03%<br>**alcohol wears off** - 1 - 0.03%<br>**alcoholic drinks consider** - 1 - 0.03%<br>**publichealth healthimprovement alcoholmisuse** - 1 - 0.03%<br>**7555 web www.alcoholics-anonymous.org.uk** - 1 - 0.03%<br>**alcohol many people** - 1 - 0.03%<br>**help alcoholics anonymous** - 1 - 0.03%<br>**et al alcohol-related** - 1 - 0.03%<br>**mean alcohol dependence** - 1 - 0.03%<br>**reduce alcohol misuse** - 1 - 0.03%<br>**alcohol-related diseases among** - 1 - 0.03%<br>**june 2010 alcohol-use** - 1 - 0.03%<br>**serious alcohol-related problems** - 1 - 0.03%<br>**references alcohol dependence** - 1 - 0.03%<br>**alcohol-use disorders physical** - 1 - 0.03%<br>**having alcohol even** - 1 - 0.03%<br>**alcohol unit may** - 1 - 0.03%<br>**alcohol even after** - 1 - 0.03%<br>**other illnesses alcohol** - 1 - 0.03%<br>**off alcohol self-help** - 1 - 0.03%<br>**alcohol dependence alcohol** - 1 - 0.03%<br>**alcohol dependence should** - 1 - 0.03%<br>**alcohol dependence alcohol-use** - 1 - 0.03%<br>**alcohol benzodiazepine medicines** - 1 - 0.03%<br>**not stop alcohol** - 1 - 0.03%<br>**leaflet called alcohol** - 1 - 0.03%<br>**alcohol-related problems may** - 1 - 0.03%<br>**alcohol do not** - 1 - 0.03%<br>**alcohol completely do** - 1 - 0.03%<br>**minimise alcohol-related harm** - 1 - 0.03%<br>**help reduce alcohol** - 1 - 0.03% |
| x-meta-description: | Alcoholism and Problem Drinking - Alcoholism is a word which many people use to mean alcohol dependence (alcohol addiction). Some people are proble... |
| link: | ; /="/"; rel="canonical" |
| x-powered-by: | ASP.NET |
| client-response-num: | 1 |
| set-cookie: | ASPSESSIONIDAACSSQCR=HMKBELPBOJOICFBPAMLINIJI; path=/ |
| cache-control: | no-cache,private |
| date: | Sat, 05 May 2012 10:44:37 GMT |
| x-meta-keywords: | alcohol problems,alcohol misuse,alcohol-related problems,problem drinking,problem drinker,national association for children of alcoholics,drinkline,drinkaware,alcohol related problems,alcohol problem,alcohol,alcohol dependence,alcoholism,alcoholics anonymous,alcoholics,alcoholic,alcohol addiction,al-anon,sensible drinking,alcohol intake |
| client-peer: | 94.245.91.12:80 |
| content-length: | 69809 |
| client-date: | Sat, 05 May 2012 10:44:35 GMT |
| content-type: | text/html |
| title: | Alcoholism and Problem Drinking | Health | Patient UK |
| server: | Microsoft-IIS/7.5 |

| |
|---|
| **Keywords found in the Anchor tags: Keyword - Times Found.** These are text links on your web page (include the 'alt' text from images in the links). These become more important by many search engines (for best results try to name them after your primary keywords). |

acute alcohol withdrawal and delirium tremens - 2

alcohol and liver disease - 2

alcohol and sensible drinking - 2

alcohol related problems - 2

alcohol dependence and harmful alcohol use - 1

www.alcoholics-anonymous.org.uk - 1

should i cut back or should i stop alcohol completely - 1

alcohol-use disorders physical complications - 1

www.dh.gov.uk en publichealth healthimprovement alcoholmisuse index.htm - 1

problems with drinking alcohol - 1

discuss alcoholism and problem drinking in the forums - 1

alcohol-use disorders preventing harmful drinking - 1

what are the symptoms of alcohol dependence - 1

after detoxification and staying off alcohol - 1

other medication sometimes used for alcohol problems - 1

**Keywords found in the IMG Alt tags: Keyword - Times Found.**
This is text found in the 'alt' tag from the images. For web pages with a lot of images those tags are important (for best results try to name them after your primary keywords).

**patient information leaflet** - 4

**professional reference article** - 4

**hitwise top 10 website** - 1

**discuss** - 1

**the information standard certified member** - 1

**sign in** - 1

**patient app** - 1

**scan me qr code** - 1

**sarah-says-widget** - 1

**loading discussions** - 1

**hon code certified** - 1

**the information standard** - 1

**URLs found in the page: URL - Times Found**
Found 280 urls from where 225 unique.

http://www.patient.co.uk/forums/ - 23
http://www.patient.co.uk/health/Alcoholism-and-Problem-Drinking/professional - 4
http://www.patient.co.uk/health/Alcoholism-and-Problem-Drinking/resources - 4
http://www.patient.co.uk/patientplus.asp - 3
http://www.patient.co.uk/support-directory.asp - 3
http://www.patient.co.uk/pils.asp - 3
http://www.patient.co.uk/mobile.asp - 3
http://www.patient.co.uk/ - 3
- 2
http://www.patient.co.uk/forums - 2
http://www.patient.co.uk/news.asp - 2
http://www.patient.co.uk/showdoc/22 - 2
http://www.patient.co.uk/patient-access.asp - 2
http://www.patient.co.uk/editorial_disclaimer.asp - 2
http://www.patient.co.uk/textbooks.asp - 2
http://www.patient.co.uk/pdf/pilsL580.pdf - 2
http://www.patient.co.uk/health/Alcohol-and-Liver-Disease.htm - 2
http://www.patient.co.uk/health/http%3A%2F%2Fwww%2Epatient%2Eco%2Euk%2Fhealth%2FAlcoholism%2Dand%2DProblem%2DDrinking%2Ehtm&send=false&layout=button_count&width=80&show_faces=false&action=like&colorscheme=light&font&height=21&appId=157980420978393" - 1
http://www.patient.co.uk/doctor/Acute-Alcohol-Withdrawal-and-Delirium-Tremens.htm - 2
http://www.patient.co.uk/health/Alcohol-and-Sensible-Drinking.htm - 2
http://www.patient.co.uk/printer.asp?doc=27000115 - 2
http://www.patient.co.uk/feedback.asp - 2
http://www.patient.co.uk/disclaimer.asp - 2
http://www.patient.co.uk/doctor/Alcohol-Related-Problems.htm - 2
http://www.patient.co.uk/dils.asp - 2
http://www.patient.co.uk/displayCategory/16777279 - 1
http://www.patient.co.uk/books.asp - 1
http://www.patient.co.uk/displayCategory/16777422 - 1
http://www.emis-online.com/about-us/careers/ - 1

http://www.patient.co.uk/offers.asp - 1
http://www.ncbi.nlm.nih.gov/entrez/query.fcgi?cmd=Retrieve&db=PubMed&dopt=Abstract&list_uids=19141482 - 1
http://www.patient.co.uk/displayCategory/16777231 - 1
http://www.ncbi.nlm.nih.gov/entrez/query.fcgi?cmd=Retrieve&db=PubMed&dopt=Abstract&list_uids=19168210 - 1
http://www.patient.co.uk/displayCategory/16777226 - 1
http://www.patient.co.uk/displayCategory/16777274 - 1
http://www.patient.co.uk/drugs/i.htm - 1
http://www.patient.co.uk/fancybox/jquery.fancybox.css - 1
http://www.patient.co.uk/health/Vaginal-Thrush.htm - 1
http://www.patient.co.uk/health/Alcoholism-and-Problem-Drinking.htm#section_5 - 1
http://www.patient.co.uk/patientplus/p.htm - 1
http://www.patient.co.uk/drugs/a.htm - 1
http://www.patient.co.uk/health/http%3A%2F%2Fwww.facebook.com%2Fpatientuk&width=140&height=612&colorscheme=light
&show_faces=true&border_color&stream=false&header=true&appId=157980420978393" - 1
http://www.patient.co.uk/showdoc/50000004 - 1
http://www.patient.co.uk/illness/x.htm - 1
http://www.patient.co.uk/health/blogs/sarah/ - 1
http://www.patient.co.uk/illness/a.htm - 1
http://www.patient.co.uk/drugs/r.htm - 1
http://www.patient.co.uk/patientplus/g.htm - 1
http://www.patient.co.uk/css/print.css - 1
http://www.patient.co.uk/showdoc/50000001 - 1
http://www.patient.co.uk/patientplus/k.htm - 1
http://www.patient.co.uk/drugs/d.htm - 1
http://guidance.nice.org.uk/CG100 - 1
http://www.nacoa.org.uk - 1
http://www.patient.co.uk/patientplus/m.htm - 1
http://www.patient.co.uk/displayCategory/16777230 - 1
http://www.patient.co.uk/patientplus/c.htm - 1
http://www.patient.co.uk/patientplus/b.htm - 1
http://www.patient.co.uk/patientplus/x.htm - 1
http://www.patient.co.uk/displayCategory/16777234 - 1
http://www.patient.co.uk/health/Irritable-Bowel-Syndrome.htm - 1
http://www.patient.co.uk/displayCategory/16777458 - 1
http://www.patient.co.uk/press.asp - 1
http://fluidsurveys.com - 1
http://www.patient.co.uk/drugs/x.htm - 1
https://www.healthonnet.org/HONcode/Conduct.html?HONConduct950113 - 1
http://www.patient.co.uk/displayCategory/16777223 - 1
http://www.drinkaware.co.uk - 1
http://www.patient.co.uk/showdoc/50000011 - 1
http://www.patient.co.uk/showdoc/50000007 - 1
http://www.patient.co.uk/displayCategory/16777460 - 1
http://www.patient.co.uk/patientplus/i.htm - 1
http://www.patient.co.uk/patientplus/d.htm - 1
http://www.patient.co.uk/health/Alcoholism-and-Problem-Drinking.htm#section_2 - 1
http://www.patient.co.uk/showdoc/50000019 - 1
http://www.patient.co.uk/sitemap.asp - 1
http://itunes.apple.com/gb/app/patient.co.uk/id492677470?mt=8&ls=1# - 1
https://patient.emisaccess.co.uk - 1
http://www.ncbi.nlm.nih.gov/entrez/query.fcgi?cmd=Retrieve&db=PubMed&dopt=Abstract&list_uids=15301399 - 1
http://www.patient.co.uk/displayCategory/16777416 - 1
http://www.patient.co.uk/showdoc/50000022 - 1
http://www.al-anonuk.org.uk - 1
http://www.patient.co.uk/awards.asp - 1
http://www.patient.co.uk/showdoc/50000013 - 1
http://www.patient.co.uk/displayCategory/16777233 - 1
http://www.patient.co.uk/illness/m.htm - 1
http://www.patient.co.uk/drugs/g.htm - 1
http://www.patient.co.uk/health/Alcoholism-and-Problem-Drinking.htm#section_8 - 1
http://www.patient.co.uk/showdoc/50000009 - 1
http://www.patient.co.uk/illness/i.htm - 1
http://www.patient.co.uk/displayCategory/16777278 - 1
http://www.patient.co.uk/showdoc/50000010 - 1
http://www.patient.co.uk/showdoc/50000016 - 1
http://www.patient.co.uk/displayCategory/16777462 - 1
http://www.patient.co.uk/drugs/o.htm - 1
https://twitter.com/patientuk - 1
http://www.patient.co.uk/illness/f.htm - 1
http://www.patient.co.uk/patientplus/u.htm - 1
http://www.patient.co.uk/displayCategory/16777225 - 1
http://www.patient.co.uk/patientplus/h.htm - 1

http://www.patient.co.uk/favicon.ico - 1
http://www.patient.co.uk/drugs/j.htm - 1
http://www.patient.co.uk/drugs/v.htm - 1
http://www.patient.co.uk/health/http%3A%2F%2Fwww%2Epatient%2Eco%2Euk&send=false&layout=button_count&width=80&show_faces=false&action=like&colorscheme=light&font&height=21&appId=157980420978393" - 1
http://www.patient.co.uk/patientplus/l.htm - 1
http://www.patient.co.uk/showdoc/50000023 - 1
http://www.patient.co.uk/illness/t.htm - 1
http://www.patient.co.uk/illness/l.htm - 1
http://www.patient.co.uk/health/Alcoholism-and-Problem-Drinking.htm - 1
http://www.patient.co.uk/showdoc/19 - 1
http://www.patient.co.uk/illness/k.htm - 1
http://www.patient.co.uk/shop-index.asp - 1
http://www.patient.co.uk/displayCategory/16777236 - 1
http://www.patient.co.uk/showdoc/50000005 - 1
http://www.patient.co.uk/displayCategory/16777275 - 1
http://www.alcoholics-anonymous.org.uk - 1
http://www.patient.co.uk/showdoc/50000020 - 1
http://www.patient.co.uk/adhow.asp - 1
http://www.patient.co.uk/showdoc/1097 - 1
http://www.patient.co.uk/patientplus/n.htm - 1
http://www.patient.co.uk/health/Haemorrhoids-(Piles).htm - 1
http://www.patient.co.uk/drugs/y.htm - 1
http://www.patient.co.uk/health/Alcoholism-and-Problem-Drinking.htm#refdis - 1
http://www.patient.co.uk/health/Alcoholism-and-Problem-Drinking.htm#section_4 - 1
http://www.patient.co.uk/css/ie6.css - 1
http://www.hitwise.com/uk/resources/top-ten-award?action=confirm&market=uk&sDomain=www.patient.co.uk&iDate=2011&iCatnum=407&Cal=1&semi= - 1
http://www.patient.co.uk/patientplus/y.htm - 1
http://www.patient.co.uk/health/Vitamin-D-Deficiency.htm - 1
http://www.patient.co.uk/showdoc/50000003 - 1
http://www.patient.co.uk/showdoc/50000014 - 1
http://www.patient.co.uk/css/ie7.css - 1
http://www.patient.co.uk/pils.asp?tab=requested - 1
http://www.patient.co.uk/drugs/n.htm - 1
http://www.patient.co.uk/illness/o.htm - 1
http://www.patient.co.uk/patientplus/s.htm - 1
http://www.theinformationstandard.org/members - 1
http://www.patient.co.uk/health/Alcoholism-and-Problem-Drinking.htm#to-content - 1
http://www.patient.co.uk/drugs/l.htm - 1
http://www.patient.co.uk/displayCategory/16777228 - 1
http://www.patient.co.uk/illness/h.htm - 1
http://www.patient.co.uk/patientplus/e.htm - 1
http://www.ncbi.nlm.nih.gov/entrez/query.fcgi?cmd=Retrieve&db=PubMed&dopt=Abstract&list_uids=21414953 - 1
http://www.patient.co.uk/showdoc/50000021 - 1
http://www.patient.co.uk/illness/v.htm - 1
http://www.patient.co.uk/patientplus/w.htm - 1
http://www.patient.co.uk/displayCategory/16777224 - 1
http://www.patient.co.uk/guidelines.asp - 1
http://www.patient.co.uk/displayCategory/16777235 - 1
http://www.patient.co.uk/drugs/m.htm - 1
http://www.patient.co.uk/drugs/z.htm - 1
http://www.patient.co.uk/health/Alcoholism-and-Problem-Drinking.htm#section_6 - 1
http://www.patient.co.uk/drugs/h.htm - 1
http://www.patient.co.uk/displayCategory/16777415 - 1
http://www.patient.co.uk/drugs/u.htm - 1
http://www.patient.co.uk/patientplus/t.htm - 1
http://www.patient.co.uk/health/Shingles.htm - 1
http://www.patient.co.uk/blogs/sarah-says/ - 1
http://www.patient.co.uk/showdoc/50000018 - 1
http://www.patient.co.uk/drugs/e.htm - 1
http://www.patient.co.uk/patientplus/a.htm - 1
http://www.dh.gov.uk/en/Publichealth/Healthimprovement/Alcoholmisuse/index.htm - 1
http://www.patient.co.uk/illness/p.htm - 1
http://www.patient.co.uk/illness/z.htm - 1
http://www.patient.co.uk/about.asp - 1
http://www.patient.co.uk/displayCategory/16777227 - 1
http://www.patient.co.uk/displayCategory/16777276 - 1
http://www.patient.co.uk/dils.asp?tab=requested - 1
http://www.patient.co.uk/pharmacy.asp - 1
http://www.patient.co.uk/health/Alcoholism-and-Problem-Drinking.htm#section_1 - 1
http://www.patient.co.uk/showdoc/50000017 - 1

http://www.patient.co.uk/PatientEquipment.asp - 1
http://www.patient.co.uk/drugs/s.htm - 1
http://guidance.nice.org.uk/CG115 - 1
http://app.fluidsurveys.com/polls/patientuk/exercise-poll/ - 1
http://www.patient.co.uk/blogs/sarah-says - 1
http://www.patient.co.uk/css/style.css - 1
http://www.patient.co.uk/displayCategory/16777277 - 1
http://www.patient.co.uk/displayCategory/16777222 - 1
http://www.patient.co.uk/drugs/t.htm - 1
http://www.patient.co.uk/drugs/q.htm - 1
http://www.patient.co.uk/patientplus/o.htm - 1
http://www.patient.co.uk/drugs/b.htm - 1
http://www.patient.co.uk/showdoc/50000006 - 1
http://www.patient.co.uk/illness/d.htm - 1
http://www.patient.co.uk/drugs/p.htm - 1
http://www.patient.co.uk/showdoc/50000002 - 1
http://www.patient.co.uk/patientplus/r.htm - 1
http://www.patient.co.uk/health/Alcoholism-and-Problem-Drinking.htm#section_7 - 1
http://www.patient.co.uk/illness/w.htm - 1
http://www.patient.co.uk/illness/j.htm - 1
http://www.patient.co.uk/drugs/f.htm - 1
http://www.patient.co.uk/displayCategory/16777461 - 1
http://www.patient.co.uk/drugs/w.htm - 1
http://www.patient.co.uk/illness/c.htm - 1
http://www.patient.co.uk/illness/e.htm - 1
http://www.patient.co.uk/favicon.gif - 1
http://www.patient.co.uk/showdoc/50000008 - 1
http://www.patient.co.uk/displayCategory/16777319 - 1
http://www.ncbi.nlm.nih.gov/entrez/query.fcgi?cmd=Retrieve&db=PubMed&dopt=Abstract&list_uids=20662805 - 1
http://www.patient.co.uk/illness/r.htm - 1
http://www.patient.co.uk/css/ie.css - 1
http://www.patient.co.uk/search.asp - 1
http://www.patient.co.uk/displayCategory/16777232 - 1
http://www.patient.co.uk/illness/g.htm - 1
http://www.patient.co.uk/blogs - 1
http://www.patient.co.uk/displayCategory/16777229 - 1
http://www.patient.co.uk/displayCategory/16777221 - 1
http://www.patient.co.uk/health/Alcoholism-and-Problem-Drinking.htm# - 1
http://www.patient.co.uk/drugs/c.htm - 1
http://www.patient.co.uk/showdoc/50000015 - 1
http://www.patient.co.uk/illness/y.htm - 1
http://www.patient.co.uk/illness/u.htm - 1
http://www.patient.co.uk/patientplus/f.htm - 1
http://www.patient.co.uk/drugs/k.htm - 1
http://www.patient.co.uk/patientplus/z.htm - 1
http://www.patient.co.uk/showdoc/50000012 - 1
http://www.patient.co.uk/illness/b.htm - 1
http://www.patient.co.uk/patientplus/j.htm - 1
http://www.patient.co.uk/patientplus/q.htm - 1
http://www.patient.co.uk/illness/s.htm - 1
http://www.patient.co.uk/patientplus/v.htm - 1
http://www.patient.co.uk/health/Alcoholism-and-Problem-Drinking.htm#section_3 - 1
http://www.patient.co.uk/illness/n.htm - 1
http://www.patient.co.uk/cookies.asp - 1
http://guidance.nice.org.uk/PH24 - 1

5 http://www.blackwellpublishing.com/journal.asp?ref=0145-6008

| Meta tags report for: http://www.blackwellpublishing.com/journal.asp?ref=0145-6008 | | |
|---|---|---|
| meta tag | length | value |
| Title: | 68 | Alcoholism: Clinical and Experimental Research - Journal Information |

| | |
|---|---|
| **Headers returned from: http://www.blackwellpublishing.com/journal.asp?ref=0145-6008** | |
| Related Keyword found<br><br>(number of times and density): | **alcoholism** - 7 - 1.79%<br>**alcohol** - 2 - 0.51%<br>**alcohol-related** - 1 - 0.26%<br>**alcohol-induced** - 1 - 0.26%<br>**alcoholism clinical** - 2 - 0.51%<br>**alcohol abuse** - 1 - 0.26%<br>**sciences alcoholism** - 1 - 0.26%<br>**alcoholism nbsp** - 1 - 0.26%<br>**alcoholism now** - 1 - 0.26%<br>**alcohol-induced syndromes** - 1 - 0.26%<br>**ncadd alcoholism** - 1 - 0.26%<br>**alcohol-related disorders** - 1 - 0.26%<br>**nbsp alcoholism** - 1 - 0.26%<br>**alcoholism alcohol-induced** - 1 - 0.26%<br>**alcohol-related disorders each** - 1 - 0.26%<br>**alcoholism clinical experimental** - 1 - 0.26%<br>**alcohol abuse cause** - 1 - 0.26%<br>**behavioral sciences alcoholism** - 1 - 0.26%<br>**nbsp alcoholism nbsp** - 1 - 0.26%<br>**alcoholism alcohol-induced syndromes** - 1 - 0.26%<br>**addiction nbsp alcoholism** - 1 - 0.26%<br>**alcoholism nbsp skip** - 1 - 0.26%<br>**sciences alcoholism clinical** - 1 - 0.26% |
| cache-control: | Private |
| set-cookie: | ASPSESSIONIDSCTCTCCA=PCMEEMNBJJLGAHOCPKHAENEF; path=/ |
| date: | Sat, 05 May 2012 15:41:29 GMT |
| x-meta-keywords: | |
| x-meta-dc.format: | text/HTML |
| client-peer: | 193.130.69.122:80 |
| client-date: | Sat, 05 May 2012 15:42:07 GMT |
| content-type: | text/html |
| x-meta-issn: | 0145-6008 |
| server: | Microsoft-IIS/6.0 |
| link: | ; /="/"; rel="stylesheet"; type="text/css" |
| x-meta-description: | |
| x-powered-by: | ASP.NET |
| client-response-num: | 1 |
| content-length: | 16155 |
| x-meta-dc.language: | en |
| x-meta-dc.type: | Text |
| title: | Alcoholism: Clinical and Experimental Research - Journal Information |
| x-meta-dc.publisher: | Blackwell Publishing Ltd |

**Keywords found in the Anchor tags: Keyword - Times Found.**
These are text links on your web page (include the 'alt' text from images in the links). These become more important by many search engines (for best results try to name them after your primary keywords).

**alcoholism** - 1

**Keywords found in the IMG Alt tags: Keyword - Times Found.**
This is text found in the 'alt' tag from the images. For web pages with a lot of images those tags are important (for best results try to name them after your primary keywords).

**alcoholism cover** - 1

**wiley knowledge for generations** - 1

**cart.gif** - 1

**alcoholism** - 1

**URLs found in the page: URL - Times Found**
Found 74 urls from where 58 unique.

http://www.wiley.com/go/rss - 4
http://www.wiley.com - 3
http://www.wiley.com/go/resources/authors - 2
http://www.wiley.com/go/help - 2
http://www.interscience.wiley.com/jpages/0145-6008 - 2
http://www.wiley.com/go/resources/librarians - 2
http://www.wiley.com/go/contact_us - 2
http://www.wiley.com/go/resources/instructors - 2
http://www.blackwellpublishing.com/journal.asp?ref=0145-6008#cont - 2
http://www.wiley.com/go/resources/booksellers - 2
http://www.wiley.com/go/sitemap - 2
http://www.wiley.com/go/copyright - 2
http://www.wiley.com/go/privacy_policy - 2
http://www.wiley.com/WileyCDA - 1
http://www.blackwellpublishing.com/offprints.asp?ref=0145-6008&site=1 - 1
http://www.wiley.com/go/resources/press - 1
http://www3.interscience.wiley.com/cgi-bin/setalert?ID=118520059&alert=on - 1
http://www3.interscience.wiley.com/journal/118520059/toc - 1
http://www.blackwellpublishing.com/aims.asp?ref=0145-6008&site=1 - 1
http://www.blackwellpublishing.com/subject_top_med.asp?subj=X - 1
http://www.blackwellpublishing.com/subs.asp?ref=0145-6008&site=1 - 1
http://www.blackwellpublishing.com/bw/c/print.css - 1
http://www.wiley.com/go/locations - 1
http://www.wiley.com/go/careers - 1
http://www.wiley.com/resources/investors - 1
http://www.blackwellpublishing.com/submit.asp?ref=0145-6008&site=1 - 1
http://mc.manuscriptcentral.com/acer - 1
http://www.blackwellpublishing.com/ads.asp?ref=0145-6008&site=1 - 1

http://www.wiley.com/go/resources/societies - 1
http://www.blackwellpublishing.com/librarians/librecfm.asp?ref=0145-6008&title=                    Alcoholism - 1
http://www.wiley.com/go/my_acct - 1
http://www.blackwellpublishing.com/subject_top_sci.asp?subj=V - 1
http://www.blackwellpublishing.com/"http://authorservices.wiley.com/bauthor/faqs_FundingBodyRequirements.asp" - 1
http://www.wiley.com/go/about - 1
http://www.wiley.com/WileyCDA/Section/id-351297.html - 1
http://www.blackwellpublishing.com/journal.asp?ref=0145-6008#side - 1
http://www.blackwellpublishing.com/"http://onlinelibrary.wiley.com/journal/10.1111/(ISSN)1530-0277/earlyview" - 1
http://www.blackwellpublishing.com/bw/wiley/css/base.css - 1
http://purl.org/dc/terms/ - 1
http://www.wiley.com/WileyCDA/Section/id-351310.html - 1
http://www.blackwellpublishing.com/editors.asp?ref=0145-6008&site=1 - 1
http://www.blackwellpublishing.com/"http://www.wiley.com/go/nihmandate" - 1
http://www.wiley.com/go/subjects - 1
http://www.blackwellpublishing.com/permis.asp?ref=0145-6008&site=1 - 1
http://www.blackwellpublishing.com/"http://olabout.wiley.com/WileyCDA/Section/id-406241.html" - 1
http://www3.interscience.wiley.com/rss/journal/118520059 - 1
http://purl.org/dc/elements/1.1/ - 1
http://www.blackwellpublishing.com/ - 1
http://www.wiley.com/go/cart - 1
http://www.blackwellpublishing.com/subject_top_nur.asp?subj=Y - 1
http://www.blackwellpublishing.com/journal.asp?ref=0145-6008&site=1 - 1
http://www.blackwellpublishing.com/bw/wiley/css/base-ie.css - 1
http://www.wiley.com/go/resources/investors - 1
http://www.wiley.com/WileyCDA/Section/id-351311.html - 1
http://www.blackwellpublishing.com/society.asp?ref=0145-6008&site=1 - 1
http://www.blackwellpublishing.com/journal.asp?ref=0145-6008#top - 1
http://www.blackwellpublishing.com/subject_top_social.asp?subj=S - 1
http://www3.interscience.wiley.com/aboutus/contactus/contactus.html - 1

6 http://www.netdoctor.co.uk/health_advice/facts/alcoholism.htm

| Meta tags report for: http://www.netdoctor.co.uk/health_advice/facts/alcoholism.htm | | |
|---|---|---|
| meta tag | length | value |
| Title: | 10 | Alcoholism |
| Description: | 86 | Find out what alcohol dependence is, and what problems are caused by too much alcohol? |
| Keywords: | 247 | alcohol,dependence,alcoholic,alcohol,drunk,drink,CAGE,cage,addiction,alcohol          consumption,physical illness,psychological        and        social        distress,dependence,Alcoholism,Tolerance,Withdrawal symptoms,withdrawel,tremor,liver cirrhosis,depression,dependence |

| eaders returned from: http://www.netdoctor.co.uk/health_advice/facts/alcoholism.htm |
|---|

| Related Keyword found (number of times and density): | **alcoholism** - 8 - 0.59% **alcohol** - 35 - 2.60% **alcoholic** - 1 - 0.07% **alcoholics** - 1 - 0.07% |
|---|---|
| x-meta-description: | Find out what alcohol dependence is, and what problems are caused by too much alcohol? |
| link: | ; media="screen"; rel="stylesheet"; type="text/css" |
| connection: | close |
| client-response-num: | 1 |
| accept-ranges: | bytes |
| date: | Sat, 05 May 2012 17:37:13 GMT |
| x-meta-keywords: | alcohol,dependence,alcoholic,alcohol,drunk,drink,CAGE,cage,addiction,alcohol consumption,physical illness,psychological and social distress,dependence,Alcoholism,Tolerance,Withdrawal symptoms,withdrawel,tremor,liver cirrhosis,depression,dependence |
| vary: | Accept-Encoding |
| client-peer: | 92.43.127.36:80 |
| client-date: | Sat, 05 May 2012 17:37:09 GMT |
| content-type: | text/html |
| title: | Alcoholism |
| client-transfer-encoding: | chunked |
| server: | Apache/2.2.18 (Unix) mod_ssl/2.2.18 OpenSSL/0.9.7d DAV/2 PHP/5.2.17 |

**Keywords found in the Anchor tags: Keyword - Times Found.**
These are text links on your web page (include the 'alt' text from images in the links). These become more important by many search engines (for best results try to name them after your primary keywords).

**smoking alcohol and drugs** - 4

**alcohol and drugs** - 2

**do i drink too much alcohol** - 1

**alcoholics anonymous** - 1

**latest smoking alcohol and drugs news** - 1

**do i have a problem with alcohol** - 1

**alcohol and drug misuse** - 1

**Keywords found in the IMG Alt tags: Keyword - Times Found.**
This is text found in the 'alt' tag from the images. For web pages with a lot of images those tags are important (for best results try to name them after your primary keywords).

netdoctor lene justesen - 1

netdoctor.co.uk - 1

symptom checker - 1

getty smoke - 1

getty le diet - 1

getty doctor - 1

getty online doctor - 1

follow us on twitter - 1

getty realage - 1

**URLs found in the page: URL - Times Found**
Found 201 urls from where 132 unique.

http://www.netdoctor.co.uk/interactive/symptom_sorter/index.php - 4
http://www.netdoctor.co.uk/videos-a-z/index.shtml - 3
http://www.netdoctor.co.uk/smoking-alcohol-and-drugs/health-centre/index.shtml - 3
http://www.netdoctor.co.uk/whoisnd.htm - 3
http://www.netdoctor.co.uk/interactive/signup.php - 3
http://www.netdoctor.co.uk/pregnancy-and-family/health-centre/index.shtml - 2
http://www.netdoctor.co.uk/mouth-and-teeth/health-centre/index.shtml - 2
http://www.netdoctor.co.uk/heart-health-and-blood-problems/health-centre/index.shtml - 2
http://www.netdoctor.co.uk/diet-and-nutrition/health-centre/index.shtml - 2
http://www.netdoctor.co.uk/disclaimer.htm - 2
http://www.netdoctor.co.uk/interactive/news/theme_news_main.php - 2
http://www.netdoctor.co.uk/liver-kidney-and-urinary-system/health-centre/index.shtml - 2
http://www.netdoctor.co.uk/examinations-tests-a-z/index.shtml - 2
http://www.netdoctor.co.uk/erectile-dysfunction/health-centre/index.shtml - 2
http://www.netdoctor.co.uk/depression/health-centre/index.shtml - 2
http://www.netdoctor.co.uk/ear-nose-and-throat/health-centre/index.shtml - 2
http://www.netdoctor.co.uk/skin-and-hair/health-centre/index.shtml - 2
http://www.netdoctor.co.uk/teenage-health/health-centre/index.shtml - 2
http://www.netdoctor.co.uk/allergy-and-asthma/health-centre/index.shtml - 2
http://www.netdoctor.co.uk/adhd/health-centre/index.shtml - 2
http://www.netdoctor.co.uk/healthyliving/complementary-therapies/index.shtml - 2
http://lediet.netdoctor.co.uk/home_lediet.html - 2
http://twitter.com/NetDoctor - 2
http://www.netdoctor.co.uk/womens-health/health-centre/index.shtml - 2
http://www.netdoctor.co.uk/diseases/index.shtml - 2
http://onlinedr.netdoctor.co.uk/ - 2
http://www.netdoctor.co.uk/childrens-health/health-centre/index.shtml - 2
http://www.netdoctor.co.uk/medicines/index.shtml - 2
http://www.netdoctor.co.uk/diabetes/health-centre/index.shtml - 2
http://www.netdoctor.co.uk/surgical-procedures-a-z/index.shtml - 2
http://www.netdoctor.co.uk/infections/health-centre/index.shtml - 2
http://www.stayquit.co.uk/uk/signup/index.php - 2
http://www.netdoctor.co.uk/wellbeing/health-centre/index.shtml - 2
http://www.netdoctor.co.uk/mens-health/health-centre/index.shtml - 2
http://www.netdoctor.co.uk/brain-and-nervous-system/health-centre/index.shtml - 2
http://www.netdoctor.co.uk/sport-and-fitness/health-centre/index.shtml - 2
http://www.netdoctor.co.uk/aches-and-pains/health-centre/index.shtml - 2
http://www.netdoctor.co.uk/digestive-health/health-centre/index.shtml - 2

http://www.netdoctor.co.uk/ - 2
http://www.netdoctor.co.uk/healthyliving/beauty/index.shtml - 2
http://www.netdoctor.co.uk/interactive/discussion/index.php?tab_id=247&c=5 - 2
http://netdoctor.privatehealth.co.uk/private-hospitals/hospitaltreatment-enquiryform/ - 2
http://www.netdoctor.co.uk/travel-health/health-centre/index.shtml - 2
http://www.netdoctor.co.uk/seniors-health/health-centre/index.shtml - 2
http://www.netdoctor.co.uk/alcoholanddrugs/index.shtml?tab_id=101 - 2
http://www.netdoctor.co.uk/sex-and-relationships/health-centre/index.shtml - 2
http://livingaids.netdoctor.co.uk/index.php - 2
http://www.netdoctor.co.uk/interactive/news/theme_news_main.php?tab_id=135 - 2
http://www.netdoctor.co.uk/lungs/health-centre/index.shtml - 2
http://netdoctor.privatehealth.co.uk/ - 2
http://www.netdoctor.co.uk/parenting/health-centre/index.shtml - 2
http://www.netdoctor.co.uk/accidents-and-first-aid/health-centre/index.shtml - 2
http://www.netdoctor.co.uk/testyourself/index.shtml - 2
http://pregnancyservice.netdoctor.co.uk/ - 2
http://onlinedr.netdoctor.co.uk - 2
http://www.stayquit.co.uk - 2
http://www.netdoctor.co.uk/cancer/health-centre/index.shtml - 2
http://www.netdoctor.co.uk/ate/health/index/600220.shtml - 2
http://www.netdoctor.co.uk/eyes/health-centre/index.shtml - 2
http://www.netdoctor.co.uk/health-services-guide/index.shtml - 2
http://www.netdoctor.co.uk/support_groups/index.shtml - 2
http://www.netdoctor.co.uk/healthy-recipes/health-centre/index.shtml - 2
http://lediet.netdoctor.co.uk - 2
http://www.netdoctor.co.uk/interactive/news/free-school-meals-a-predictor-of-substance-abuse-id801351049-t135.html - 1
http://www.netdoctor.co.uk/interactive/discussion/Yellow-Tongue-nasty-taste-in-mouth-t64941-f11.html - 1
http://www.netdoctor.co.uk/interactive/video/index.php?&lineupId=1219349298001&tab_id=317 - 1
http://www.netdoctor.co.uk/all-diseases-and-conditions.shtml - 1
http://www.netdoctor.co.uk/ate/health/smoking_alcohol_drugs/600244.shtml?tab_id=99 - 1
http://www.netdoctor.co.uk/health_advice/facts/drugabuse.htm - 1
http://www.netdoctor.co.uk/health_advice/facts/../../diseases/facts/insomnia.htm - 1
http://www.netdoctor.co.uk/interactive/news/firmer-action-needed-to-bring-down-global-smoking-rates-id801349070-t135.html - 1
http://www.netdoctor.co.uk/ate/smokingalcoholdrugs/204666.html - 1
http://www.netdoctor.co.uk/interactive/interactivetests/smokingtest.php - 1
http://www.netdoctor.co.uk/all-health-A-Z.shtml - 1
http://www.netdoctor.co.uk/health_advice/facts/../../health_advice/facts/drugabuse.htm - 1
http://www.netdoctor.co.uk/interactive/news/cigarette-pack-health-warnings-beneficial-for-ex-smokers-id801350246-t135.html - 1
http://www.netdoctor.co.uk/interactive/discussion/Alcohol-and-drug-misuse-f11.html - 1
http://www.netdoctor.co.uk/style/print.css - 1
http://www.netdoctor.co.uk/smokingalcoholanddrugs/support_groups/006017.htm - 1
http://ad.doubleclick.net/clk;240459219;60197509;m?http://realage.netdoctor.co.uk - 1
http://www.netdoctor.co.uk/ate/smokingalcoholdrugs/204028.html - 1
http://www.netdoctor.co.uk/ate/smokingalcoholdrugs/205967.html - 1
http://www.netdoctor.co.uk/interactive/discussion/Cannabis-smoking---worried-about-my-partner-t64796-f11.html - 1
http://www.netdoctor.co.uk/interactive/discussion/Champix-t6901-f11.html - 1
http://www.netdoctor.co.uk/interactive/discussion/index.php?tab_id=102&c=19 - 1
http://www.facebook.com/pages/NetDoctor/136143599746047?v=wall&ref=ts - 1
http://www.netdoctor.co.uk/smoking-alcohol-and-drugs/health-centre/index.shtml?tab_id=101 - 1
http://www.omniture.com - 1
http://www.netdoctor.co.uk/interactive/video/index.php?&lineupId=1219349298001&videoId=1205617764001&tab_id=317 - 1
http://www.netdoctor.co.uk/interactive/discussion/search.php?mode=viewdiscussion&u=187230&search_author=clawd - 1
http://www.netdoctor.co.uk/style/smoking_themeV2.css - 1
http://www.netdoctor.co.uk/medicines/100003484.html - 1
http://www.netdoctor.co.uk/style/styleV2.css - 1
http://www.netdoctor.co.uk/interactive/discussion/search.php?mode=viewdiscussion&u=173328&search_author=AuntSally - 1
http://www.netdoctor.co.uk/privacy.htm - 1
http://www.netdoctor.co.uk/health_advice/facts/alcoholism.htm - 1
http://www.netdoctor.co.uk/contactus.shtml - 1
http://www.netdoctor.co.uk/health_advice/facts/../../alcoholanddrugs/alcohol_toomuch.htm - 1
http://www.netdoctor.co.uk/alcoholanddrugs/alcohol_toomuch.htm - 1
http://www.netdoctor.co.uk/smokingalcoholdrugs/support_groups/006076.htm - 1
http://www.netdoctor.co.uk/smoking/index.shtml?tab_id=100 - 1
http://www.netdoctor.co.uk/style/style-ie.css - 1
http://www.netdoctor.co.uk/medicines/100005109.html - 1
http://www.netdoctor.co.uk/health_advice/facts/../../smoking/heartfailure.htm - 1
http://www.netdoctor.co.uk/interactive/discussion/search.php?mode=viewdiscussion&u=12787&search_author=Kath131 - 1
http://www.netdoctor.co.uk/style/style-ie6.css - 1
http://www.netdoctor.co.uk/health_advice/facts/smokehealth.htm - 1
http://www.netdoctor.co.uk/sitemap.html - 1
https://onlinedr.netdoctor.co.uk/stop_smoking/meds/champix - 1
http://www.netdoctor.co.uk/ads.htm - 1

http://www.netdoctor.co.uk/smoking-alcohol-and-drugs/health-centre/index.shtml?tab_id=354 - 1
http://www.netdoctor.co.uk/health_advice/facts/smokequit.htm - 1
http://www.netdoctor.co.uk/all-health-services.shtml - 1
http://www.netdoctor.co.uk/health_advice/facts/.../.../alcoholanddrugs/depression.htm - 1
http://www.netdoctor.co.uk/interactive/news/australian-study-finds-tv-ads-encourage-child-drinking-id801346251-t135.html - 1
http://www.netdoctor.co.uk/all-healthy-living.shtml - 1
http://onlinedr.netdoctor.co.uk/?tab_id=268 - 1
http://www.netdoctor.co.uk/alcoholanddrugs/menshealth/facts/addiction.htm - 1
http://www.netdoctor.co.uk/smokingalcoholanddrugs/support_groups.shtml?tab_id=98 - 1
http://www.netdoctor.co.uk/interactive/news/young-scots-leading-healthier-lifestyles-id801354782-t135.html - 1
http://www.netdoctor.co.uk - 1
http://www.netdoctor.co.uk/health_advice/facts/.../.../diseases/facts/cirrhosisliver.htm - 1
http://www.netdoctor.co.uk/interactive/video/index.php?&lineupId=1219349298001&videoId=1205617737001&tab_id=317 - 1
http://www.netdoctor.co.uk/smokingalcoholanddrugs/support_groups/006119.htm - 1
http://www.netdoctor.co.uk/interactive/discussion/search.php?mode=viewdiscussion&u=186951&search_author=Dilg - 1
http://www.netdoctor.co.uk/interactive/discussion/Husband-cant-stop-drinking-t65118-f11.html - 1
http://www.netdoctor.co.uk/smokingalcoholanddrugs/support_groups/006019.htm - 1
http://www.netdoctor.co.uk/interactive/discussion/search.php?mode=viewdiscussion&u=66975&search_author=Male_nurse - 1
http://www.netdoctor.co.uk/ate/smokingalcoholdrugs/205909.html - 1
http://www.netdoctor.co.uk/interactive/discussion/Coming-Off-Tramadol-After-13-years-t51678-f11.html - 1
http://www.netdoctor.co.uk/interactive/video/index.php?&lineupId=1219349298001&videoId=1809488587&tab_id=317 - 1
http://www.netdoctor.co.uk/health_advice/facts/.../.../diseases/depression/anxietydisorders_000017.htm - 1

7 http://www.medicinenet.com/alcohol_abuse_and_alcoholism/article.htm

| Meta tags report for: http://www.medicinenet.com/alcohol_abuse_and_alcoholism/article.htm |
|---|
| Found no meta tags. |
| Headers returned from: http://www.medicinenet.com/alcohol_abuse_and_alcoholism/article.htm |

| Related Keyword found<br><br>(number of times and density): | **alcohol** - 49 - 1.28%<br>**alcoholism** - 42 - 1.10%<br>**alcoholic** - 1 - 0.03%<br>**alcoholics** - 1 - 0.03%<br>**alcohol-use** - 1 - 0.03% |
|---|---|
| x-powered-by: | ASP.NET |
| client-response-num: | 1 |
| set-cookie: | ASPSESSIONIDCSRSCCAS=PEBDCEPBDPCODJBPBOFJPJFD; path=/ |
| cache-control: | private |
| date: | Sat, 05 May 2012 19:14:43 GMT |
| client-peer: | 207.231.206.235:80 |

| content-length: | 108717 |
|---:|:---|
| client-date: | Sat, 05 May 2012 19:14:39 GMT |
| content-type: | text/html |
| server: | Microsoft-IIS/7.5 |

**Keywords found in the Anchor tags: Keyword - Times Found.** These are text links on your web page (include the 'alt' text from images in the links). These become more important by many search engines (for best results try to name them after your primary keywords).

health risks of alcohol abuse slideshow pictures - 2

what is alcohol abuse - 2

alcohol abuse and alcoholism index – 2

how can someone find more information or get help or support to treat alcohol abuse and alcoholism - 1

is it safe to drink alcohol while pregnant - 1

what is the treatment for alcoholism - 1

how is alcoholism diagnosed - 1

what causes alcoholism is alcoholism hereditary - 1

what medications treat alcoholism - 1

what is alcoholism - 1

can an alcoholic just cut back or stop drinking - 1

what differentiates alcohol abuse from alcoholism - 1

what are the stages of alcoholism - 1

blood alcohol level - 1

can alcoholism be prevented - 1

what are alcohol abuse and alcoholism symptoms and signs in teenagers women men and the elderly - 1

children of alcoholics - 1

patient discussions alcohol abuse and alcoholism experience - 1

alcoholism - 1

alcohol and teens - 1

what are risk factors for alcoholism - 1

patient discussions alcohol abuse and alcoholism treatments - 1

what is the prognosis of alcoholism - 1

what are the long-term physical and psychological effects of alcohol abuse and alcoholism - 1

**alcoholism and alcohol abuse facts** - 1

**Keywords found in the IMG Alt tags: Keyword - Times Found.**
This is text found in the 'alt' tag from the images. For web pages with a lot of images those tags are important (for best results try to name them after your primary keywords).

**doctor to patient** - 2

**xml** - 2

**symptom checker your guide to symptoms signs pinpoint your pain** - 1

**medicinenet doctors** - 1

**truste online privacy certification** - 1

**roxanne dryden-edwards md** - 1

**medicinenet** - 1

**adult skin problems slideshow** - 1

**discover the connection between viral hepatitis alcoholism cirrhosis and liver cancer** - 1

**health and living recipes nutrition exercise and more** - 1

**otc and prescription drug abuse** - 1

**this website is certified by health on the net foundation click to verify** - 1

**adult skin conditions recognize these skin conditions** - 1

**melissa conrad stöppler md** - 1

**health risks of alcohol abuse slideshow pictures** - 1

**view slideshow pictures** - 1

**RLs found in the page: URL - Times Found**
Found 370 urls from where 238 unique.

- 40
http://www.medicinenet.com/script/main/art.asp?articlekey=8709&page=2 - 5
http://www.medicinenet.com/alcohol_abuse_and_alcoholism/article.htm# - 5
http://www.medicinenet.com/diseases_and_conditions/article.htm - 5
http://www.medicinenet.com/health_and_living/focus.htm - 5
http://www.medicinenet.com/slideshows/article.htm - 4
http://www.medicinenet.com/script/main/art.asp?articlekey=8709&questionid=928 - 4
http://www.medicinenet.com/image_collection/article.htm - 3
http://www.facebook.com/MedicineNet - 3
http://www.medicinenet.com/script/main/art.asp?articlekey=1917 - 3
http://www.medicinenet.com/script/main/hp.asp - 3
http://www.medicinenet.com/symptoms_and_signs/article.htm - 3
http://www.medicinenet.com/medications/article.htm - 3
http://www.medicinenet.com/quizzes_a-z_list/article.htm - 3
http://www.medicinenet.com/script/main/art.asp?articlekey=107525 - 3
http://www.medicinenet.com/mental_health/focus.htm - 3
https://pref.health.webmd.com/mnet/Newslettersignup.asp - 3
http://www.medicinenet.com/script/main/forum.asp?articlekey=8709 - 3
http://www.medicinenet.com/script/main/art.asp?articlekey=8709&page=3 - 2
http://www.medicinenet.com/script/main/art.asp?articlekey=8709&page=9 - 2
http://www.medicinenet.com/script/main/art.asp?articlekey=342 - 2
http://www.medicinenet.com/script/main/art.asp?articlekey=110443 - 2

http://www.medicinenet.com/script/main/art.asp?articlekey=8709&page=11 - 2
http://www.medicinenet.com/womens_health/focus.htm - 2
http://www.medicinenet.com/script/main/art.asp?articlekey=97800 - 2
http://www.medicinenet.com/pregnancy/focus.htm - 2
http://www.medicinenet.com/rss/dailyhealth.xml - 2
http://www.medicinenet.com/script/main/art.asp?articlekey=41529 - 2
http://www.medicinenet.com/cold_and_flu/focus.htm - 2
http://www.medicinenet.com/high_blood_pressure/focus.htm - 2
http://www.medicinenet.com/script/main/art.asp?articlekey=8709&page=5 - 2
http://www.medicinenet.com/chronic_pain/focus.htm - 2
http://www.medicinenet.com/script/main/art.asp?articlekey=122404 - 2
http://www.medicinenet.com/allergies/focus.htm - 2
http://www.medicinenet.com/script/main/art.asp?articlekey=374 - 2
http://www.medicinenet.com/skin/focus.htm - 2
http://www.medicinenet.com/arthritis/focus.htm - 2
http://www.medicinenet.com/diabetes/focus.htm - 2
http://www.medicinenet.com/sexual_health/focus.htm - 2
http://www.medicinenet.com/healthy_kids/focus.htm - 2
http://www.medicinenet.com/script/main/art.asp?articlekey=8709&page=10 - 2
http://www.medicinenet.com/digestion/focus.htm - 2
http://twitter.com/medicinenet - 2
http://www.medicinenet.com/script/main/art.asp?articlekey=8709&page=7 - 2
https://plus.google.com/117041701306844597651 - 2
http://www.medicinenet.com/script/main/art.asp?articlekey=80650 - 2
http://www.medicinenet.com/script/main/art.asp?articlekey=8709&page=8 - 2
http://www.medicinenet.com/cancer/focus.htm - 2
http://www.medicinenet.com/mens_health/focus.htm - 2
http://www.medicinenet.com/script/main/art.asp?articlekey=42985 - 2
http://www.medicinenet.com/script/main/art.asp?articleKey=99154 - 2
http://www.medicinenet.com/script/main/art.asp?articlekey=8709&questionid=717 - 2
http://www.medicinenet.com/script/main/art.asp?articlekey=8709&page=4 - 2
http://www.medicinenet.com/script/main/art.asp?articlekey=52962 - 2
http://www.medicinenet.com/script/main/art.asp?articlekey=368 - 2
http://www.medicinenet.com/pets/pethealth/focus.htm - 2
http://www.medicinenet.com/procedures_and_tests/article.htm - 2
http://www.medicinenet.com/script/main/art.asp?articlekey=8709&page=6 - 2
http://www.medicinenet.com/heart/focus.htm - 2
http://www.medicinenet.com/hearing/focus.htm - 2
http://www.medicinenet.com/script/main/art.asp?articlekey=322 - 2
http://www.medicinenet.com/weight_loss/article.htm - 2
http://www.medicinenet.com/alcohol_abuse_and_alcoholism/city.htm - 2
http://www.medicinenet.com/infectious_disease/focus.htm - 2
http://www.medicinenet.com/script/main/art.asp?articlekey=505 - 2
http://www.medicinenet.com/script/main/art.asp?articlekey=119180 - 2
http://www.medterms.com/script/main/hp.asp - 2
http://www.medicinenet.com/script/main/art.asp?articlekey=18915 - 1
http://www.medicinenet.com/script/main/art.asp?articlekey=109861 - 1
http://www.medicinenet.com/site_map/article.htm - 1
http://www.medicinenet.com/script/main/art.asp?articlekey=157484 - 1
http://www.medicinenet.com/script/main/art.asp?articlekey=101262 - 1
http://www.medicinenet.com/script/main/art.asp?articlekey=89487 - 1
http://www.webmd.com/ - 1
http://www.medicinenet.com/script/main/art.asp?articlekey=52905 - 1
http://www.medicinenet.com/cholesterol/focus.htm - 1
http://www.medicinenet.com/script/main/art.asp?articlekey=8709&page=9#info - 1
http://www.medicinenet.com/script/main/art.asp?articlekey=106752 - 1
http://www.medicinenet.com/script/main/art.asp?articlekey=157402 - 1
http://www.medicinenet.com/rss/article.htm - 1
http://www.medicinenet.com/rss/general/mental_health.xml - 1
http://www.medicinenet.com/script/main/art.asp?articlekey=89980 - 1
http://www.medicinenet.com/script/main/art.asp?articlekey=8709&page=10#codependency - 1
http://www.medicinenet.com/script/main/alphaidx.asp?p=a_48 - 1
http://www.medicinenet.com/rss/weeklyhealth.xml - 1
http://www.medicinenet.com/diet_and_weight_management/article.htm - 1
http://www.medicinenet.com/asthma/focus.htm - 1
http://www.medicinenet.com/script/main/art.asp?articlekey=8709&page=5#diagnosis - 1
http://www.medicinenet.com/thyroid/focus.htm - 1
http://www.medicinenet.com/script/main/art.asp?articlekey=8893 - 1
http://www.medicinenet.com/script/main/art.asp?articlekey=8709&page=6#treatment - 1
http://www.medicinenet.com/script/main/art.asp?articlekey=54999 - 1
http://www.medicinenet.com/script/main/art.asp?articlekey=24418 - 1
http://www.medicinenet.com/script/main/art.asp?articlekey=8709&page=8#alcoholic - 1

http://www.medicinenet.com/migraine/focus.htm - 1
http://www.medicinenet.com/script/main/art.asp?articlekey=440 - 1
http://images.medicinenet.com/css/medicinenet/global/1.1/global.min.css - 1
http://www.medicinenet.com/script/main/art.asp?articlekey=107851 - 1
http://www.medicinenet.com/fat_and_fats_quiz/quiz.htm - 1
http://www.medicinenet.com/script/main/art.asp?articlekey=8709&page=4#causes - 1
http://www.medicinenet.com/script/main/art.asp?articlekey=8709&page=4#symptoms - 1
http://www.medicinenet.com/hiv/focus.htm - 1
http://www.medicinenet.com/script/main/art.asp?articlekey=8709&page=2 - 1
http://www.medicinenet.com/script/main/forum.asp?articlekey=8709&articletype=pharm - 1
http://www.medicinenet.com/script/main/art.asp?articlekey=12596 - 1
http://www.medicinenet.com/script/main/art.asp?articlekey=114760 - 1
http://www.medicinenet.com/script/main/art.asp?articlekey=26119 - 1
http://images.medicinenet.com/css/sponsorbox.css - 1
http://images.medicinenet.com/css/medicinenet/refIndex.css - 1
http://www.medicinenet.com/script/main/art.asp?articlekey=107837 - 1
http://www.medicinenet.com/script/main/art.asp?articlekey=431 - 1
http://www.webmd.com/depression/recognizing-depression-symptoms/physical-symptoms - 1
http://www.webmd.com/drug-medication/medicationawareness/default.htm - 1
http://www.medicinenet.com/images/fav/MN.ico - 1
http://www.medicinenet.com/script/main/art.asp?articlekey=497 - 1
http://www.medicinenet.com/sleep/focus.htm - 1
http://www.medicinenet.com/script/main/art.asp?articlekey=120152 - 1
http://www.medicinenet.com/script/main/art.asp?articlekey=105631 - 1
http://www.medicinenet.com/script/main/art.asp?articlekey=81207 - 1
http://www.medicinenet.com/script/main/forum.asp?articlekey=8709&articletype=dt - 1
http://www.medicinenet.com/script/main/art.asp?articlekey=106763 - 1
http://www.medicinenet.com/script/main/art.asp?articlekey=191 - 1
http://www.medicinenet.com/script/main/art.asp?articlekey=79201 - 1
http://www.medicinenet.com/happiness_quiz/quiz.htm - 1
http://www.medicinenet.com/nutrition_food_and_recipes/article.htm - 1
http://www.medicinenet.com/script/main/art.asp?articlekey=41916 - 1
http://www.medicinenet.com/script/main/art.asp?articlekey=38058 - 1
http://www.medicinenet.com/depression/focus.htm - 1
http://www.medicinenet.com/script/main/art.asp?articlekey=180 - 1
http://www.medicinenet.com/alcohol_abuse_and_alcoholism/related-conditions/index.htm - 1
http://www.medicinenet.com/supplements-vitamins/article.htm - 1
http://www.medscape.com/ - 1
https://data.webmd.com/sdclive/SdcForm.aspx?FormId=MedNetContact - 1
http://www.wbmd.com/ - 1
http://www.medicinenet.com/script/main/art.asp?articlekey=16745 - 1
http://www.medicinenet.com/script/main/art.asp?articlekey=6930 - 1
http://www.medicinenet.com/script/main/art.asp?articlekey=915 - 1
http://www.medicinenet.com/travel_health/focus.htm - 1
http://images.medicinenet.com/css/medicinenet/ie_fix/ie.css - 1
http://www.medicinenet.com/script/main/art.asp?articlekey=63261 - 1
http://www.medicinenet.com/script/main/art.asp?articlekey=470 - 1
http://www.medicinenet.com/script/main/art.asp?articlekey=52909 - 1
http://www.rxlist.com/pharmacy/local_locations_pharmacies.htm - 1
http://www.medicinenet.com/rss/medicinenet_slideshows.xml - 1
http://www.medicinenet.com/script/main/art.asp?articlekey=8709&page=7#medications - 1
http://www.medicinenet.com/script/main/art.asp?articlekey=113903 - 1
http://www.medicinenet.com/script/main/art.asp?articlekey=2001 - 1
http://www.medicinenet.com/script/main/art.asp?articlekey=106753 - 1
http://www.medicinenet.com/script/main/art.asp?articlekey=25825 - 1
http://www.medicinenet.com/alcohol_abuse_and_alcoholism/images-quizzes/index.htm - 1
http://www.medicinenet.com/rss/specialty/alcohol_abuse_and_alcoholism.xml - 1
http://www.medicinenet.com/script/main/art.asp?articlekey=107516 - 1
http://www.medicinenet.com/script/main/art.asp?articlekey=20052 - 1
http://www.medicinenet.com/script/main/art.asp?articlekey=276 - 1
http://www.medicinenet.com/script/main/art.asp?articlekey=10929 - 1
http://www.medicinenet.com/script/main/art.asp?articlekey=12507 - 1
http://www.medicinenet.com/script/main/art.asp?articlekey=38063 - 1
http://www.medicinenet.com/script/main/art.asp?articlekey=107517 - 1
http://www.medicinenet.com/script/main/art.asp?articlekey=8709&page=9#effects - 1
http://www.medicinenet.com/script/main/art.asp?articlekey=488 - 1
http://www.medicinenet.com/slideshows-picks/article.htm - 1
https://www.healthonnet.org/HONcode/Conduct.html?HONConduct594277 - 1
http://www.medicinenet.com/script/main/art.asp?articlekey=8709&page=4#symptoms - 1
http://www.medicinenet.com/script/main/submit-viewer-comments.asp?questionid=717 - 1
http://www.medicinenet.com/oralhealth/focus.htm - 1
http://www.medicinenet.com/script/main/forum.asp?articlekey=8709&articletype=news - 1

http://www.medicinenet.com/skin-problems-pictures-slideshow/article.htm - 1
http://www.medicinenet.com/script/main/art.asp?articlekey=8709&page=2#what - 1
http://www.medicinenet.com/script/main/forum.asp?articlekey=8709&articletype=itn - 1
http://www.medicinenet.com/alcohol_abuse_and_alcoholism/glossary.htm - 1
http://www.medicinenet.com/script/main/art.asp?articlekey=384 - 1
http://www.privacy.truste.com/privacy-seal/WebMD,-LLC/validation?rid=7b3bf16d-37b4-42e0-aec7-836ef1d3680d - 1
http://www.medicinenet.com/script/main/art.asp?articlekey=103748 - 1
https://www.hon.ch/HONcode/Conduct.html?HONConduct594277 - 1
http://www.webmd.com/mobile - 1
http://www.medicinenet.com/script/main/art.asp?articlekey=467 - 1
http://www.medicinenet.com/script/main/art.asp?articlekey=33915 - 1
http://www.webmd.com/healthy-aging/drug-costs-12/questions-about-generic-drugs - 1
http://www.medicinenet.com/script/main/art.asp?articlekey=8709&page=10#prevention - 1
http://www.healthonnet.org/HONcode/Conduct.html - 1
http://images.medicinenet.com/css/medicinenet/header/1.4/header.min.css - 1
http://www.medicinenet.com/script/main/art.asp?articlekey=439 - 1
http://www.medicinenet.com/script/main/art.asp?articlekey=334 - 1
http://www.medicinenet.com/script/main/art.asp?articlekey=54633 - 1
http://www.medicinenet.com/script/main/art.asp?articlekey=39156 - 1
http://www.medicinenet.com/script/main/art.asp?articlekey=2787 - 1
http://www.medicinenet.com/senior_health/focus.htm - 1
http://www.medicinenet.com/script/main/art.asp?articlekey=8709&page=11#prognosis - 1
http://www.medicinenet.com/script/main/art.asp?articlekey=125353 - 1
http://www.medicinenet.com/script/main/art.asp?articlekey=8709&page=1#facts - 1
http://www.rxlist.com/script/main/hp.asp - 1
http://www.medicinenet.com/script/main/art.asp?articlekey=10181 - 1
http://www.medicinenet.com/alzheimers/focus.htm - 1
http://www.medicinenet.com/script/main/art.asp?articlekey=99154 - 1
http://www.rxlist.com/pill-identification-tool/article.htm - 1
http://www.medicinenet.com/script/main/art.asp?articlekey=100985 - 1
http://www.medicinenet.com/script/main/art.asp?articlekey=157774 - 1
http://www.medicinenet.com/skin_conditions_picture_quiz/quiz.htm - 1
http://www.medicinenet.com/eyesight/focus.htm - 1
http://www.medicinenet.com/script/main/art.asp?articlekey=152720 - 1
http://www.medicinenet.com/script/main/art.asp?articlekey=120794 - 1
http://emedicine.medscape.com/ - 1
http://www.medicinenet.com/script/main/art.asp?articlekey=157595 - 1
http://www.medicinenet.com/script/main/art.asp?articlekey=8709&page=8#safe - 1
http://images.medicinenet.com/css/medicinenet/html5/html5reset-1.6.1.css - 1
http://www.medicinenet.com/script/main/submit-viewer-comments.asp?questionid=928 - 1
http://www.medicinenet.com/alcohol_abuse_and_alcoholism/index.htm#Health-Features - 1
http://www.medicinenet.com/script/main/art.asp?articlekey=22133 - 1
http://www.medicinenet.com/lungs/focus.htm - 1
http://www.medicinenet.com/script/main/art.asp?articlekey=509 - 1
http://www.medicinenet.com/script/main/art.asp?articlekey=2009 - 1
http://www.medicinenet.com/script/main/art.asp?articlekey=8709&page=9#pregnant - 1
http://www.medicinenet.com/superfoods_quiz/quiz.htm - 1
http://www.medicinenet.com/exercise_and_fitness/article.htm - 1
http://www.medicinenet.com/script/main/art.asp?articlekey=8709 - 1
http://www.webmd.com/oral-health/healthy-teeth-1/default.htm?vid=liveEvent0310&ch=3 - 1
http://www.medicinenet.com/neurology/focus.htm - 1
http://www.medicinenet.com/script/main/art.asp?articlekey=8709&page=8#drinking - 1
http://www.medicinenet.com/osteoarthritis-pain/focus.htm - 1
http://www.medicinenet.com/script/main/art.asp?articlekey=12510 - 1
http://www.medicinenet.com/script/main/art.asp?articlekey=12578 - 1
http://www.webmd.com/policies/about-privacy-policy - 1
http://www.medicinenet.com/script/main/art.asp?articlekey=8709&page=3#risks - 1
http://www.medicinenet.com/script/main/forum.asp?articlekey=8709&articletype=proc - 1
http://www.medicinenet.com/slideshows-popular/article.htm - 1
http://www.medicinenet.com/script/main/art.asp?articlekey=157501 - 1
http://www.medicinenet.com/menopause/focus.htm - 1
http://www.medicinenet.com/script/main/art.asp?articlekey=107539 - 1
http://www.webmd.boots.com/ - 1
http://images.medicinenet.com/css/medicinenet/master/2.0/master.min.css - 1
http://www.medicinenet.com/script/main/art.asp?articlekey=8709&page=3#alcoholism - 1
http://www.medicinenet.com/prevention_and_wellness/article.htm - 1
http://www.medicinenet.com/script/main/forum.asp?articlekey=8709&articletype=img - 1
http://www.medicinenet.com/script/main/art.asp?articlekey=9090 - 1
http://www.medicinenet.com/script/main/art.asp?articlekey=359 - 1
http://www.emedicinehealth.com/script/main/hp.asp - 1
http://www.medicinenet.com/script/main/art.asp?articlekey=8709&page=6#stages - 1

http://www.medicinenet.com/script/main/art.asp?articlekey=8709&page=3#abuse - 1

8 http://www.alcoholics-anonymous.org.uk/newcomers/?PageID=69

| Meta tags report for: http://www.alcoholics-anonymous.org.uk/newcomers/?PageID=69 | | |
|---|---|---|
| meta tag | length | value |
| Title: | 41 | Alcoholics Anonymous (A.A.) Great Britain |
| Description: | 23 | Alcoholics Anonymous GB |
| Keywords: | 123 | AA, Alcoholics Anonymous, GSB, General Service Board, meetings, alcoholism, recovery, drink problem, drinking problem, A.A. |

| Headers returned from: http://www.alcoholics-anonymous.org.uk/newcomers/?PageID=69 | |
|---|---|
| Related Keyword found (number of times and density): | **alcoholism** - 7 - 1.03%<br>**alcoholic** - 6 - 0.88%<br>**alcohol** - 4 - 0.59%<br>**alcoholics** - 3 - 0.44% |
| x-meta-title: | Alcoholics Anonymous (A.A.) Great Britain |
| x-meta-description: | Alcoholics Anonymous GB |
| link: | <../section-newcomers.css>; rel="stylesheet"; type="text/css" |
| x-powered-by: | ASP.NET |
| connection: | close |
| client-response-num: | 1 |
| set-cookie: | CFID=509496;expires=Mon, 28-Apr-2042 20:17:26 GMT;path=/ |
| date: | Sat, 05 May 2012 20:17:26 GMT |
| x-meta-keywords: | AA, Alcoholics Anonymous, GSB, General Service Board, meetings, alcoholism, recovery, drink problem, drinking problem, A.A. |
| client-peer: | 80.82.141.157:80 |
| client-date: | Sat, 05 May 2012 20:17:22 GMT |
| content-type: | text/html; charset=UTF-8 |
| content-language: | en-GB |
| title: | Alcoholics Anonymous (A.A.) Great Britain |
| server: | Microsoft-IIS/6.0 |

**Keywords found in the Anchor tags: Keyword - Times Found.**
These are text links on your web page (include the 'alt' text from images in the links). These become more important by many search engines (for best results try to name them after your primary keywords).

**alcoholics anonymous** - 1

**about alcoholism** - 1

**Keywords found in the IMG Alt tags: Keyword - Times Found.**
This is text found in the 'alt' tag from the images. For web pages with a lot of images those tags are important (for best results try to name them after your primary keywords).

**alcoholics anonymous** - 1

**menu bottom** - 1

**menu top** - 1

**URLs found in the page: URL - Times Found**
Found 26 urls from where 25 unique.

http://www.alcoholics-anonymous.org.uk/newcomers/../index.cfm - 2
http://www.alcoholics-anonymous.org.uk/newcomers/?PageID=76 - 1
http://www.alcoholics-anonymous.org.uk/newcomers/?PageID=2 - 1
http://www.alcoholics-anonymous.org.uk/newcomers/?PageID=68 - 1
http://www.alcoholics-anonymous.org.uk/newcomers/../?PageID=2 - 1
http://www.alcoholics-anonymous.org.uk/newcomers/?PageID=78 - 1
http://www.alcoholics-anonymous.org.uk/newcomers/../?PageID=43 - 1
http://www.alcoholics-anonymous.org.uk/newcomers/../?PageID=4 - 1
http://www.alcoholics-anonymous.org.uk/newcomers/?PageID=82 - 1
http://www.alcoholics-anonymous.org.uk/ - 1
http://www.alcoholics-anonymous.org.uk/?PageID=2 - 1
http://www.alcoholics-anonymous.org.uk/newcomers/ - 1
http://www.alcoholics-anonymous.org.uk/newcomers/?PageID=69 - 1
http://www.alcoholics-anonymous.org.uk/newcomers/?PageID=71 - 1
http://www.alcoholics-anonymous.org.uk/newcomers/?PageID=72 - 1
http://www.alcoholics-anonymous.org.uk/?PageID=4 - 1
http://www.alcoholics-anonymous.org.uk/newcomers/?PageID=79 - 1
http://www.alcoholics-anonymous.org.uk/?PageID=43 - 1
http://www.alcoholics-anonymous.org.uk/newcomers/?PageID=101 - 1
http://www.alcoholics-anonymous.org.uk/?PageID=3 - 1
http://www.alcoholics-anonymous.org.uk/newcomers/?PageID=47 - 1
http://www.alcoholics-anonymous.org.uk/newcomers/?PageID=70 - 1
http://www.alcoholics-anonymous.org.uk/newcomers/../?PageID=3 - 1
http://www.alcoholics-anonymous.org.uk/newcomers/?PageID=243 - 1
http://www.alcoholics-anonymous.org.uk/newcomers/../section-newcomers.css - 1

9  http://www.tandf.co.uk/journals/titles/07347324.asp

| Meta tags report for: http://www.tandf.co.uk/journals/titles/07347324.asp | | |
|---|---|---|
| **meta tag** | **length** | **value** |
| Title: | 34 | Taylor & Francis Journals: Welcome |

| Keywords returned from: http://www.tandf.co.uk/journals/titles/07347324.asp | |
|---|---|
| Related Keyword found<br><br>(number of times and density): | **alcoholism** - 9 - 2.14%<br>**alcoholic** - 4 - 0.95% |
| x-meta-description: | |
| link: | ; /="/"; rel="stylesheet"; type="text/css" |
| x-powered-by: | ASP.NET |
| connection: | close |
| client-response-num: | 1 |
| cache-control: | private |
| set-cookie: | ASPSESSIONIDQARATDCB=KCOMNFICNJABMCABEIFBBMML; path=/ |
| date: | Sat, 05 May 2012 23:34:04 GMT |
| x-meta-keywords: | |
| client-peer: | 213.212.74.245:80 |
| client-date: | Sat, 05 May 2012 23:33:59 GMT |
| content-type: | text/html |
| title: | Taylor & Francis Journals: Welcome |
| server: | Microsoft-IIS/6.0 |

| Keywords found in the Anchor tags: Keyword - Times Found. |
|---|
| These are text links on your web page (include the 'alt' text from images in the links). These become more important by many search engines for best results try to name them after your primary keywords). |

| **alcoholism treatment quarterly** - 1 |
|---|

| Keywords found in the IMG Alt tags: Keyword - Times Found. |
|---|

This is text found in the 'alt' tag from the images. For web pages with a lot of images those tags are important (for best results try to name them after your primary keywords).

**taylor francis journals welcome** - 2

**journals resources** - 1

**printer friendly page** - 1

**begin search** - 1

**author resources** - 1

**top** - 1

**related websites** - 1

**alcoholism treatment quarterly** - 1

**journal listings** - 1

URLs found in the page: URL - Times Found
Found 60 urls from where 57 unique.

http://www.tandf.co.uk/ - 2
http://www.tandf.co.uk/journals/customer.asp - 2
http://www.tandf.co.uk/journals/titles/07347324.asp#top - 2
http://www.tandfonline.com/WATQ - 1
- 1
http://www.tandf.co.uk/journals/ethics.asp - 1
http://www.tandf.co.uk/journals/ifa.asp - 1
http://www.tandf.co.uk/journals/new-journals.asp - 1
http://www.tandf.co.uk/journals/permissions.asp - 1
http://www.taylorandfrancis.com - 1
http://www.tandf.co.uk/journals/onlinesamples.asp - 1
http://www.tandf.co.uk/css/master.css - 1
http://www.tandf.co.uk/journals/journal.asp?issn=0734-7324&linktype=1 - 1
http://www.tandf.co.uk/addiction-abs - 1
http://www.tandf.co.uk/journals/offers.asp - 1
http://www.tandf.co.uk/journals/printview?issn=0734-7324 - 1
http://www.tandf.co.uk/journals/arenas.asp - 1
http://www.tandf.co.uk/journals/developingworld.asp - 1
http://www.tandf.co.uk/jcp - 1
http://www.tandf.co.uk/journals/online.asp - 1
http://www.tandf.co.uk/journals/pr.asp - 1
http://www.tandf.co.uk/journals/iopenaccess.asp - 1
http://www.tandf.co.uk/era - 1
http://www.tandf.co.uk/journals/access/ASAM2012.pdf - 1
http://www.tandf.co.uk/journals/contact.asp - 1
http://www.tandf.co.uk/journals/special.asp - 1
http://www.informaworld.com/journals_reprints - 1
http://www.tandf.co.uk/journals/recommend/journals.asp - 1
http://journalauthors.tandf.co.uk - 1
http://www.tandf.co.uk/libsite - 1
http://www.tandf.co.uk/journals/terms.asp - 1
http://www.tandf.co.uk/journals/sublist.asp - 1
http://www.tandf.co.uk/journals/catalogue/index.asp - 1
http://www.tandf.co.uk/swa - 1
http://www.tandf.co.uk/journals/alphalist.asp - 1
http://journalauthors.tandf.co.uk/preparation/copyright.asp - 1
http://www.tandf.co.uk/journals/ifirst.asp - 1
http://www.taylorandfrancisgroup.com/careers/ - 1
http://www.tandf.co.uk/journals/eupdates.asp - 1
http://www.tandf.co.uk/journals/publish.asp - 1
http://www.tandf.co.uk/journals/privacy.asp - 1
http://www.tandf.co.uk/journals/journal.asp?issn=0734-7324&linktype=145 - 1
http://journalauthors.tandf.co.uk/othercategories/authornewsletter.asp - 1

http://www.tandf.co.uk/journals/advertising.asp - 1
http://www.tandf.co.uk/journals/society - 1
http://www.tandf.co.uk/ergo-abs - 1
http://www.tandf.co.uk/journals/alerting.asp - 1
http://www.tandf.co.uk/journals/members.asp - 1
http://www.garlandscience.com - 1
http://www.tandf.co.uk/journals/journal.asp?issn=0734-7324&linktype=44 - 1
http://www.ebookstore.tandf.co.uk - 1
http://www.routledge.com - 1
http://www.taylorandfrancisgroup.com/ - 1
http://www.tandfonline.com/action/pricing?journalCode=watq - 1
http://www.tandf.co.uk/journals/favicon/favicon.ico - 1
http://www.tandf.co.uk/journals/subscription.asp - 1
http://www.tandf.co.uk/journals/journal.asp?issn=0734-7324&linktype=2 - 1

10 http://www.nhs.uk/news/2012/03march/Pages/lsd-acid-alcoholism-treatment.aspx

| Meta tags report for: http://www.nhs.uk/news/2012/03march/Pages/lsd-acid-alcoholism-treatment.aspx | | |
|---|---|---|
| **meta tag** | **length** | **value** |
| Title: | 63 | Could an acid trip stop alcoholism? - Health News - NHS Choices |
| Description : | 228 | LSD â€œhelps alcoholics to give up drinkingâ€, BBC News has today reported. This unusual claim is based on a review examining research into the powerful hallucinogen and its potential to treat alcoholism. The review analysed... |
| Keywords: | 103 | National Health Service (NHS),acid,trip,LSD,1960s,health news,medical news,alcohol,alcoholism,treatment |

| Headers returned from: http://www.nhs.uk/news/2012/03march/Pages/lsd-acid-alcoholism-treatment.aspx | |
|---|---|
| Related Keyword found (number of times and density): | **alcohol** - 22 - 0.70%<br>**alcoholism** - 17 - 0.54%<br>**alcoholic** - 3 - 0.10%<br>**alcoholics** - 3 - 0.10%<br>**alcohol-focused** - 1 - 0.03%<br>â€œalcohol - 1 - 0.03% |
| x-meta-dc.creator: | NHS Choices |
| x-meta-dc.description: | LSD â€œhelps alcoholics to give up drinkingâ€, BBC News has today reported. This unusual claim is based on a review examining research into the powerful hallucinogen and its potential to treat alcoholism. The review analysed... |
| set-cookie: | cookie=R2817298101; path=/ |
| x-meta-dcsext.server: | NHC10WEB12PRP |
| date: | Sat, 05 May 2012 23:49:44 GMT |
| x-meta-keywords: | National Health Service (NHS),acid,trip,LSD,1960s,health news,medical news,alcohol,alcoholism,treatment |
| vary: | * |
| x-meta-dc.format: | text/html |
| client-date: | Sat, 05 May 2012 23:49:13 GMT |
| x-meta-dc.rights: | http://www.nhs.uk/termsandconditions/Pages/TermsConditions.aspx |
| x-meta-dc.subject: | National Health Service (NHS),acid,trip,LSD,1960s,health news,medical news,alcohol,alcoholism,treatment |
| server: | Microsoft-IIS/7.5 |
| x-meta-wt.cg-n: | News |
| x-meta-description: | LSD â€œhelps alcoholics to give up drinkingâ€, BBC News has today reported. This unusual claim is based on a review examining research into the powerful hallucinogen and its potential to treat alcoholism. The review analysed... |
| x-meta-dc.coverage: | England |
| x-meta-egms.accessibility: | Double-A |
| x-meta-dc.date.issued: | 2012-14-03 |
| content-length: | 108500 |
| x-meta-dc.title: | Could an acid trip stop alcoholism? - Health News - NHS Choices |
| x-meta-dc.publisher: | Department of Health |
| title: | Could an acid trip stop alcoholism? - Health News - NHS Choices |
| x-meta-wt.sv: | NHC10WEB12PRP |
| cache-control: | private |
| client-peer: | 217.64.234.65:80 |
| x-meta-dcsext.realurl: | /news/2012/03march/Pages/lsd-acid-alcoholism-treatment.aspx |
| content-type: | text/html; charset=utf-8 |
| x-meta-wt.ti: | Could an acid trip stop alcoholism? - Health News - NHS Choices |
| link: | ; /="/"; rel="meta"; title="ICRA labels"; type="application/rdf+xml" |
| pics-label: | (pics-1.1 "http://www.icra.org/pics/vocabularyv03/" l gen true for "http://nhs.uk" r (n 2 s 1 v 0 1 0 oa 0 ob 0 oc 0 od 0 oe 0 of 0 og 0 oh 0 c 1) gen true for "http://www.nhs.uk" r (n 2 s 1 v 0 1 0 oa 0 ob 0 oc 0 od 0 oe 0 of 0 og 0 oh 0 c 1)) |
| x-powered-by: | UrlRewriter.NET 1.8.0 |
| client-response-num: | 1 |
| x-meta-dc.language: | eng |
| x-ua-compatible: | IE=8 |
| x-meta-wt.cg-s: | 2012 |
| x-meta-dc.identifier: | http://www.nhs.uk |
| expires: | Sun, 06 May 2012 00:49:44 GMT |

**Keywords found in the Anchor tags: Keyword - Times Found.**
These are text links on your web page (include the 'alt' text from images in the links). These become more important by many search engines (for best results try to name them after your primary keywords).

alcohol search - 1

getting support for alcohol problems - 1

can lsd cure alcoholism trials show 59 per cent of problem drinkers improve

caring for my alcoholic wife - 1

caring for an alcoholic - 1

alcohol articles - 1

lsd helps alcoholics to give up drinking - 1

3 comments about could an acid trip stop alcoholism health news nhs choices - 1

lysergic acid diethylamide lsd for alcoholism meta-analysis of randomized s - 1

lsd could treat alcoholism because trips make you reassess addiction - 1

**Keywords found in the IMG Alt tags: Keyword - Times Found.**
This is text found in the 'alt' tag from the images. For web pages with a lot of images those tags are important (for best results try to name them after your primary keywords).

facebook share - 2

google bookmarks - 2

windows live messenger share - 2

rss feed - 2

twitter share - 2

nhs choices saved pages - 2

email share - 2

link to directgov public services all in one place - 1

go to nhs choices homepage - 1

the information standard certified member - 1

bazian source image - 1

its your choice - 1

| URLs | found | in | the | page: | URL | - | Times | Found |
|------|-------|-----|-----|-------|-----|---|-------|-------|

Found 209 urls from where 189 unique.

http://www.nhs.uk/css/bth.css - 3
http://www.nhs.uk/css/screen.css - 2
- 2
http://www.nhs.uk/tools/pages/toolslibrary.aspx - 2
https://www.google.com/bookmarks/mark?op=add&bkmk=http%3a%2f%2fwww.nhs.uk%2fnews%2f2012%2f03march%2fPages%2flsd-acid-alcoholism-treatment.aspx&title=Could+an+acid+trip+stop+alcoholism%3f+-+Health+News+-+NHS+Choices&WT.mc_id=30411 - 2
http://www.nhs.uk/NHSChoices/shared/RSSFeedGenerator/RSSFeed.aspx?site=News - 2
http://www.nhs.uk/video/pages/MediaLibrary.aspx - 2
http://www.nhs.uk/carersdirect - 2
http://www.facebook.com/sharer.php?u=http%3a%2f%2fwww.nhs.uk%2fnews%2f2012%2f03march%2fPages%2flsd-acid-alcoholism-treatment.aspx&t=Could+an+acid+trip+stop+alcoholism%3f+-+Health+News+-+NHS+Choices&WT.mc_id=60411 - 2
http://www.nhs.uk/news/2012/03march/Pages/lsd-acid-alcoholism-treatment.aspx?savefavourite=true - 2
http://www.chooseandbook.nhs.uk/ - 2
http://www.nhs.uk/Personalisation/Registration.aspx?ReturnUrl=http%3a%2f%2fwww.nhs.uk%2fnews%2f2012%2f03march%2fPages%2flsd-acid-alcoholism-treatment.aspx - 2
http://www.nhs.uk/servicedirectories/Pages/ServiceSearch.aspx - 2
http://www.nhs.uk/accessibility/Pages/Accessibility.aspx - 2
http://profile.live.com/badge/?url=http%3a%2f%2fwww.nhs.uk%2fnews%2f2012%2f03march%2fPages%2flsd-acid-alcoholism-treatment.aspx&title=Could+an+acid+trip+stop+alcoholism%3f+-+Health+News+-+NHS+Choices&screenshot=http://www.nhs.uk/img/social-sharing/nhsc-wl-icn.jpg&WT.mc_id=40411 - 2
http://www.nhs.uk/aboutNHSChoices/Pages/ContactUs.aspx - 2
http://www.nhs.uk/Personalisation/Login.aspx - 2
http://twitter.com/home?status=Could+an+acid+trip+stop+alcoholism%3f+-+Health+News+-+NHS+Choices%20-%20http%3a%2f%2fwww.nhs.uk%2fnews%2f2012%2f03march%2fPages%2flsd-acid-alcoholism-treatment.aspx&WT.mc_id=50411 - 2
http://www.nhs.uk/Personalisation/Registration.aspx?ReturnUrl= - 1
http://www.youtube.com/nhschoices - 1
http://www.nhs.uk/news/2012/05may/Pages/warfarin-aspirin-anti-bloodclot-trial.aspx - 1
http://www.nhs.uk/News/Pages/NewsArticles.aspx?TopicId=Mental+health - 1
http://www.nhs.uk/news/pages/newsarticles.aspx?TopicID=QA+articles - 1
http://www.nhs.uk/conditions/pregnancy-and-baby/pages/pregnancy-and-baby-care.aspx - 1
http://www.talktofrank.com/drug/lsd - 1
http://www.aboutNHSChoices/aboutnhschoices/termsandconditions/Pages/termsandconditions.aspx - 1
http://www.nhs.uk/Conditions/Arthritis - 1
http://www.nhs.uk/servicedirectories/Pages/ServiceSearchAdditional.aspx?ServiceType=StopSmokingService - 1
http://www.nhs.uk/news/2012/01january/pages/psilocybin-mushroom-brain-scans.aspx - 1
http://www.nhs.uk/conditions/asthma - 1
http://www.nhs.uk/NHSEngland/links/Pages/links-make-it-happen.aspx - 1
http://www.show.scot.nhs.uk/ - 1
http://www.nhs.uk/servicedirectories/Pages/ServiceSearch.aspx?ServiceType=Dentist - 1
http://www.nhs.uk/css/if-ie6.css - 1
http://www.nhs.uk/css/print.css - 1
http://www.nhs.uk/news/2012/03march/Pages/lsd-acid-alcoholism-treatment.aspx#commentCountLink - 1
http://www.aboutNHSChoices/professionals/syndication/Pages/Webservices.aspx - 1
http://www.nhs.uk/news - 1
http://www.nhs.uk/news/pages/newsarticles.aspx?TopicId=Lifestyle%2fexercise - 1
http://www.nhs.uk/Livewell/alcohol/Pages/Caringforanalcoholic.aspx - 1
http://www.facebook.com/nhschoices - 1
http://www.nhs.uk/includes/labels.rdf - 1

http://www.nhs.uk/livewell/smoking - 1
http://www.nhs.uk/css/reset-ie7.css - 1
http://www.nhsdirect.wales.nhs.uk/ - 1
http://www.dh.gov.uk/en/index.htm - 1
http://www.nhs.uk/News/Pages/NewsArticles.aspx?TopicId=Food%2fdiet - 1
http://www.nhs.uk/Planners/vaccinations/Pages/Landing.aspx - 1
http://www.nhs.uk/News/Pages/NewsArticles.aspx?TopicId=Cancer - 1
http://www.nhs.uk/CarersDirect/yourself/timeoff/Pages/Accessingrespitecare.aspx - 1
http://www.nhs.uk/news/2012/03march/Pages/lsd-acid-alcoholism-treatment.aspx#footer-tab2 - 1
http://www.nhs.uk/CarersDirect/guide - 1
http://www.nhs.uk/news/pages/newsarticles.aspx?TopicId=Cancer - 1
http://www.nhs.uk/livewell/drugs/pages/dodrugsdamagebrain.aspx - 1
http://www.nhs.uk/Commentspolicy/Pages/Moderationrules.aspx - 1
http://www.nhs.uk/CarersDirect/moneyandlegal/carersbenefits/Pages/Overview.aspx - 1
http://www.nhs.uk/Conditions/Diabetes-type2 - 1
http://www.telegraph.co.uk/science/science-news/9131181/LSD-could-treat-alcoholism-because-trips-make-you-reassess-addiction.html - 1
http://www.nhs.uk/news/pages/newsarticles.aspx?TopicId=Genetics%2fstem+cells - 1
http://www.nhs.uk/CarersDirect/workandlearning - 1
http://www.nhs.uk/aboutNHSChoices/aboutnhschoices/Aboutus/Pages/LinkingtoChoices.aspx - 1
http://www.nhs.uk/aboutNHSChoices/aboutnhschoices/termsandconditions/Pages/commentspolicy.aspx - 1
http://www.nhs.uk/CarersDirect/moneyandlegal - 1
http://www.nhs.uk/choices/pages/sitemap.aspx - 1
http://www.nhs.uk/news/pages/newsarticles.aspx?TopicId=Heart%2flungs - 1
http://www.nhs.uk/Livewell/Hayfever - 1
http://www.nhs.uk/News/Pages/NewsArticles.aspx?TopicId=Older+people - 1
http://www.nhs.uk/conditions/measles - 1
http://www.nhs.uk/livewell/mentalhealth - 1
http://m.direct.gov.uk/categoriesController?action=category&id=9 - 1
http://www.nhs.uk/news/pages/newsglossary.aspx#peerreview - 1
http://www.nhs.uk/news/2012/03march/Pages/../../../../Pages/comments.aspx?contentId=22702&AreaId=2 - 1
http://www.nhs.uk/servicedirectories/Pages/ServiceSearch.aspx?ServiceType=Hospital&ServiceName=Maternity - 1
http://www.nhs.uk/livewell/Pages/Livewellhub.aspx - 1
http://www.nhs.uk/news/pages/newsarticles.aspx?TopicId=Medication - 1
http://www.nhs.uk/news/pages/newsarticles.aspx?TopicId=Older+people - 1
http://www.nhs.uk/CarersDirect/guide/assessments/Pages/Carersassessments.aspx - 1
http://www.nhs.uk/Pages/LinkListing.aspx - 1
http://www.nhs.uk/conditions/whooping-cough/pages/introduction.aspx - 1
http://www.nhs.uk/News/Pages/NewsArticles.aspx?TopicId=Medication - 1
http://www.nhs.uk/NHSEngland/AboutNHSservices/Pages/NHSServices.aspx - 1
http://www.nhs.uk/LiveWell/c25k/Pages/couch-to-5k.aspx - 1
http://www.nhs.uk/css/if-ie7.css - 1
http://www.nhs.uk/CarersDirect/young - 1
http://www.nhs.uk/livewell/drugs/pages/drugsoverview.aspx - 1
http://www.nhs.uk/pages/languagehub.aspx - 1
http://www.nhs.uk/news/2012/04april/Pages/mobile-phone-cancer-brain-tumour-evidence.aspx - 1
http://www.nhs.uk/CarersDirect/yourself - 1
http://www.aboutNHSChoices/aboutnhschoices/Aboutus/Pages/languageshub.aspx - 1
http://www.nhs.uk/news/2012/03march/Pages/../../../../Pages/comments.aspx?contentId=22703&AreaId=2 - 1
http://www.nhs.uk/livewell/tiredness-and-fatigue - 1
http://www.nhs.uk/news/pages/newsarticles.aspx?TopicId=Obesity - 1
http://www.nhs.uk/NHSEngland/thenhs/records/proms/Pages/aboutproms.aspx - 1
http://www.nhs.uk/CHQ/Pages/home.aspx - 1
http://www.nhs.uk/livewell/pages/topics.aspx - 1
http://www.nhscarerecords.nhs.uk/summary/ - 1
http://www.nhs.uk/News/Pages/NewsArticles.aspx?TopicId=Pregnancy%2fchild - 1
http://www.nhs.uk/img/favicon.ico - 1
http://www.nhs.uk/NHSEngland/thenhs/nhshistory/Pages/NHShistory1948.aspx - 1
http://www.nhs.uk/news/pages/newsglossary.aspx#Randomisedcontrolledtrial(RCT) - 1
http://www.nhs.uk/livewell/loseweight - 1
http://www.nhsdirect.nhs.uk/ - 1
http://www.nhs.uk/Personalisation/Login.aspx?ReturnUrl=http%3a%2f%2fwww.nhs.uk%2fnews%2f2012%2f03march%2fPages%2flsd-acid-alcoholism-treatment.aspx -1
http://www.nhs.uk/Conditions/Cancer-of-the-testicle - 1
http://www.nhs.uk/servicedirectories/Pages/ServiceSearch.aspx?ServiceType=AandE - 1
http://www.nhs.uk/news/pages/newsarticles.aspx?TopicId=Pregnancy%2fchild - 1
http://www.nhs.uk/Conditions/Back-pain - 1
http://www.nhs.uk/conditions/kidney-infection - 1
http://www.hscni.net/ - 1
http://www.nhs.uk/news/pages/newsarticles.aspx?TopicId=Medical+practice - 1
http://www.nhs.uk/servicedirectories/Pages/ServiceSearchAdditional.aspx?ServiceType=Mentalhealth - 1
http://www.nhs.uk/Conditions/Pages/hub.aspx - 1

http://www.nhs.uk/Conditions/menopause - 1
http://www.nhs.uk/Pages/HomePage.aspx - 1
http://www.nhs.uk/livewell/Summerhealth/Pages/Summerhealthhome.aspx - 1
http://www.nhs.uk/css/base.css - 1
http://www.nhs.uk/news/2012/03march/Pages/lsd-acid-alcoholism-treatment.aspx#footer-tab3 - 1
http://www.nhs.uk/news/2012/05may/Pages/gmc-medication-prescribing-errors-report.aspx - 1
http://www.nhs.uk/ - 1
http://www.nhs.uk/livewell/healthy-eating - 1
http://www.nhs.uk/choiceinthenhs/Pages/choicehome.aspx - 1
http://www.nhs.uk/News - 1
http://www.nhs.uk/Conditions/Coronary-heart-disease - 1
http://www.nhs.uk/News/Pages/NewsArticles.aspx?TopicId=Heart%2flungs - 1
http://www.nhs.uk/servicedirectories/Pages/ServiceSearch.aspx?ServiceType=Optician - 1
http://www.nhs.uk/servicedirectories/Pages/ServiceSearch.aspx?ServiceType=WalkInCentre - 1
http://www.nhs.uk/News/Pages/NewsArticles.aspx?TopicId=Obesity - 1
http://www.nhs.uk/Conditions/Pages/BodyMap.aspx?Index=A - 1
http://www.nhschoicestraining.co.uk - 1
https://www.healthspace.nhs.uk - 1
http://www.nhs.uk/Conditions/Cancer-of-the-breast-female - 1
http://www.nhs.uk/Livewell/alcohol - 1
http://www.nhs.uk/news/pages/newsarticles.aspx?TopicId=Food%2fdiet - 1
http://www.nhs.uk/servicedirectories/Pages/ServiceSearch.aspx?ServiceType=GP - 1
http://www.nhs.uk/Conditions/eczema-(atopic) - 1
http://www.nhs.uk/livewell/fitness - 1
http://www.nhs.uk/aboutNHSChoices/aboutnhschoices/Personalaccounts/Pages/NHSChoicesaccount.aspx - 1
http://www.nhs.uk/news/pages/newsglossary.aspx#oddsratio - 1
http://www.nhs.uk/news/2012/04april/Pages/autism-neurotransmitter-drug-clue.aspx - 1
http://www.nhs.uk/Livewell/alcohol/Pages/Caringforanalcoholicrealstory.aspx - 1
http://www.nhs.uk/servicedirectories/Pages/ServiceSearch.aspx?ServiceType=Pharmacy - 1
http://twitter.com/nhschoices - 1
http://www.nhs.uk/news/2012/03march/Pages/lsd-acid-alcoholism-treatment.aspx#mainContent - 1
http://www.nhs.uk/servicedirectories/Pages/ServiceSearch.aspx?ServiceType=Hospital - 1
http://www.nhs.uk/news/2012/03march/Pages/lsd-acid-alcoholism-treatment.aspx#footer-tab1 - 1
http://www.nhs.uk/News/Pages/NewsArticles.aspx?TopicId=Special+reports - 1
http://www.dailymail.co.uk/sciencetech/article-2111687/Theres-consistent-beneficial-effect-LSD-treating-alcoholics-say-researchers.html - 1
http://www.nhs.uk/servicedirectories/Pages/ServiceSearchAdditional.aspx?ServiceType=SexualHealthService - 1
http://www.nhs.uk/news/pages/newsarticles.aspx?TopicId=Special+reports - 1
http://www.nhscareers.nhs.uk/ - 1
http://www.nhs.uk/News/Pages/NewsArticles.aspx?TopicId=QA+articles - 1
http://www.nhs.uk/news/2012/03march/Pages/lsd-acid-alcoholism-treatment.aspx#main-navigation - 1
http://www.nhs.uk/medicine-guides/pages/default.aspx - 1
http://www.nhs.uk/News/Pages/NewsArticles.aspx?TopicId=Lifestyle%2fexercise - 1
http://www.nhs.uk/news/2012/03march/Pages/../../../Pages/comments.aspx?contentId=22784&AreaId=2 - 1
http://www.nhs.uk/aboutNHSChoices/aboutnhschoices/Aboutus/Pages/Editorialpolicy.aspx - 1
http://www.nhs.uk/aboutNHSChoices/aboutnhschoices/termsandconditions/Pages/Privacypolicy.aspx - 1
http://www.bbc.co.uk/news/health-17297714 - 1
http://www.nhs.uk/aboutNHSChoices/Pages/AboutNHSChoices.aspx - 1
http://www.nhs.uk/aboutNHSChoices/aboutnhschoices/Aboutus/Pages/the-information-standard.aspx - 1
http://www.direct.gov.uk/ - 1
http://www.nhs.uk/servicedirectories/Pages/ServiceSearch.aspx?ServiceType=Hospital&ServiceMode=Consultant - 1
http://www.nhs.uk/News/Pages/NewsArticles.aspx?TopicId=Medical+practice - 1
http://www.nhs.uk/aboutNHSChoices/professionals/healthandcareprofessionals/quality-accounts/Pages/quality-accounts-2010-2011.aspx - 1
http://www.nhs.uk/CarersDirect/carers-learning-online/Pages/Welcome.aspx - 1
http://www.nhs.uk/News/Pages/NewsArticles.aspx?TopicId=Diabetes - 1
http://www.nhs.uk/News/Pages/NewsArticles.aspx?TopicId=Genetics%2fstem+cells - 1
http://www.nhs.uk/nhsdirect/pages/symptoms.aspx - 1
http://www.nhs.uk/news/pages/newsglossary.aspx#confidenceinterval - 1
http://www.nhs.uk/css/reset-ie8.css - 1
http://www.healthunlocked.com/nhschoices/ - 1
http://www.nhs.uk/news/pages/newsarticles.aspx?TopicId=Diabetes - 1
http://www.nhs.uk/conditions/rickets/pages/introduction.aspx - 1
http://www.nhs.uk/News/Pages/NewsArticles.aspx?TopicId=Neurology - 1
http://www.nhs.uk/Livewell/alcohol/Pages/Alcoholsupport.aspx - 1
http://www.nhs.uk/news/2012/05may/Pages/girls-put-off-exercise-school-sport.aspx - 1
http://www.nhs.uk/CarersDirect/guide/rights/Pages/carers-rights.aspx - 1
http://www.nhs.uk/choiceinthenhs - 1
http://www.nhs.uk/news/pages/newsarticles.aspx?TopicId=Neurology - 1
http://www.nhs.uk/aboutnhschoices/aboutnhschoices/nhschoicesmobile/pages/nhschoicesmobile.aspx - 1
http://www.nhs.uk/aboutNHSChoices/aboutnhschoices/e-newsletters/Pages/newsletters-home.aspx - 1
http://www.nhs.uk/News/Pages/NewsIndex.aspx - 1

http://www.nhs.uk/news/pages/newsarticles.aspx?TopicId=Mental+health - 1
http://www.nhs.uk/servicedirectories/Pages/ServiceSearchAdditional.aspx?ServiceType=Alcohol - 1
http://jop.sagepub.com/content/early/2012/03/08/0269881112439253.abstract - 1
http://www.nhs.uk/css/if-ie8.css - 1
http://www.nhs.uk/css/reset-ie6.css - 1
http://www.nhs.uk/livewell/sexualhealth - 1
http://www.nhs.uk/News/Pages/NewsArticles.aspx?TopicId=Swine+flu - 1

## 2. Bing SERP Analysis:

Other Documents are attached in CD provided.

# Appendix 5: Test Case Data

```php
<?php
/**
 * PHPUnit
 *
 * Copyright (c) 2001-2012, PHPUnit test
 * All rights reserved.
 * @package    PHPUnit
 * @subpackage Extensions_PhptTestCase
 *
 */

if
(PHPUnit_Util_Filesystem::fileExistsInIncludePath('PEAR/RunTest.
php')) {
    $currentErrorReporting = error_reporting(E_ERROR | E_WARNING
| E_PARSE);
    require_once 'PEAR/RunTest.php';
    error_reporting($currentErrorReporting);
}

/**
 * Wrapper to run .phpt test cases.
 *
 * @package    PHPUnit
 * @subpackage Extensions_PhptTestCase
 * @since      Class available since Release 3.1.4
 */
class        PHPUnit_Extensions_PhptTestCase        implements
PHPUnit_Framework_Test, PHPUnit_Framework_SelfDescribing
{
    /**
     * The filename of the .phpt file.
     *
     * @var    string
     */
    protected $filename;

    /**
     * Options for PEAR_RunTest.
     *
     * @var    array
     */
    protected $options = array();

    /**
     * Constructs a test case with the given filename.
     *
     * @param  string $filename
     * @param  array  $options
     */
    public  function  __construct($filename,  array  $options  =
array())
```

```php
    {
        if (!is_string($filename)) {
            throw PHPUnit_Util_InvalidArgumentHelper::factory(1,
'string');
        }

        if (!is_file($filename)) {
            throw new PHPUnit_Framework_Exception(
              sprintf(
                'File "%s" does not exist.',
                $filename
              )
            );
        }

        $this->filename = $filename;
        $this->options  = $options;
    }

    /**
     * Counts  the  number  of  test  cases  executed  by
run(TestResult result).
     *
     * @return integer
     */
    public function count()
    {
        return 1;
    }

    /**
     * Runs  a  test  and  collects  its  result  in  a  TestResult
instance.
     *
     * @param  PHPUnit_Framework_TestResult $result
     * @param  array                        $options
     * @return PHPUnit_Framework_TestResult
     */
    public function run(PHPUnit_Framework_TestResult $result =
NULL, array $options = array())
    {
        if (!class_exists('PEAR_RunTest', FALSE)) {
            throw    new    PHPUnit_Framework_Exception('Class
PEAR_RunTest not found.');
        }

        if (isset($GLOBALS['_PEAR_destructor_object_list']) &&
            is_array($GLOBALS['_PEAR_destructor_object_list'])
&&
            !empty($GLOBALS['_PEAR_destructor_object_list'])) {
            $pearDestructorObjectListCount                    =
count($GLOBALS['_PEAR_destructor_object_list']);
        } else {
            $pearDestructorObjectListCount = 0;
```

```php
        }

        if ($result === NULL) {
            $result = new PHPUnit_Framework_TestResult;
        }

        $coverage = $result->getCollectCodeCoverageInformation();
        $options  = array_merge($options, $this->options);

        if (!isset($options['include_path'])) {
            $options['include_path'] = get_include_path();
        }

        if ($coverage) {
            $options['coverage'] = TRUE;
        } else {
            $options['coverage'] = FALSE;
        }

        $currentErrorReporting  =  error_reporting(E_ERROR   |
E_WARNING | E_PARSE);
        $runner                       = new PEAR_RunTest(new
PHPUnit_Extensions_PhptTestCase_Logger, $options);

        if ($coverage) {
            $runner->xdebug_loaded = TRUE;
        } else {
            $runner->xdebug_loaded = FALSE;
        }

        $result->startTest($this);

        PHP_Timer::start();

        $buffer = $runner->run($this->filename, $options);
        $time   = PHP_Timer::stop();

        error_reporting($currentErrorReporting);

        $base         = basename($this->filename);
        $path         = dirname($this->filename);
        $coverageFile  =  $path   .   DIRECTORY_SEPARATOR   .
str_replace(
                          '.phpt', '.xdebug', $base
                        );
        $diffFile          = $path  .  DIRECTORY_SEPARATOR  .
str_replace(
                          '.phpt', '.diff', $base
                        );
        $expFile           = $path  .  DIRECTORY_SEPARATOR  .
str_replace(
                          '.phpt', '.exp', $base
                        );
```

```
        $logFile              =  $path  .   DIRECTORY_SEPARATOR   .
str_replace(
                          '.phpt', '.log', $base
                      );
        $outFile              =  $path  .   DIRECTORY_SEPARATOR   .
str_replace(
                          '.phpt', '.out', $base
                      );
        $phpFile              =  $path  .   DIRECTORY_SEPARATOR   .
str_replace(
                          '.phpt', '.php', $base
                      );

        if (is_object($buffer) && $buffer instanceof PEAR_Error)
{
            $result->addError(
              $this,
              new RuntimeException($buffer->getMessage()),
              $time
            );
        }

        else if ($buffer == 'SKIPPED') {
            $result->addFailure($this,                        new
PHPUnit_Framework_SkippedTestError, 0);
        }

        else if ($buffer != 'PASSED') {
            $expContent = file_get_contents($expFile);
            $outContent = file_get_contents($outFile);

            $result->addFailure(
              $this,
              new PHPUnit_Framework_ComparisonFailure(
                $expContent,
                $outContent,
                $expContent,
                $outContent
              ),
              $time
            );
        }

        foreach (array($diffFile, $expFile, $logFile, $phpFile,
$outFile) as $file) {
            if (file_exists($file)) {
                unlink($file);
            }
        }

        if ($coverage && file_exists($coverageFile)) {
            eval('$coverageData        =         '        .
file_get_contents($coverageFile) . ';');
            unset($coverageData[$phpFile]);
```

```
            $result->getCodeCoverage()->append($coverageData,
$this);
            unlink($coverageFile);
        }

        $result->endTest($this, $time);

        // Do not invoke PEAR's destructor mechanism for PHP 4
        // as it raises an E_STRICT.
        if ($pearDestructorObjectListCount == 0) {
            unset($GLOBALS['_PEAR_destructor_object_list']);
        } else {
            $count                                              =
count($GLOBALS['_PEAR_destructor_object_list'])               -
$pearDestructorObjectListCount;

            for ($i = 0; $i < $count; $i++) {
array_pop($GLOBALS['_PEAR_destructor_object_list']);
            }
        }

        return $result;
    }

    /**
     * Returns the name of the test case.
     *
     * @return string
     */
    public function getName()
    {
        return $this->toString();
    }

    /**
     * Returns a string representation of the test case.
     *
     * @return string
     */
    public function toString()
    {
        return $this->filename;
    }
}
```

## Appendix 6: Application Screen Shots

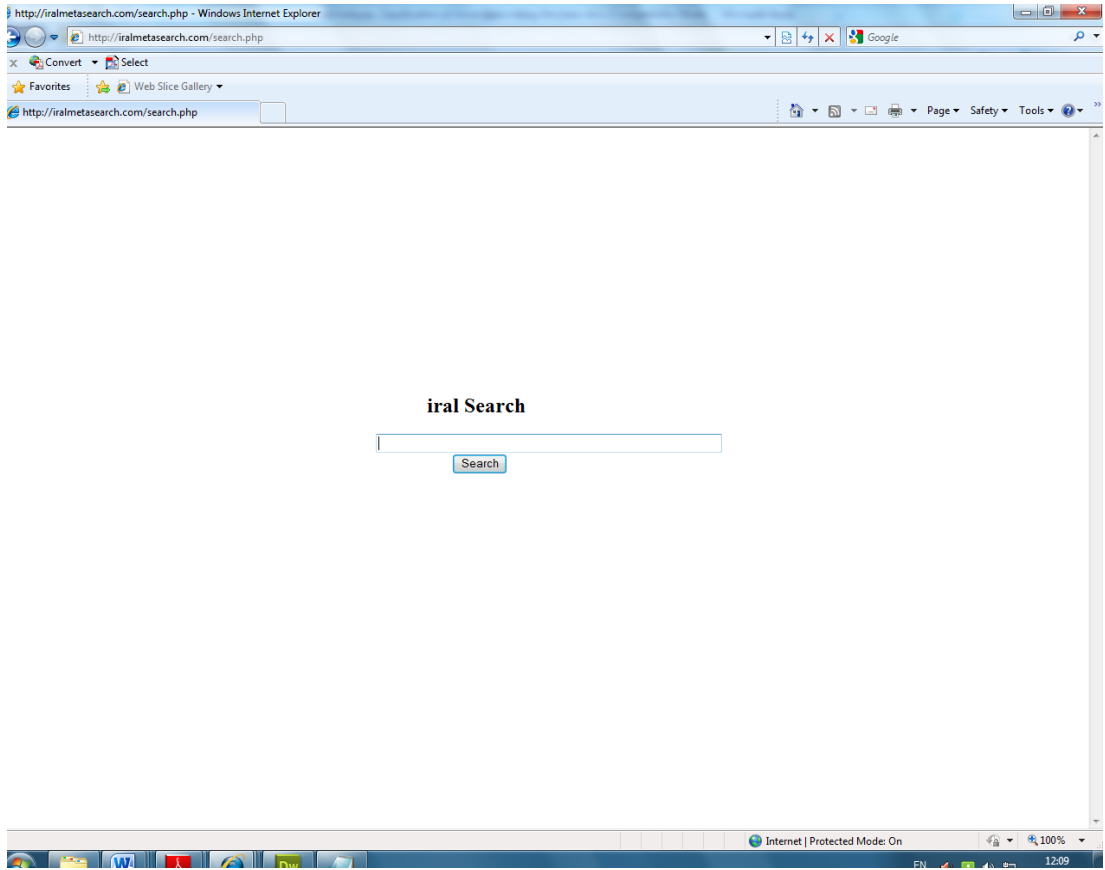

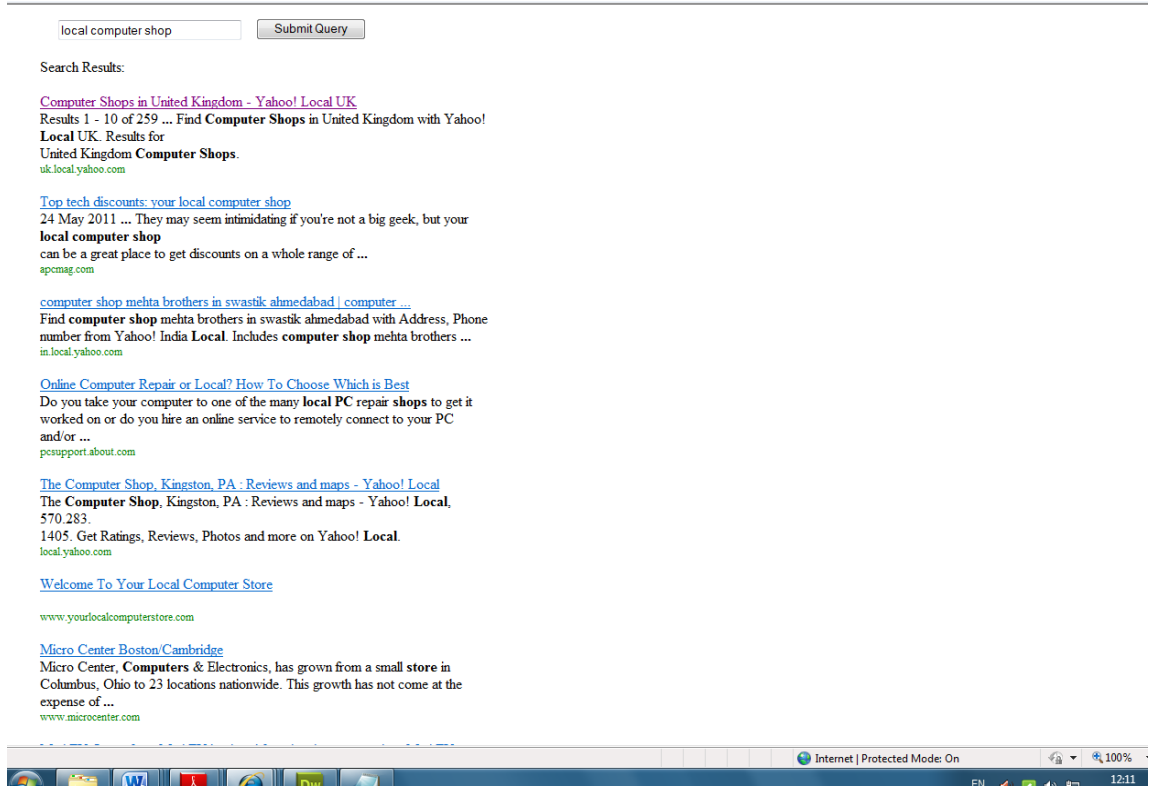

## Appendix 7: Legal and Ethical Issues

The Data Protection Act, 1998 identifies an individual's personal information consisting of his/her racist or ethnic origin, political opinions, religious beliefs and/or sexual life as *sensitive personal data*. The proposed application does not include any such above mentioned data either from individual for interviews/ users of the system / Company professional interviewed.

# Appendix 8: Contest Form

**Northumbria University**
**CEIS Research Ethics Sub-Committee**
**CONSENT FORM – C**

**Project Title: Intelligent Search Optimization using Artificial fuzzy logic**
**Name of the Researcher or Project Consultant: Jai Manral**
**Name of participant:**

**Participating Organization:** I consent to take part in this project. ☐

I have had the project explained to me by the researcher/ consultants **and been given an information sheet**. I have read and understand the purpose of the study.

I am willing to be interviewed.

I understand and am happy that the discussions I will be involved in may be audio-taped and notes will be taken.

I understand I can withdraw my consent at any time, without giving a reason and without prejudice.

I know that my name and details will be kept confidential and will not appear in any printed documents.

The tapes and any personal information will be kept secure and confidential. They will be kept by the researcher/project consultants until the end of the project. They will then be disposed of in line with Northumbria University's retention policy.

Anonymised summaries (if required) will be produced from the discussions to be used in the project report and in other publications. None of the participants will be identified in the project report or in other publications based on this project. Copies of any reports or publications will be available on request to participants.

I have been given a copy of this Consent Form.

**Signed:**
**Date:**

## Appendix 9: Terms of Reference

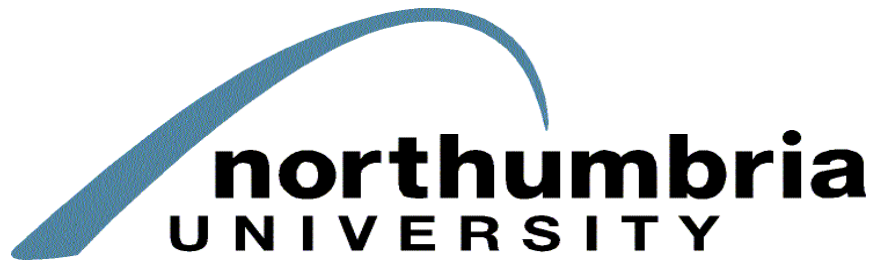

SCHOOL OF COMPUTING, ENGINEERING AND INFORMATION SCIENCES

**IS0749 Research and Project Management Assignment**
**Part C:** *Research Proposal*

*Project Title:* **Intelligent Web Search Optimisation using Artificial Fuzzy Logic.**

Student:              Jai Manral (10037359)
Programme:         MSc Computer Science
Supervisor:         Alamgair Hossain
Second Marker:     Michael Brockway
Planned Review Date: 5 /12/2011

# Table of Contents

## 1. Aim

The Aim of this research proposal is to analyse and implement the best techniques used in the search engine optimization for websites and to develop an information retrieval system using Meta search engine to see the optimization effects on the ranking of the websites in search engines using artificial fuzzy logics.

## 2. Background

### 2.1 Motivation

In today's world online search engines plays a major role as source of information in the web for the users i.e. *Consumers* (Berman and Katona, 2011). Online search is ubiquitous and with the rise in the e-commerce the competition in the ranking, on the search engines has increased rapidly (Fuxue *et al.*, 2011). With nearly 146 billion online searches every month, there is a high rise in the competition for websites to rank higher in the search engines. There are unlimited opportunities in this field but to take the advantage, it is important for the websites to rank higher in the search engines. In order for the websites to rank higher in the search results they need to follow some techniques which are known as Search Engine Optimization. As mentioned in (Frydenberg and Miko, 2011) SEO is the process of promoting a website so that it can be visible by the search engines.

There are various methods described for the optimization of the websites. As mentioned in (Fuxue *et al.*, 2011) the SEO techniques are observed from four aspects: Structure of the web-pages, keywords used, content and links. There are other researchers which believe that keywords and back links are solely responsible for the ranking of the websites (Frydenberg and Miko, 2011). There are some researchers which focus mainly on the social networking (facebook, twitter) for the ranking of the websites (Chen *et al., 2010*). Malaga (2007) in his paper states that the keywords play the major role in the optimization techniques. The SEO techniques can be of two types which are the black hat techniques and the white hat techniques (Malaga, 2010). The purpose of this research is to find out the best white hat approaches and to rank them in order to their importance.

Search engine from time to time generate the guidelines for the optimization but the process they adapt i.e. *how they rank the optimization techniques is still not know.*

The researcher of this project thus aims to find out the new techniques for optimizing the websites and the techniques used in ranking in search engine by using the artificial intelligence system. As there are different search engines available (Google, Yahoo, Bing, Microsoft) and each of them has different strategy of ranking the websites, (Hai *et al.*, 2011), thus the researcher has planned to develop a Meta Search Engine, for the practical results for the experiment.

### 2.2 Research Methods

Online businesses rely on search engines to generate traffic to their Web sites through the use of both organic and sponsored search results. (Frydenberg and Miko, 2011)

As (Stumme *et al.*, 2006) stated that the enormous success of the search engine has make it necessary to improve the search ranks thus leading a competitive market and new challenges for the website optimizers to make their websites visible in the search results.

SEO techniques requires an understanding of how exactly the search engines work and rank the pages for any particular keywords (Frydenberg and Miko, 2011).

There are various steps needed in order to make the study and achieve the desired the results:

In order to understand the optimization techniques the researcher of this report will create a set of the optimization techniques through vast study of the journals and data collected from the website development firms (Abster-iT and Portebella Rain Creative Design). The use of Web Mining is essential for studying the optimization techniques. Web Mining according to (Stumme *et al.*, 2006) can be sub divided in to three steps: Web Content Mining, Web Structure Mining and Web Content Mining. The researcher will undertake the web mining process in order to generate the techniques for optimization. The next step is to put these optimization techniques into the set, *say Q,* in the ranking order by using the fuzzy logics and neural networks.

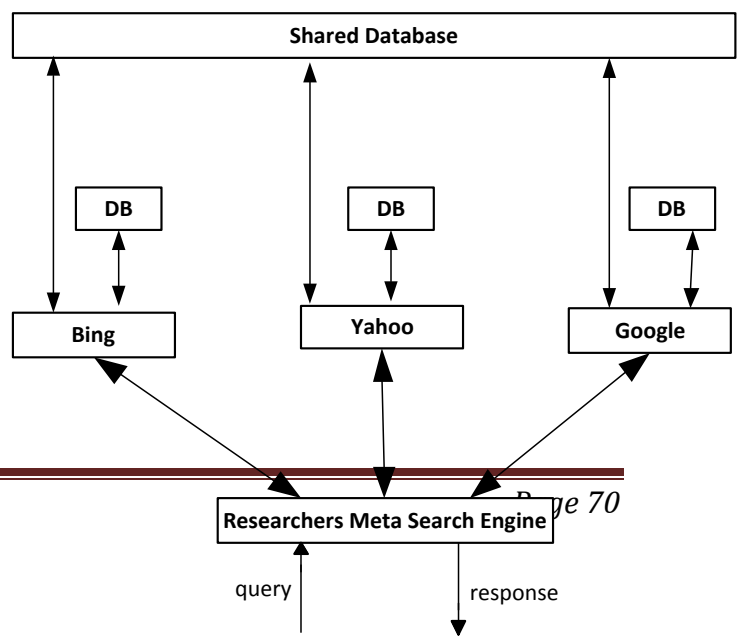

The next step is to create the Meta Search Engine. The Meta search engine has been useful in the earlier experiments conducted by the researcher's such as Thorsten for their study of Optimizing Search Engines using Clickthrough Data. The point of using Meta search engine in this research is to get the best results from the various search engines such as Google, Yahoo, MSN and Bing, so that the research topic will not be based on the specific search engines optimization techniques. The other benefit of Meta search engine usage will be that the researcher will not have to develop the database and can cover the larger segment of the documents throughout the web (Joachims, 2002).

Next step is to extract the data from Meta Search Engine and arrange them in the new ranking order using the Support Vector Machine (SVM) algorithm. It will be based on the non-linear ranking function and the parameters (ranks) of the vector will be based from the *set Q,* described earlier. As described in the Joachims (2002), the SVM algorithms are successfully used as ranking algorithm for the Meta search engine results. This will generate the new list of the refined web-pages according to the optimizing techniques.

The last step is to make the comparing of the resulted list to the list originated from the search engines (Google, Yahoo, Bing) separately and find the ranking (optimizing) parameters they used for the information retrieval.

Thus the researcher will be able to find the best techniques needed to optimize the web-pages and improve their ranking in the search engines.

### 3. Objectives

The undertaken research tries to *find important SEO techniques used for ranking result pages by search engine; how they are important in indexing and categorizing web-data and by using them, optimize search results from search engine: Google and Bing, to show their importance in generating better results to user query.* It is important to achieve following objectives in order to accomplish the aim of this research:

- To find relevant literature from articles, web sources and interviews to understand the working of search engines, i.e. how they retrieve data from web, categorize them and present it to users.
- To develop new ranking algorithm for retrieving better results to user query.
- To demonstrate the weightage of those SEO factors important in optimizing websites and improve ranking in search engines result pages.
- To develop an intelligent Meta search engine, this can retrieve better results than present search engines.
- To help webmasters by providing important SEO techniques for their websites.

### 4. Resources / Constraints

All the resources will be collected from the internet. The majority of the resources will be used from the research papers. For Web mining the detail study of the websites and the structure can be done using the internet and consulting to the Web-site Development firms (Abster-iT and Portobella Rain Creative Design). Creating of Meta search engine can be done on the software's provided by the university. This can be done on the computer labs at the Pandon Building. The Support Vector Machine and other fuzzy logics, including the neural networks can be designed in the Artificial Intelligence labs provided by the university.

Apart from the study materials the researcher needs to perform several interviews with the people who are in the business of websites. The researcher has therefore contacted some firms (Abster-iT and Portobella Rain Creative Designs) in and around the Newcastle for this purpose.

The major constrains of this research is that the field of optimization is vast and due to the limitation of the time it will be infeasible to obtain the perfect results. As the research is based on the United Kingdom thus, the search engines data which the researcher requires for this project will be extracted from the UK servers. This can alter the results if the same research is performed from different part of the world. There will be limitation for the 'keywords' to be used in search engines and the number of websites to be monitored.

## 5. Research Ethics

The materials to be used for the purpose of the research are in public domains. There is no external ethical requirement for the research work. All the data collected from the internet can be used publically and no handling of the confidential data is required. The data to be collected from the Web Designing Firms (Abster-iT and Portobella Rain Creative Design) are not confidential and can be used for the research purposes. There is no risk/ harm to the research participants and the University to carry out this project. All the materials such as the survey and the questionnaire are to be kept safely after the study has been completed. This research does not include any study which might cause the misuse of the findings to the society in person or in World Wide Web.

## 6. Schedule of Activities

| Task Name | Duration | Start | Finish | Predecessors |
|---|---|---|---|---|
| **Dissertation** | **88 days?** | **Sun 15/01/12** | **Tue 15/05/12** | |
| **Chapter 1** | **12 days** | **Sun 15/01/12** | **Mon 30/01/12** | |
| Literature Review | 6 days | Sun 15/01/12 | Fri 20/01/12 | |
| Data Analysis | 6 days | Mon 23/01/12 | Mon 30/01/12 | 3 |
| **Practical Work** | **38 days** | **Tue 31/01/12** | **Thu 22/03/12** | |
| Web Mining | 12 days | Tue 31/01/12 | Wed 15/02/12 | 4 |
| Creating (set Q) fuzzy logics | 10 days | Thu 16/02/12 | Wed 29/02/12 | 6 |
| Creating Meta Search Engine | 6 days | Thu 01/03/12 | Thu 08/03/12 | 7 |

| SVM Algorithm Design | 10 days | Fri 09/03/12 | Thu 22/03/12 | 8 |
|---|---|---|---|---|
| **Result Analysis** | **26 days** | **Fri 23/03/12** | **Fri 27/04/12** | |
| Use Meta search engine live | 8 days | Fri 23/03/12 | Tue 03/04/12 | 9 |
| Use SVM for Ranking | 6 days | Wed 04/04/12 | Wed 11/04/12 | 11 |
| Compare the result | 6 days | Thu 12/04/12 | Thu 19/04/12 | 12 |
| Analyze the results | 6 days | Fri 20/04/12 | Fri 27/04/12 | 13 |
| **Project evaluation** | 5 days | Mon 30/04/12 | Fri 04/05/12 | 14 |
| **Conclusion** | 4 days? | Mon 07/05/12 | Thu 10/05/12 | 15 |
| **Recommendation** | 2 days | Fri 11/05/12 | Mon 14/05/12 | 16 |
| **Dissertation Submission** | 1 day | Tue 15/05/12 | Tue 15/05/12 | 17 |

## Appendix 10: Turnitin UK Originality Report

TurnitinUK Originality Report

intelligenet search engine optimization using artificial intelligence by Jai Manral
From May 2012 Final Dissertation Submission (MSc Computing Projects)
Processed on 16-May-2012 10:04 AM BST
ID: 17217437
Word Count: 15942

Similarity Index
3%
Similarity by Source
Internet Sources:
3%
Publications:
0%
Student Papers:
1%
sources:
1
< 1% match (Internet from 25/10/10)
http://las.org.sg/sjlim/SJLIM20094Sampath.pdf
2
< 1% match (Internet from 17/5/10)
http://echa.europa.eu/doc/consultations/recommendations/tech_reports/tech_rep_mda.pdf
3
< 1% match (student papers from 30/04/12)
Submitted to University of East London on 2012-04-30
4
< 1% match (Internet from 9/10/09)
http://www.cyrodiil.net/forums/index.php?action=printpage;topic=1557.0
5
< 1% match (Internet from 16/2/12)
http://www.ict-rocket.eu/documents/Thesis/Irfan_MScProject_UniS_2009.pdf
6
< 1% match (Internet from 29/8/10)
http://www.hongsheng168.com/map.asp
7
< 1% match (Internet from 1/10/10)
http://mumble.net/~campbell/darcs/trc-testing/mit-test.pkg
8
< 1% match (Internet from 14/4/12)
http://saatviksolutions.net/seo-blog/seo/what-is-search-engine-optimization-seo.html
9
< 1% match (student papers from 26/04/12)
Submitted to University of Central England in Birmingham on 2012-04-26
10
< 1% match (student papers from 01/12/10)
Submitted to University of Sunderland on 2010-12-01

< 1% match (Internet from 15/5/12)

http://library.iyte.edu.tr/tezler/master/gidamuh/T000800.pdf

< 1% match (student papers from 30/09/10)

Submitted to University of Bradford on 2010-09-30

< 1% match (student papers from 30/08/10)

Submitted to University of Bradford on 2010-08-30

< 1% match (Internet)

http://e88.de/b_13593.html

< 1% match (Internet from 27/8/03)

http://db.uwaterloo.ca/~tozsu/courses/cs856/webdb.pdf

< 1% match (student papers from 03/05/12)

Submitted to University of Ulster on 2012-05-03

< 1% match (Internet from 12/4/12)

http://www.medinfo.jp/products/204890_etoc.html

< 1% match (Internet from 18/3/10)

http://www.grin.com/e-book/37396/franchising-as-a-method-of-internationalization-subway-case

< 1% match (Internet from 11/1/04)

http://www.modernspecialty.com/stats/mondart.pdf

< 1% match (Internet)

http://www.nemaug.org/DSOER/KABALE.pdf

< 1% match (Internet from 25/3/08)

http://www.workoninternet.com/article_18960.html

< 1% match (Internet from 20/4/11)

http://webusage.315169.free-press-release.com/

< 1% match (Internet from 8/3/11)

http://alexandria.tue.nl/extra2/200711913.pdf

< 1% match (Internet from 18/10/11)

http://www.rentpals.com/Alcoholic.aspx

< 1% match (publications)

Wei-Cing Hong. "Meta-Searching Chinese Health Information on the Internet", 2007 9th International Conference on e-Health Networking Application and Services, 06/2007

< 1% match (student papers from 02/09/11)

Submitted to University of Newcastle upon Tyne on 2011-09-02

< 1% match (Internet from 19/9/09)

http://www.adobe.com/devnet/seo/articles/techniques_ria_04.html